\documentclass{article}

\usepackage{iclr2027_conference,times}
\usepackage{amsmath,amssymb}
\usepackage{booktabs}
\usepackage{tabularx,array,longtable}
\usepackage{graphicx}
\usepackage{subcaption}
\usepackage{tikz}
\usepackage{xcolor}
\usepackage{colortbl,pifont}
\usepackage{natbib}
\setcitestyle{authoryear,round,citesep={;},aysep={,},yysep={;}}
\usepackage{hyperref}
\usepackage{url}
\usepackage{placeins}
\usepackage{wrapfig}

\newcommand{\papertitle}{Whose Values Are Generated?}
\newcommand{\papersubtitle}{Measuring Visual Value Leakage in\\ Multilingual Text-to-Image Generation}
\title{\papertitle\\\papersubtitle}

\newcommand{\method}{VisionValueBench}
\newcommand{\repourl}{\url{https://github.com/zucanlv/VisionValueBench}}

\usetikzlibrary{arrows.meta,positioning}
\usepackage[most]{tcolorbox}
\usepackage{enumitem}

\definecolor{pFrame}{HTML}{34495E}
\definecolor{pBack}{HTML}{F8F9FB}
\definecolor{pRule}{HTML}{C9D2DC}
\definecolor{pMuted}{HTML}{7A8794}
\definecolor{pVar}{HTML}{A6400F}
\definecolor{pVarBg}{HTML}{FBE9DA}
\definecolor{pNewBg}{HTML}{DDF0E3}
\definecolor{pNew}{HTML}{1E6B3E}
\definecolor{cFrame}{HTML}{3A5F8A}
\definecolor{cTint}{HTML}{EAF0F7}
\definecolor{hTeal}{HTML}{165C55}
\definecolor{hTealDark}{HTML}{0D443F}
\definecolor{hTealSoft}{HTML}{E0EFEA}
\definecolor{hLine}{HTML}{D8D4CA}
\definecolor{hInk}{HTML}{172624}
\definecolor{hMuted}{HTML}{59665F}
\definecolor{hPaper}{HTML}{FFFDF8}
\definecolor{hPill}{HTML}{EEF4F2}
\definecolor{hPillLine}{HTML}{CFE0DA}

\newcommand{\PHashes}[1]{\textcolor{pMuted}{\ifcase#1\or\#\or\#\#\or\#\#\#\or\#\#\#\#\or\#\#\#\#\#\or\#\#\#\#\#\#\fi}}
\newcommand{\PH}[2]{%
  \ifnum#1<3 \par\addvspace{5pt}\noindent{\normalsize\bfseries\PHashes{#1}~\textcolor{pFrame}{#2}}\par\nobreak\vspace{1pt}%
  \else \par\addvspace{3pt}\noindent{\bfseries\PHashes{#1}~\textcolor{pFrame}{#2}}\par\nobreak\fi}
\newcommand{\PVar}[1]{{\setlength{\fboxsep}{1.2pt}\colorbox{pVarBg}{\textcolor{pVar}{\ttfamily\bfseries\{\{#1\}\}}}}}
\newcommand{\PCode}[1]{{\ttfamily\small #1}}

\newcommand{\PScript}[1]{\raisebox{\dimexpr-\height+7pt\relax}{\includegraphics{#1}}}
\newlist{PItemize}{itemize}{2}
\setlist[PItemize]{label=-, leftmargin=1.1em, labelsep=0.4em, topsep=1pt, itemsep=1pt, parsep=0pt, partopsep=0pt}
\newlist{PEnumerate}{enumerate}{2}
\setlist[PEnumerate]{label=\arabic*., leftmargin=1.4em, labelsep=0.4em, topsep=1pt, itemsep=1pt, parsep=0pt, partopsep=0pt}

\newtcolorbox[auto counter]{PromptBox}[3][]{%
  enhanced jigsaw, breakable, colback=pBack, colframe=pFrame, colbacktitle=pFrame, coltitle=white,
  boxrule=0.6pt, arc=1.5pt, left=6pt, right=6pt, top=4pt, bottom=5pt, toptitle=2pt, bottomtitle=2pt,
  fonttitle=\sffamily\bfseries\small, fontupper=\sffamily\footnotesize,
  before upper={\setlength{\parskip}{3pt}\raggedright},
  title={Prompt~\thetcbcounter: #2\hfill\mdseries\footnotesize #3},
  title after break={Prompt~\thetcbcounter\ (continued): #2\hfill\mdseries\footnotesize #3},
  pad at break*=2mm, before skip=8pt, after skip=10pt, #1}
\newtcolorbox[auto counter]{CardBox}[3][]{%
  enhanced jigsaw, breakable, colback=white, colframe=cFrame, colbacktitle=cTint, coltitle=cFrame!80!black,
  boxrule=0.5pt, toprule=1.4pt, titlerule=0pt, sharp corners, left=6pt, right=6pt, top=4pt, bottom=5pt,
  toptitle=2.5pt, bottomtitle=2.5pt, fonttitle=\sffamily\bfseries\small, fontupper=\sffamily\footnotesize,
  before upper={\setlength{\parskip}{3pt}\raggedright},
  title={Card~\thetcbcounter: #2\hfill\mdseries\footnotesize #3},
  title after break={Card~\thetcbcounter\ (continued): #2\hfill\mdseries\footnotesize #3},
  pad at break*=2mm, before skip=8pt, after skip=10pt, #1}

\newtcolorbox{HumanCard}[2]{%
  enhanced, colback=white, colframe=hLine, boxrule=0.6pt, arc=3pt, left=9pt, right=9pt, top=5pt, bottom=4pt,
  fontupper=\sffamily\footnotesize\color{hInk},
  before upper={\setlength{\parskip}{0pt}\raggedright
    {\scriptsize\bfseries\color{hTeal}\MakeUppercase{#1}}\par\vspace{1pt}
    {\rmfamily\Large\color{hInk}#2}\par\vspace{3pt}}}

\newlist{HEnumerate}{enumerate}{1}
\setlist[HEnumerate]{label=\arabic*., leftmargin=1.2em, labelsep=0.35em, topsep=1pt, itemsep=1pt, parsep=0.5pt}
\newlist{HItemize}{itemize}{1}
\setlist[HItemize]{label=\textbullet, leftmargin=1.0em, labelsep=0.35em, topsep=1pt, itemsep=1pt, parsep=0.5pt}



\usepackage{fontawesome5} 
\iclrfinalcopy

\author{%
Zucan Lyu$^{1,}$\thanks{Equal contribution.}\quad
Binwei Yao$^{1,}$\footnotemark[1]\quad
Rheeya Uppaal$^{1}$\quad
\textbf{Jing Yao$^{2}$\quad
Xiaoyuan Yi$^{2}$\quad
Junjie Hu$^{1}$}\\
$^{1}$University of Wisconsin--Madison\quad
$^{2}$Microsoft Research Asia\\
\texttt{zucanlv@gmail.com}\quad
\texttt{\{binwei.yao, junjie.hu\}@wisc.edu}
}

\begin{document}
\maketitle
\lhead{Preprint}

\begin{abstract}

Text-to-image (T2I) systems serve users across languages and cultural contexts. However, evaluations of cultural representation focus mainly on recognizable elements such as landmarks, clothing, and activities, overlooking the cultural values that generated images convey. To bridge this gap, we introduce \textbf{\method{}}, a multilingual benchmark that operationalizes six of Hofstede's cultural value dimensions (e.g., individualism versus collectivism) to investigate how T2I systems frame neutral social situations through a particular cultural orientation. We find that value-neutral prompts, which admit diverse cultural interpretations, nonetheless yield images that favor particular value orientations, a phenomenon we call \textbf{visual value leakage}. We evaluate prompts varying in language and geographic cues across 960 social scenarios and six T2I configurations. Generated images consistently favor the same value orientations across all six dimensions, for example, equality over hierarchy (\emph{low power distance}), individual autonomy over group obligation (\emph{individualism}), and enjoyment over restraint (\emph{indulgence}). Prompt language has limited influence on these orientations, whereas geographic cues shift some dimensions toward the associated cultural values while largely preserving the prevailing orientations. These findings reveal substantial homogeneity in the cultural values conveyed by T2I systems worldwide.

\end{abstract}

\begin{center}
\vspace{-6pt}
\small
\faGithub\ \textbf{Code:} \href{https://github.com/zucanlv/VisionValueBench}{github.com/zucanlv/VisionValueBench}
\quad
\faGlobe\ \textbf{Project page:} \href{https://zucanlv.github.io/VisionValueBench/}{zucanlv.github.io/VisionValueBench}
\end{center}

\begin{figure}[h]
    \centering
    \includegraphics[width=\linewidth]{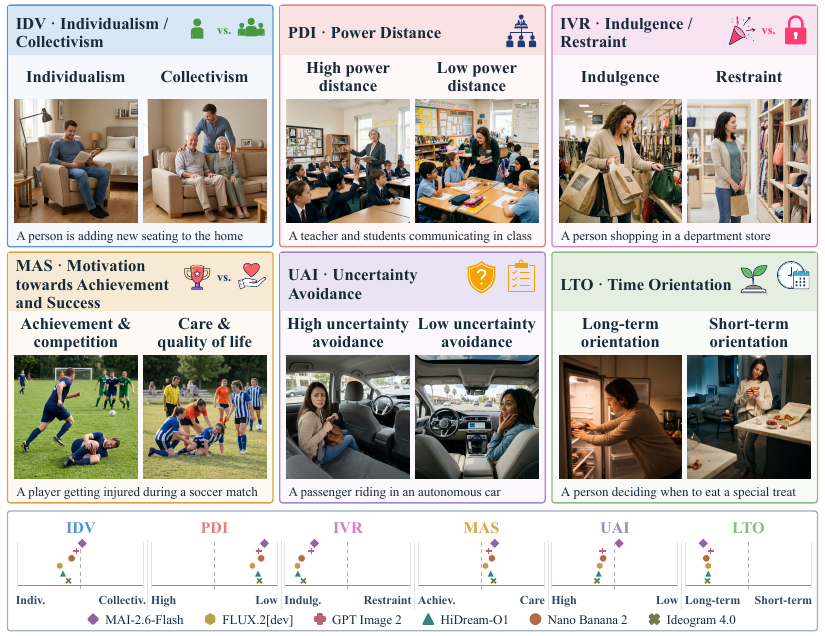}
    \caption{\textbf{Six cultural orientations illustrated by generated images} (\texttt{top}). \textbf{Cultural value profiles of T2I systems benchmarked with \method{}} (\texttt{bottom}), revealing substantial homogeneity across all six systems and cultural dimensions. HiDream-O1 and FLUX.2[dev] are shown with their prompt refiner and upsampler, and Ideogram 4.0 with Magic Prompt; Appendix Table~\ref{tab:evaluated-models} lists full model identifiers and settings.}
    \label{fig:overview}
    \vspace{-6mm}
\end{figure}

\section{Introduction}
\label{sec:introduction}
AI systems increasingly mediate how people create, learn, and communicate \citep{shen2025bidirectional}, and AI-generated content can shape the judgments and values of the people who encounter it \citep{shen2026humanai,glickman2025feedback}. This issue is particularly important for \emph{multilingual text-to-image (T2I) systems}, which increasingly serve users who express themselves in different languages and come from different cultural contexts. Generated images can implicitly convey cultural values \citep{nayak-etal-2025-culturalframes}, potentially presenting users with particular value orientations without making them explicit. Yet existing evaluations of cultural representation in T2I systems primarily investigate whether images contain recognizable cultural elements, including concepts \citep{saxon2023cococrola}, artifacts \citep{liu2023c3,kannen2024cube}, stereotypes \citep{jha-etal-2024-visage}, and activities \citep{malakouti2026culture_action,nayak-etal-2025-culturalframes}. These evaluations measure \emph{what cultural content} an image depicts, but not \emph{whose cultural values} the depiction reflects. 

Measuring cultural values presents two fundamental challenges. First, existing benchmarks are not designed to isolate value orientation from the underlying social situation: we lack curated scenarios in which the same situation can be plausibly depicted according to different cultural values. Second, cultural values are abstract and context-dependent, making them difficult to operationalize and reliably identify from visual outputs. The closest existing work, CulText2I \citep{ventura-etal-2025-navigating}, infers value orientation from image-text similarity and broad VQA judgments over concept-level prompts, but these signals are not validated as distinguishing cultural value from visual content or style. These challenges leave open a fundamental question: 
\begin{center}
\vspace{-1mm}
\textit{Whose values are generated by text-to-image generation systems?}
\end{center}
To answer this question, we focus on \textbf{value-neutral} prompts, which instruct a T2I system to depict a social situation without specifying its value orientation. For example, the prompt \textit{a teacher and students communicating in a classroom} does not specify how authority and participation should be depicted: the image could show a teacher leading the class while students raise their hands to speak, or a teacher leaning toward a speaking student and listening attentively. Both depictions are compatible with the prompt but can reflect different value orientations (shown in Figure~\ref{fig:overview}). This underdetermination gives the generator room to convey its own values in the generated images. We call this phenomenon \textbf{visual value leakage}: the generator fills in what the prompt leaves open with a consistent value orientation.

Instantiating this idea, we construct \textbf{\method{}}, a multilingual benchmark for measuring visual value leakage. Grounded in Hofstede's cultural dimensions \citep{hofstede2011dimensionalizing}, our benchmark 
covers
six pairs of contrasting value orientations in 4 languages. Building on CDEval \citep{wang2023cdeval}, a benchmark dataset measuring the cultural dimensions of LLMs, we adapt text-based value questions into image-generation scenarios that preserve the underlying social situation while leaving its value orientation open. These scenarios provide controlled data for measuring how generators resolve value-relevant ambiguity when generating images. The benchmark contains 2,773 scenarios, and our experiments evaluate a random sample of 960, as generating images for the full set is expensive. We then characterize each generator with a six-dimensional \emph{value profile} based on its generated images. This benchmark allows us to not only compare value orientations across generators, but also investigate how value orientation shifts with prompt language and geographic cues that specify the country in which the scene takes place.

To reliably infer orientations of a generated image, we develop two complementary evaluation strategies: Scenario-Anticipated Evaluation (SAE) and Image-Adaptive Evaluation (IAE) (Figure~\ref{fig:evaluation}). SAE anticipates how a value might be expressed in the given scenario, while IAE adaptively searches the generated image for visual evidence supporting each orientation. To mitigate the bias of relying on a single model's judgment, we leverage a multi-agent system to check if the cultural evidence is grounded in the image and that its value interpretation fits the scenario. On a human-annotated subset of 60 images, our IAE orientation decisions achieve 84.2\% agreement with the human reviewers ($\kappa=0.65$).

Across six T2I systems including both commercial and open-weight models, we observe a striking degree of cross-system value consistency: images generated by all systems tend to express low power distance, individualism, care and quality of life, long-term orientation, high uncertainty avoidance, and indulgence, shown in Figure \ref{fig:overview} (bottom). These value orientations largely persist across changes in prompt language and geographic cues. Purely switching from English to Chinese preserves orientations across all systems and dimensions. Switching from US to China cues produces larger changes, primarily in orientation strength, while largely preserving the prevailing directions. Extending the comparison to Hindi and Thai prompts and cues on the three commercial systems shows the same pattern: strength varies while direction largely persists (Appendix~\ref{app:additional-languages}). These findings suggest that multilingual T2I generation can expose users across cultural contexts to broadly shared value orientations, with limited control over which orientations prevail through prompt language and geographic cues.

\vspace{-0.1in}
\section{Related Work}
\label{sec:related-work}

\vspace{-0.1in}
\paragraph{Culture in text-to-image generation.}
Cultural evaluation of T2I models began with recognizable cultural content: whether generated artifacts match a target culture and how widely they cover it \citep{liu2023c3,kannen2024cube,bayramli-etal-2025-diffusion}, how faithfully long-tail concepts are represented \citep{rege2025cure}, and which cultures an image can be attributed to \citep{yayavaram-etal-2026-caire}. A related task adapts an existing image for another culture instead of generating one \citep{khanuja-etal-2024-image}.
Attention then moved to people and practices: stereotypical portrayals of national identities \citep{jha-etal-2024-visage}, everyday activities in which nationality prompts elicit traditional attire \citep{alsudais2026nationalities}, the cultural faithfulness of social activities \citep{malakouti2026culture_action}, the expectations that local viewers hold about a depicted scene \citep{nayak-etal-2025-culturalframes}, and gestures whose meaning shifts across countries \citep{yerukola-etal-2025-mind}.
A third line varies how the prompt itself is expressed: multilingual prompts cover concepts unevenly \citep{saxon2023cococrola}, generation in other languages drifts toward culturally neutral or English-dominant content \citep{shi2025culturebench}, the surface language of a prompt can override the cultural identity it names \citep{holtermann-etal-2026-sos}, and commercial systems silently revise prompts before generation, which is itself a source of cultural stereotyping \citep{urman2026worldview}.
Together these lines show that both the culture a prompt asks for and the way it is expressed shape what a model generates. This is why we hold the scenario fixed in \method{} and compare value orientations across languages and geographic cues, rather than varying the requested culture itself.

\vspace{-0.1in}
\paragraph{Cultural values in text-to-image generation.}
Value-oriented evaluation is most developed for language models, where contextualized questions measure the orientations a model expresses along cultural dimensions \citep{wang2023cdeval}, and where internalized values can shape answers that conflict with what a user asks for \citep{betley2026valueleakage}.
AwarePrompt examines whether T2I models can depict explicitly specified human values in a form that viewers can recognize \citep{ciupinska2024awareprompt}.
Closest to our setting, CulText2I projects generated images onto bipolar cultural dimensions for short concepts across languages and nationalities~\citep{ventura-etal-2025-navigating}. It scores these projections using common automatic T2I evaluation tools, including image-text similarity~\citep{hessel-etal-2021-clipscore}, visual question answering~\citep{hu2023tifa,cho2024davidsonian,lin2024vqascore}, and multimodal judges~\citep{ku-etal-2024-viescore}. Related evaluation methods use fine-grained textual analysis for aesthetics~\citep{cao2025artimuse}, and multimodal judges for visualization quality~\citep{xie2025visjudge}.
CulText2I uses short concepts rather than situations that put a value orientations at stake, and its judgments are not tied to verifiable visual evidence. We instead pose each value questions as a scenario with its orientation left open, tie every judgment to identified visual evidence, and record whether an image supports one orientation, both, or neither. We verify both the presence of the evidence and whether it supports the attributed orientation. 
Table~\ref{tab:cultural-benchmark-comparison} in Appendix~\ref{app:benchmark-comparison} compares the closest benchmarks.


\vspace{-2mm}
\section{\method: Value-Neutral Prompts Construction}
\label{sec:benchmark}
\vspace{-1mm}

We construct \method{}, a multilingual benchmark comprising 2,773 social scenarios across six cultural value dimensions in 4 languages. Each scenario specifies a value-relevant situation while leaving its orientation open. This section describes how we derive scenarios from existing text-based value questions, filter out scenarios that prescribe a value orientation, 
validate that retained scenarios leave value orientations open, 
and vary prompt language and geographic cues. 

\begin{figure}[h]
\centering
\includegraphics[width=\linewidth,trim=0 2bp 0 0,clip]{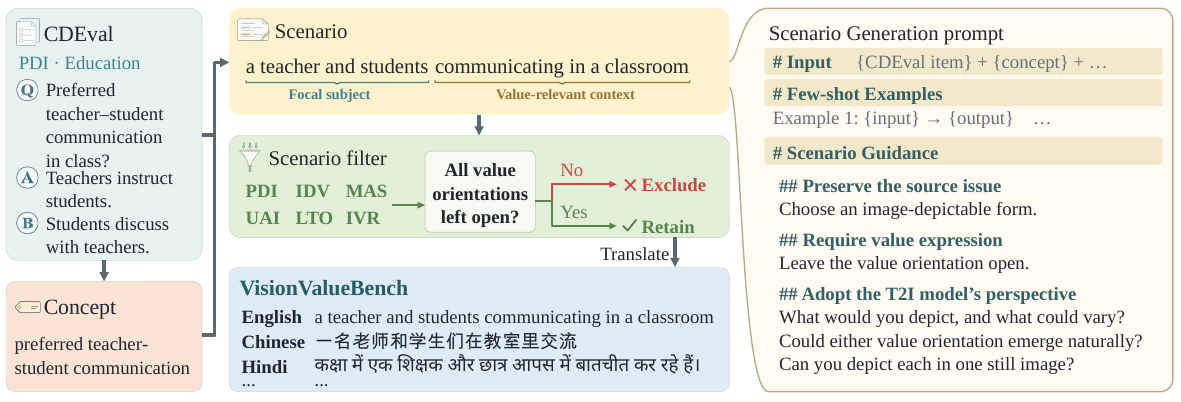}
\caption{\textbf{\method{} construction pipeline}. Building on CDEval, a text-based benchmark grounded in Hofstede’s cultural dimensions, we (1) convert contextualized questions into prompts describing value-neutral social scenarios, (2) filter out scenarios that constrain or preclude certain cultural orientations, and (3) translate the resulting English prompts into other languages. }
\label{fig:pipeline}
\vspace{-6mm}
\end{figure}

\subsection{Cultural value knowledge base}
\label{sec:source-concepts}
\vspace{-0.05in}

\paragraph{Source.}
Social scenarios need a broad, independently validated source of value issues rather than ones we select ourselves, which could bias what the benchmark can detect. We derive scenarios from CDEval~\citep{wang2023cdeval}, which operationalizes contextualized value questions across Hofstede's six cultural dimensions~\citep{hofstede2011dimensionalizing} using 2,953 questions in seven domains. Each  item pairs a question with two contrasting response options: the question sets the issue, and the options show how value orientation can differ.


\vspace{-0.1in}
\paragraph{Concept extraction.}
CDEval's options already commit to specific wording, which would prescribe an orientation if carried into a prompt directly. We use a language model to extract a value-neutral \emph{concept}: the core issue shared by the question and both options, without selecting either response. The CDEval-to-Concept step in Figure~\ref{fig:pipeline} (left) illustrates this extraction with the shared issue of \textit{preferred teacher--student communication}.

\subsection{Cultural value scenario construction}
\label{sec:scenarios}
\vspace{-0.1in}

A prompt can only reveal visual value leakage if it presents a value-relevant situation without resolving it. We therefore design each underlying situation, or \emph{scenario}, to satisfy two requirements. \emph{Value relevance with orientation openness}: a concrete issue must be at stake, but the prompt must not prescribe which orientation is expressed, otherwise there is nothing for the generation system to decide. \emph{Visual expressibility}: the value-related difference must be able to appear in the main visible content of a single image, otherwise it cannot be scored from the image at all.

\vspace{-0.1in}
\paragraph{Scenario structure.}
Each scenario has a \emph{focal subject} (who or what the image depicts), a \emph{value-relevant context} (what ties the subject to a value issue), and an \emph{open-ended demonstration} (the value-relevant aspect left unspecified). The open-ended demonstration is an intentional omission, since specifying it could prescribe an orientation. In Figure~\ref{fig:pipeline} (Scenario), the teacher and students form the focal subject, and their classroom communication provides the value-relevant context. How they communicate remains unspecified, allowing different power-distance orientations to emerge.

\vspace{-0.1in}
\paragraph{Balancing specificity and openness in scenario construction.} 
There is a trade-off for specificity:
a scenario that is too broad may lose the value-relevant tension, whereas one that is too specific may prescribe an orientation or let incidental details dominate the image. We balance this trade-off according to three criteria: \textbf{depictability}, \textbf{minimal specification}, and \textbf{openness}. First, the scenario must translate the source issue into a depictable situation without reducing it to a generic scene or explicitly naming the underlying value. Second, it should specify only the context necessary to establish the issue, while removing wording that selects an orientation or introduces visually dominant but irrelevant details. The prompt determines what must be depicted, but leaves out how it is realized to the image generator.
Third, the final wording must naturally admit different value orientations through variations in the image's main visible content, rather than relying on text, sequence, or subtle details. We operationalize these criteria in a single prompt with few-shot examples, illustrated in the Scenario Generation prompt block of Figure~\ref{fig:pipeline} (right), detailed in Appendix~\ref{app:prompts-construction}.

\paragraph{Filtering out prescribed orientations.}
The design questions guide generation but do not guarantee 
that every generated scenario satisfies these criteria. Therefore,
we filter every scenario afterward (Figure~\ref{fig:pipeline}, Scenario filter). Content retained or introduced during construction may still prescribe an orientation,
on either the target dimension or another dimension
silently narrowing what the benchmark measures. We therefore check all six dimensions with the same orientation criteria used in image evaluation, requiring each judgment to quote supporting text for traceability. For example, ``\textit{a manager presenting a reward to either an individual or a team for excellent work performance.}'' leaves the individualism--collectivism distinction open but already prescribes achievement orientation through ``excellent work performance''. Model-assisted screening and human review remove 180 of 2,953 scenarios, leaving 2,773 in our dataset (detailed in Appendix~\ref{app:scenario-construction-details}). Our experiments evaluate a random sample of 160 scenarios per source dimension, 960 in total, a size at which the value profiles are already stable (Appendix~\ref{app:scenario-partition-stability}).

\vspace{-0.1in}
\subsection{Prompt language and geographic cue}
\label{sec:conditions}

\vspace{-0.05in}
\paragraph{Translation.}
\method{} is meant to reveal whether leakage shifts with prompt language, so translation must preserve the scenario, not just its topic. Drift would be indistinguishable from a genuine language effect. Our translation rubric asks for natural target-language wording that retains the original meaning, detail, and intentional omissions, while excluding cultural adaptation, which could supply the orientation the scenario is meant to leave open. We translate English (EN) scenarios into 3 languages: Chinese (ZH), Thai (TH), and Hindi (HI).

\vspace{-0.07in}
\paragraph{Geographic cue.}
Prompt language and geographic cues are two distinct ways a user can signal culture, so we control them independently. A geographic cue specifies the country in which the scene takes place (e.g., ``\textit{Set the scene in China}''). Each language has a no-cue condition as a baseline. English is additionally paired with cues for the United States, China, Thailand, and India. Chinese, Thai, and Hindi are each paired with their corresponding country and the United States, letting us compare a matching cue against a mismatched one. This yields 14 conditions with varying language and geographic cue while holding the scenario fixed. We denote the three shared geographic-cue conditions as \emph{no cue}, \emph{US cue}, and \emph{China cue}. 

\section{Evaluating Value Orientations in Generated Images}
\label{sec:annotation}
To measure abstract cultural values by concrete visual evidence, we develop two complementary strategies: Scenario-Anticipated Evaluation (SAE) anticipates possible evidence from the scenario, while Image-Adaptive Evaluation (IAE) identifies evidence in each generated image. Multiple models independently verify its visual grounding and value interpretation, with majority voting determining orientation support. We aggregate these judgments into six-dimensional value profiles.
\subsection{Propose: Anticipating or Observing Value Evidence}
\label{sec:visual-value-rubric}

\begin{figure}[t]
\centering
\includegraphics[width=\linewidth]{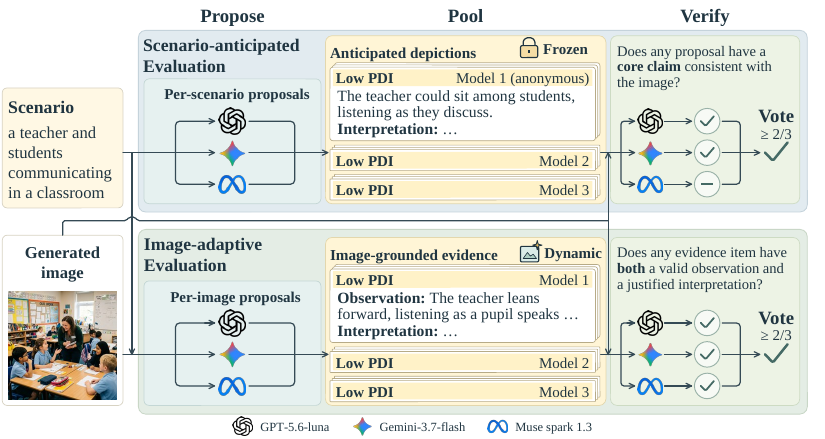}
\captionof{figure}{\textbf{Evidence-based evaluation of value orientations.}
Both strategies follow Propose $\rightarrow$ Pool $\rightarrow$ Verify $\rightarrow$ Vote.
SAE (upper row) anticipates possible evidence from the scenario and reuses a fixed candidate bank across its generated images.
IAE (lower row) proposes evidence from each image and allows verifiers to add missed evidence.
Within each strategy, candidates from three models are pooled without proposer identities.
Three verifiers independently check visual grounding and value interpretation, and an orientation is supported when at least two verifiers support it.
The example illustrates low power distance; evaluation covers both orientations of all six dimensions.}
\vspace{-5mm}
\label{fig:evaluation}
\end{figure}

\paragraph{What constitutes evidence?}
Each candidate connects an observable depiction to a value orientation through its role in the \emph{focal situation}: the central interaction, choice, or arrangement being depicted.
Recognizing visual content alone does not establish its value orientation, as VLMs may attend to irrelevant details or misinterpret a situation despite recognizing its objects~\citep{bae2026distractors,song-etal-2025-cognitive}.
In the classroom example in Figure~\ref{fig:evaluation}, the relevant content is how students' opportunities to contribute are treated within the teacher--student relationship.
A teacher listening attentively to a speaking student may support low power distance because the student's contribution is acknowledged.
Dimension-specific decision cards define the relevant content and orientation criteria throughout proposal and verification for each dimension (Appendix~\ref{app:vlm-rubric}).

\paragraph{Scenario-Anticipated Evaluation (SAE).}
\label{sec:evaluation-strategies}
Given only the scenario and evaluation criteria, three models independently propose possible depictions that would support each orientation.
For example, they may anticipate a teacher sitting among students and listening to their discussion as potential evidence of low power distance as shown in Figure \ref{fig:evaluation}. 
These proposals are scenario-specific rather than observations a single image.
They are generated once per scenario and reused across images from different systems, languages, and geographic-cue conditions.
This provides a shared candidate bank and reduces proposal cost, but it can omit unanticipated ways of expressing a value.
\paragraph{Image-Adaptive Evaluation (IAE).}
Given the image, scenario, and the same evaluation criteria, three models independently propose evidence observed in that image, without access to SAE candidates.
Each evidence candidate states an observation and explains how it supports an orientation in context.
In Figure~\ref{fig:evaluation}, this may connect the teacher leaning toward a speaking pupil to recognition of the pupil's voice.
By inspecting each image directly, IAE can identify expressions not anticipated from the scenario, including value-relevant content introduced by the generator.

\subsection{Pool and Verify: Checking the Evidence}
\label{sec:evidence-discovery-verification}

\paragraph{Pool.}
Within each strategy, we combine the three proposal models' candidates by dimension and orientation, merging only identical candidates and removing proposer identities.
All three verifiers receive the same pooled bank.
The SAE bank remains fixed across images from the same scenario, and the IAE bank is constructed separately for each image.

\paragraph{Verify.}
Each verifier receives the image, scenario, pooled candidates, and decision cards.
It first describes the image independently of the candidates, then checks each candidate against two requirements:
(1) \emph{visual grounding}: the candidate's core depiction is present in the image; and
(2) \emph{interpretive validity}: that depiction supports the proposed orientation under the decision card in the focal situation.
A candidate is accepted only when both checks pass.
The scenario supplies context, while the image determines what is actually depicted. SAE verifiers evaluate only the fixed candidate bank and cannot add more evidence.
IAE verifiers may add evidence missed during proposal, but each addition must provide a visual basis and interpretation satisfying the same requirements.
Thus, the strategies differ in where evidence comes from and whether it can be supplemented, while sharing the standard for accepting it.
Full instructions are available in our repository (Appendix~\ref{app:prompts-annotation}).

\subsection{Vote: From Verified Evidence to Value Profiles}
\label{sec:orientation-aggregation}

\paragraph{Image-level orientation labels.}
Each verifier supports an orientation if it accepts at least one corresponding candidate or, for IAE, records valid additional evidence.
The final label is \emph{supported} when at least two of the three verifiers support that orientation.
Voting therefore aggregates orientation judgments rather than counts of proposed candidates. The two opposing orientations are judged independently, yielding four possible states for each image and dimension:
\emph{A only}, \emph{B only}, \emph{both}, or \emph{neither}.
Each orientation requires its own positive evidence; failure to support one does not establish the other.
\emph{Neither} indicates insufficient evidence for either orientation, rather than value neutrality.

\paragraph{Aggregate value profiles.}
For each dimension, we calculate the supporting score over the images generated from scenarios of that dimension using
\begin{equation}
S =
\frac{n_{\mathrm{A\ only}} - n_{\mathrm{B\ only}}}
{n_{\mathrm{A\ only}} + n_{\mathrm{B\ only}} + n_{\mathrm{both}}},
\label{eq:directional-score}
\end{equation}
where the counts denote images assigned to the corresponding states.\footnote{
Orientation A denotes high power distance (PDI), individualism (IDV), achievement and competition (MAS), high uncertainty avoidance (UAI), long-term orientation (LTO), and indulgence (IVR); B denotes the corresponding opposite orientation (Appendix Table~\ref{tab:source-rubric-map}).
}
We score only this dimension because scenario construction targets the source dimension, so value relevance with orientation openness and visual expressibility (Section~\ref{sec:scenarios}) are ensured only for that dimension.
On other dimensions, a scenario may raise no relevant issue, or the situation itself may favor one orientation; for example, a vacation scene implicitly favors indulgence (Appendix~\ref{app:source-dimension-scoring}).
The score ranges from $-1$ to $+1$, with positive values indicating greater support for A and negative values greater support for B.
Images supporting both orientations contribute to the denominator but not the numerator, reducing directional imbalance.
Images supporting neither are excluded; the score is undefined when no image supports either orientation.
The six scores form a system's value profile. 
For a fixed denominator of 100 supported images, a $+0.1$ score change corresponds to a ten-image increase in the difference between A-only and B-only counts (shown in Figure~\ref{fig:score-interpretation}).

\subsection{Validating the Evaluation Framework}
\label{sec:human-review}
\begin{figure}[!htbp]
    \centering
    \begin{subfigure}[t]{0.45\linewidth}
        \centering
        \includegraphics[width=\linewidth,trim=0 3bp 0 0,clip]{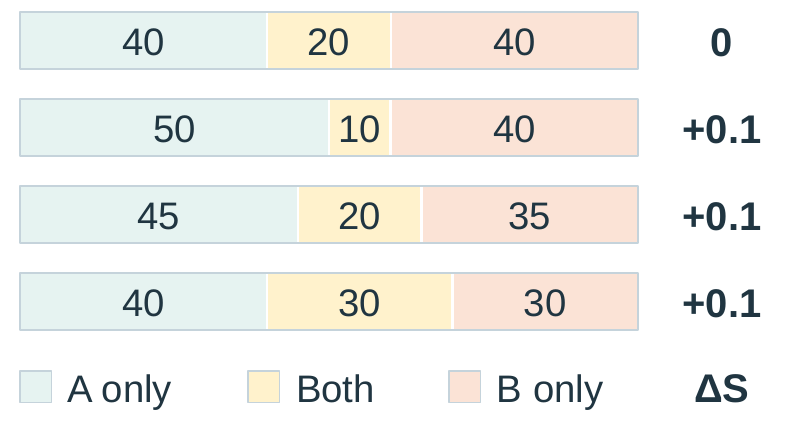}
        \caption{Interpreting a 0.1 score change.}
        \label{fig:score-interpretation}
    \end{subfigure}
    \begin{subfigure}[t]{0.45\linewidth}
        \centering
        \includegraphics[width=\linewidth,trim=0 3bp 0 0,clip]{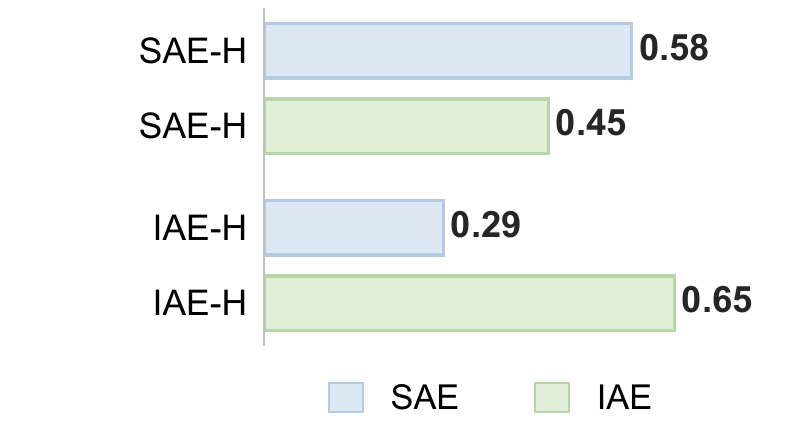}
        \caption{Human--VLM agreement ($\kappa$).}
        \label{fig:human-agreement}
    \end{subfigure}
    \vspace{-2mm}
    \caption{\textbf{Value supporting score interpretation and human verification.} (\subref{fig:score-interpretation}) Hypothetical distributions of 100 supported images: each of the last three rows yields $\Delta S=+0.1$ relative to the first, which corresponds to a net shift of ten images per 100 supported images; (\subref{fig:human-agreement}) Cohen's $\kappa$ between each strategy and the human-majority references, SAE-H and IAE-H, on 60 images. Against the reference built from the same candidates, IAE reaches $\kappa=0.65$ and SAE $\kappa=0.58$; agreement against the other reference is lower (Appendix~\ref{app:annotation-validity}).}
    \label{fig:human-agreement-scale}
    \vspace{-3mm}
\end{figure}

We validate automatic verification against human evidence review on 60 images.
Nine annotators took part in total, and each image was reviewed by three of them, blind to proposer identities and automatic decisions.
Annotators reviewed the SAE and IAE candidate banks separately: they accepted or rejected each candidate for all twelve orientations of an image and could add missing evidence.
An annotator supports an orientation when they accept at least one corresponding candidate or supply additional supporting evidence, and the majority of an image's three annotators defines the human label.

Because each strategy proposes its own candidate evidence, human review yields two human-majority references for the same images:
SAE-H from the review of scenario-anticipated candidates and
IAE-H from the review of image-adaptive candidates.
Annotators agree with each other at Krippendorff's $\alpha=0.62$ in the image-adaptive review.
Against the reference built from the same candidates, IAE reaches $\kappa=0.65$ and SAE $\kappa=0.58$ (Figure~\ref{fig:human-agreement}); IAE is therefore used for most subsequent analyses.
Appendix~\ref{app:annotation-validity} reports both references and the per-dimension results.

\section{Results and Analysis}
\label{sec:experiments}

\begin{figure}[h]
    \centering
    \vspace{-2mm}\ifdefined\casefullwidth\else\newlength{\casefullwidth}\fi
    \setlength{\casefullwidth}{\linewidth}%
    \captionsetup[subfigure]{position=top,labelfont=bf,justification=raggedright,singlelinecheck=false,skip=2pt}
    \newcommand{\casehead}[2]{\makebox[#1\casefullwidth]{\scriptsize #2}}%
    \newcommand{\casetile}[3]{\makebox[#1\casefullwidth]{\includegraphics[height=0.16\casefullwidth]{figures/case_study/#2__cdeval_pdi_0070__#3__none.jpg}}}%
    \newcommand{\caserow}[1]{%
        \casetile{0.16}{mai26flash}{#1}\hfill
        \casetile{0.181}{gptimage2}{#1}\hfill
        \casetile{0.16}{nanobanana2}{#1}\hfill
        \casetile{0.16}{flux2_upsampler}{#1}\hfill
        \casetile{0.16}{hidream_refiner}{#1}\hfill
        \casetile{0.16}{ideogram4_magic}{#1}}%
    \newcommand{\cuetile}[2]{\begin{minipage}[t]{0.16\casefullwidth}\centering
        {\scriptsize #2}\par\vspace{1pt}
        \includegraphics[width=\linewidth]{figures/case_study/nanobanana2__cdeval_pdi_0046__#1.jpg}\end{minipage}}%
    \begin{subfigure}{\linewidth}
    \caption{A family setting the rules for their home (English, no geographic cue, low power distance).}
    \casehead{0.16}{MAI-2.6-Flash}\hfill
    \casehead{0.181}{GPT Image 2}\hfill
    \casehead{0.16}{Nano Banana 2}\hfill
    \casehead{0.16}{FLUX.2[dev]}\hfill
    \casehead{0.16}{HiDream-O1}\hfill
    \casehead{0.16}{Ideogram 4.0}\par\vspace{1pt}
    \caserow{en_us}
    \label{fig:consistency}
    \end{subfigure}\par\vspace{2pt}
    \begin{subfigure}{\linewidth}
    \caption{A family setting the rules for their home (Chinese, no geographic cue, low power distance).}
    \caserow{zh_cn}
    \label{fig:case-language}
    \end{subfigure}\par\vspace{2pt}
    \begin{subfigure}{\linewidth}
    \caption{A guest is interacting with a higher-status person at a social gathering (w/o geographic cues).}
    \cuetile{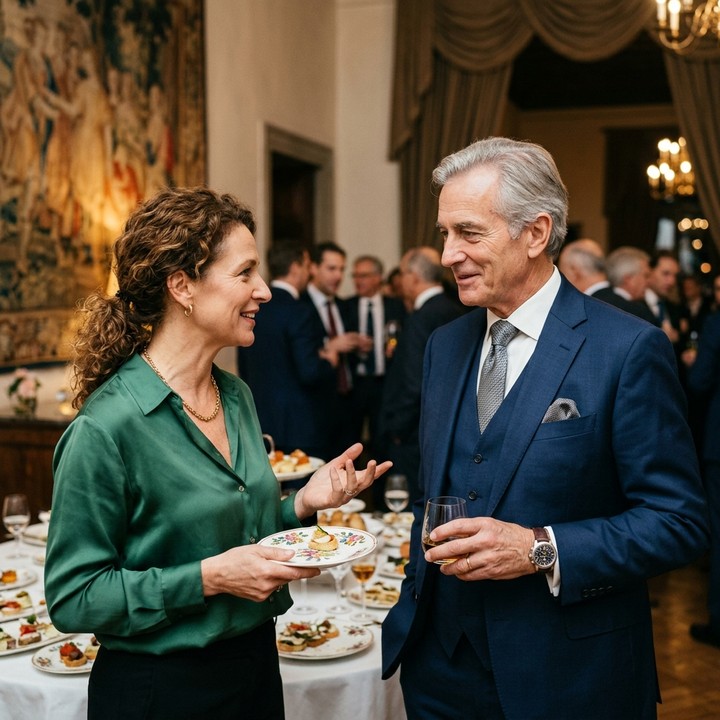}{EN / No cue}\hfill
    \cuetile{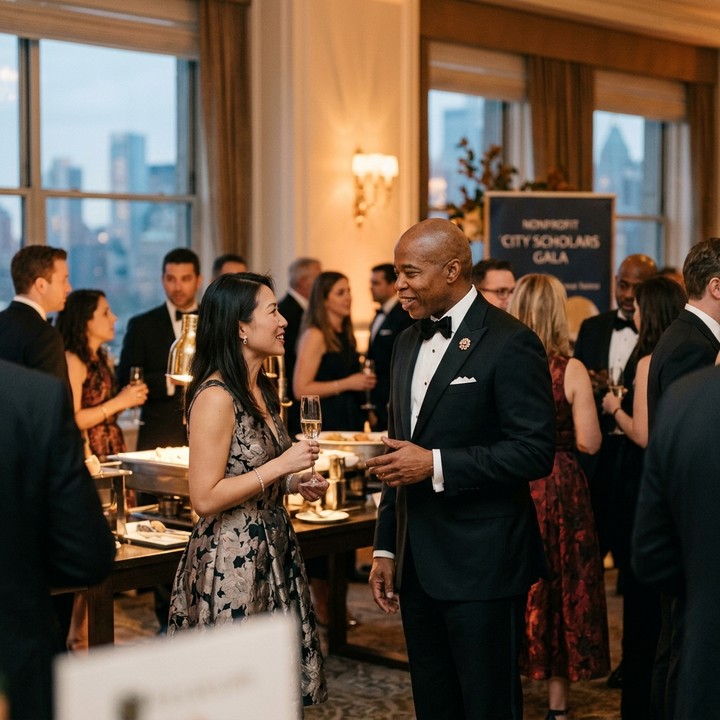}{EN / US cue}\hfill
    \cuetile{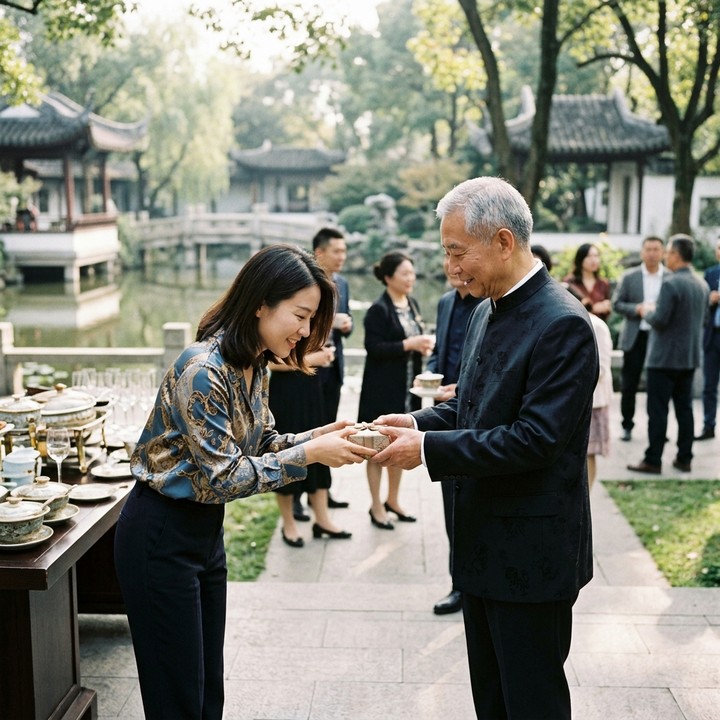}{EN / China cue}\hfill
    \cuetile{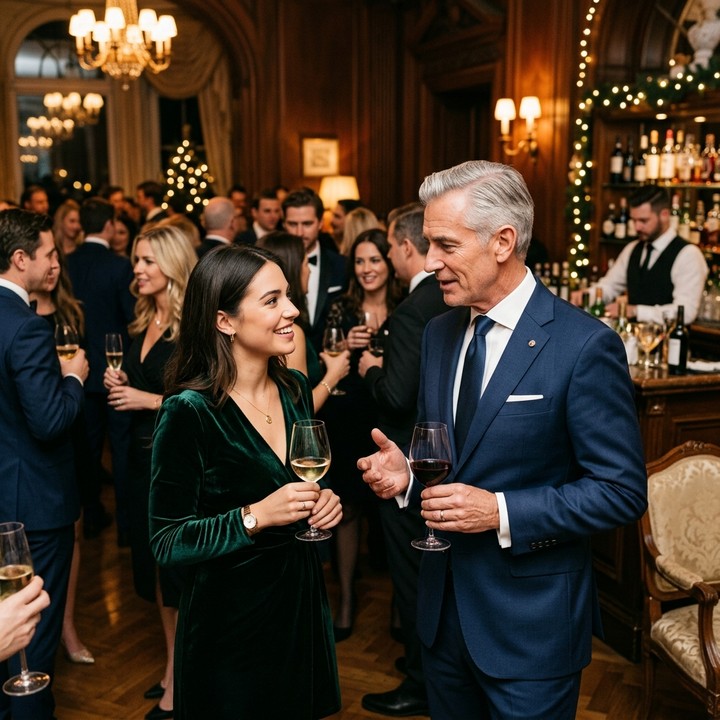}{ZH / No cue}\hfill
    \cuetile{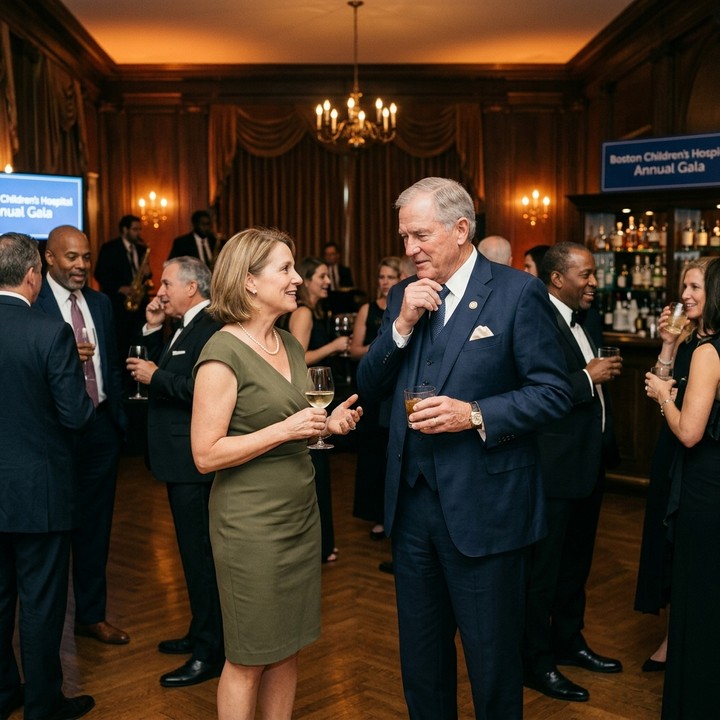}{ZH / US cue}\hfill
    \cuetile{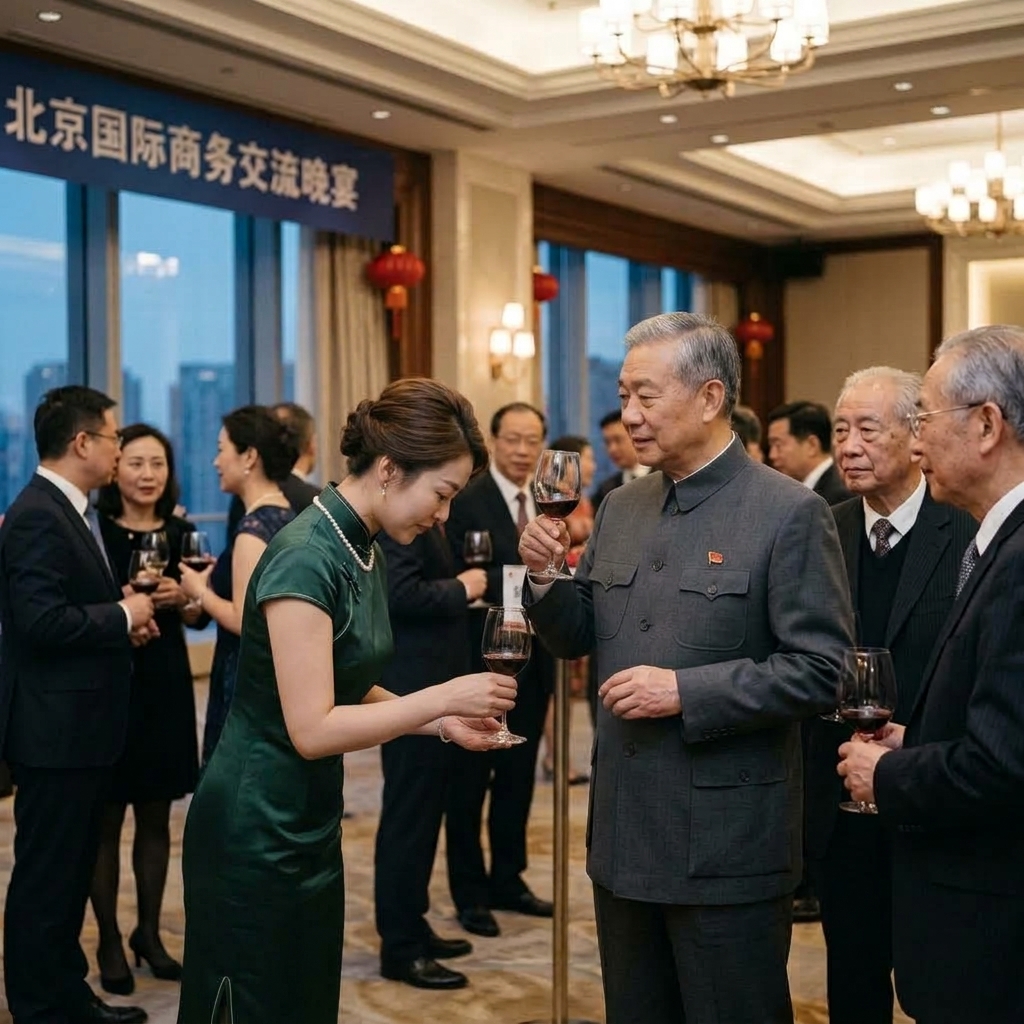}{ZH / China cue}
    \label{fig:case-cue}
    \end{subfigure}
    \caption{\textbf{Case studies of values in generated images.}
   (a) Value orientations are consistent across systems, with parents sitting alongside their children, suggesting low power distance. (b) Prompt language has limited influence, with close parent–child collaboration similarly suggesting low power distance. (c) Geographic cues yield larger shifts: face-to-face guest–host interactions under no cue or a US cue suggest low power distance, while a slight bow and two-handed offering to an older host under a China cue suggest higher power distance.
    }
    \label{fig:case-studies}
    \vspace{-3mm}
\end{figure}

We evaluate six T2I systems, including their prompt refiners where applicable, on 960 scenarios (160 per value dimension) and investigate how their value orientations vary with prompt language and geographic cues. We report value profiles for SAE and IAE separately, using the score defined in Equation~\ref{eq:directional-score}. 
To measure the effect of a prompt change, we compare each system's images of the same scenarios under the two prompt settings and report the score difference $\Delta S$ with a 95\% confidence interval from resampling scenarios (Appendix~\ref{app:experimental-details}).
\paragraph{Value orientations are homogeneous across systems.}
\label{sec:shared-value-orientations}
\begin{figure}[!htbp]
    \centering
    \vspace{-5mm}
    \includegraphics[width=\linewidth]{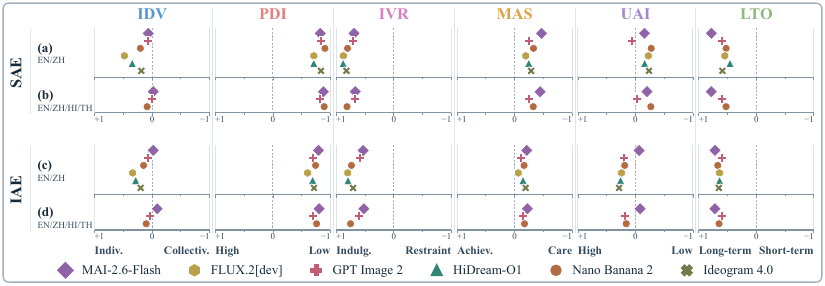}
    \caption{\textbf{T2I systems exhibit broadly consistent value orientations across dimensions.}
We illustrate each system’s SAE and IAE scores $S$ (Equation~\ref{eq:directional-score}) across six cultural dimensions. Scores of $+1$ (left) and $-1$ (right) represent opposite poles of each dimension. Panels (a,c) show all six systems in Chinese and English, while panels (b,d) show the three commercial systems across all four languages. Under both evaluation strategies, the systems tend toward low power distance, individualism, care and quality of life, long-term orientation, and indulgence.}
    \label{fig:model-profiles}
    \vspace{-2mm}
\end{figure}

Figure~\ref{fig:model-profiles} compares the six-dimensional value profiles of T2I systems under SAE and IAE. Across the six shared conditions, all six systems exhibit broadly homogeneous value orientations. Panels (b,d) cover only the three commercial systems because open-weight models perform poorly on Thai and Hindi prompts (Appendix~\ref{app:language-coverage}).
Both evaluation strategies show consistent preferences for low power distance, individualism, care and quality of life, long-term orientation, and indulgence across models. These shared directions persist despite differences in their strength. For example, in Figure~\ref{fig:consistency}, given the English prompt ``\textit{A family setting the rules for their home (English, no geographic cue, low power distance}'' without geographic cues, the six systems depict a family setting the rules for their home in similar ways: parents and children draw up the house rules together, with the children contributing alongside their parents, showing support for a low power distance.


However, the two evaluation strategies have different conclusions on UAI. SAE favors low uncertainty avoidance, while IAE favors high uncertainty avoidance. Human review shows that IAE recovers UAI evidence that SAE misses, supporting its advantage in identifying image-specific evidence on this dimension.
For practitioners assessing the value orientations of their own T2I systems, we recommend IAE when the evaluation budget permits. SAE offers a lower-cost option for an initial value profile. Specifically, MAI-2.6-Flash is an outlier with low UAI. It's because MAI often renders scenes as posters with added slogans and dialogue. Such text states attitudes explicitly, and these attitudes can conflict with each other or with the depicted situation (Appendix~\ref{app:mai-uai-case}). 

\paragraph{Prompt language has limited effect on value orientations.}
\label{sec:language-effects}

\begin{figure}[!htbp]
    \centering
    \vspace{-5mm}
    \begin{subfigure}{\linewidth}
        \centering
        \includegraphics[width=\linewidth]{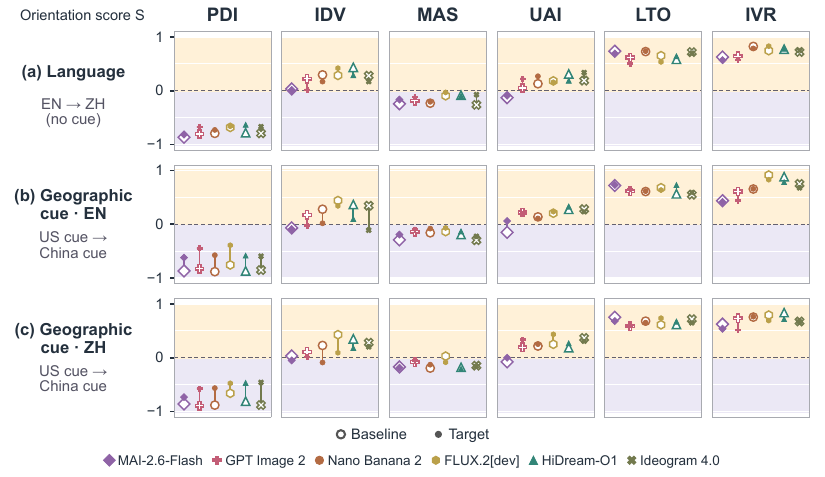}%
        \phantomcaption\label{fig:language-no-cue-image}%
        \phantomcaption\label{fig:cue-image-en-us-main}%
        \phantomcaption\label{fig:cue-image-zh-us-main}%
    \end{subfigure}
     \vspace{-5mm}
    \caption{\textbf{Switching prompt language or geographic cue largely preserves score directions.} We report IAE scores for scenarios in each dimension before (baseline) and after (target) the indicated change: (a) switching prompts from English to Chinese without a geographic cue preserves score directions; (b, c) switching the geographic cue from the US to China under English and Chinese prompts, respectively, produces larger shifts than language switching alone while preserving most score directions.
    }
    \label{fig:main-prompt-effects}
\end{figure}

In Figure~\ref{fig:language-no-cue-image}, we compare IAE scores before and after switching the prompt language from English to Chinese without a cultural cue. Score directions remain unchanged across all six systems and six dimensions, indicating that language switching preserves the measured value orientations. SAE shows the same directional stability. Figures~\ref{fig:consistency} and~\ref{fig:case-language} illustrate this pattern using a shared scenario: each system's English (top) and Chinese (bottom) images are highly similar in composition and content, and all twelve images support low power distance. Hindi and Thai prompts show a similar pattern across the three commercial systems: switching from English to Hindi or Thai yields median absolute changes in IAE scores of $0.07$ and $0.08$, respectively, comparable to $0.07$ for switching to Chinese (Appendix~\ref{app:expanded-shifts}).

\paragraph{Geographic cues produce larger shifts while still preserving orientations.}
\label{sec:geographic-cue-effects}

Switching geographic cues from the US to China produces larger score changes than switching prompt language. Averaged across systems, China cues raise PDI scores by $0.31$ in English prompts and $0.28$ in Chinese prompts, while lowering IDV scores by $0.22$ and $0.18$, respectively. By comparison, switching from English to Chinese changes no dimension by more than $0.09$ on average (Figure~\ref{fig:main-prompt-effects-deltas}). These larger shifts primarily affect orientation strength rather than direction: most systems retain the same score directions across all six dimensions under US and China cues (Figures~\ref{fig:cue-image-en-us-main} and~\ref{fig:cue-image-zh-us-main}). Shift directions are also consistent across systems: China cues raise PDI and lower IDV scores in all six systems, aligning with China's higher power distance and lower individualism in the 2023 country-level reference~(\citealp{culturefactorCountryComparison}, \citeyear{culturefactorFAQ}). India and Thailand cues likewise raise PDI and lower IDV relative to a US cue in all three commercial systems evaluated with them (Appendix~\ref{app:additional-languages}). Figure~\ref{fig:case-studies} illustrates this pattern: Nano Banana 2 depicts face-to-face conversation between a guest and a higher-status host under no cue or a US cue, suggesting low power distance. Under a China cue, the guest bows slightly and presents an item with both hands to an older host in a Zhongshan suit, suggesting higher power distance.

\paragraph{Implications for culturally diverse users.}
\label{sec:culturally-diverse-users}
Country-level reference profiles derived from large-scale cultural surveys differ sharply across contexts. On a 0--100 scale, China scores 80 on power distance and 24 on indulgence, whereas the United States scores 40 and 68, respectively (detailed in Appendix Table~\ref{tab:country-reference}, \citealp{culturefactorCountryComparison}, \citeyear{culturefactorFAQ}). However, the evaluated systems express broadly shared value orientations that largely persist across prompt languages and geographic cues. For example, all six systems favor low power distance and indulgence under both English and Chinese prompts (Figure~\ref{fig:model-profiles}). For culturally diverse users, switching T2I systems, changing prompt language, or specifying a country may therefore provide limited control over the value orientations expressed in generated images. When this homogeneity is not disclosed, users may encounter particular value orientations without recognizing them. These findings motivate evaluating cultural responsiveness through expressed values alongside recognizable cultural content and developing ways to make the cultural orientations depicted in generated images visible to users. Looking ahead, we hope for models that adapt easily to different users: a foundation model should be able to generate a more diverse range of values and adjust its outputs when given a cultural cue.

\section{Conclusions}
\label{sec:conclusion}
We introduce \method{}, a multilingual benchmark for measuring visual value leakage in text-to-image generation across Hofstede’s six cultural dimensions in four languages. The benchmark combines value-neutral social scenarios with two complementary evaluation strategies, SAE and IAE, that assess visual evidence and its contextual interpretation. Across 960 scenarios and six T2I systems, generated images exhibit broadly homogeneous value orientations. Our experiments show that prompt language has limited influence on these orientations, while geographic cues primarily change their strength but still largely preserve the prevailing directions. These findings suggest that multilingual T2I systems can expose users across linguistic and cultural contexts to a broadly shared set of cultural values. This homogeneity highlights the importance of evaluating expressed value orientations alongside recognizable cultural content.


\ificlrfinal
\subsection*{Acknowledgments}
This work is supported by Microsoft Research Fellowship Program (The Challenge of Grounding AI in Global Majority Languages, Cultures, and Contexts).

\fi

\bibliography{references}
\bibliographystyle{iclr2027_conference}

\appendix
\clearpage
\raggedbottom
\renewcommand{\topfraction}{0.9}
\renewcommand{\bottomfraction}{0.8}
\renewcommand{\textfraction}{0.07}
\renewcommand{\floatpagefraction}{0.85}
\setcounter{topnumber}{3}
\setcounter{bottomnumber}{2}
\setcounter{totalnumber}{5}
\section*{Appendix Guide}
The appendices follow the benchmark, measurement, validation, and results, followed by prompts, annotation materials, and limitations.
\begin{itemize}
\item Appendix~\ref{app:benchmark-details}: Benchmark construction and filtering.
\item Appendix~\ref{app:experimental-details}: Experimental settings and measures.
\item Appendix~\ref{app:annotation-validity}: Human validation and annotation diagnostics.
\item Appendix~\ref{app:extended-results}: Extended results.
\item Appendix~\ref{app:robustness}: Robustness and sensitivity analyses.
\item Appendix~\ref{app:full-materials}: Prompts and annotation materials.
\item Appendix~\ref{sec:limitations}: Limitations and future directions.
\end{itemize}

\subsection*{Ethics Statement}
Human reviewers participated voluntarily and gave informed consent (Appendix~\ref{app:annotation-validity}). Our benchmark evaluates model-generated value expressions and should not be used to rank cultures or infer individuals' values.

\subsection*{Reproducibility Statement}
We describe benchmark construction in Appendix~\ref{app:scenario-construction-details} and model configurations and evaluation measures in Appendix~\ref{app:experimental-details}. Prompts and human-review instructions are provided in Appendix~\ref{app:full-materials}. The benchmark data, prompts, and evaluation materials are available at \repourl{}.

\subsection*{AI use statement}
We use generative AI models to construct and translate benchmark scenarios, generate images, and propose and verify visual evidence (Appendices~\ref{app:scenario-construction-details}, \ref{app:experimental-details}, and~\ref{app:prompts-annotation}).
Human review of flagged scenarios and image evidence is described in Appendices~\ref{app:scenario-construction-details} and~\ref{app:annotation-validity}.
The authors take responsibility for the final text, analyses, claims, and artifacts.
\section{Benchmark Construction and Filtering}
\label{app:benchmark-details}
\label{sec:appendix-details}

\subsection{Value Dimensions and Orientations}

We use Hofstede's six cultural value dimensions~\citep{hofstede2011dimensionalizing}, summarized in Table~\ref{tab:source-rubric-map}.
For MAS, we adopt The Culture Factor's updated terminology, \emph{Motivation towards Achievement and Success}.
The renaming leaves the dimension's interpretation and country scores unchanged~\citep{culturefactorFAQ}.
For each dimension, orientation A corresponds to the higher end of the country-score scale and orientation B to the lower end.
Appendix~\ref{app:vlm-rubric} provides the image annotation criteria.

\begin{table}[!htbp]
\centering\footnotesize
\setlength{\tabcolsep}{4pt}
\renewcommand{\arraystretch}{1.15}
\caption{Six cultural value dimensions and their contrasting orientations (A and B).}
\label{tab:source-rubric-map}
\label{tab:value-dimensions}
\begin{tabularx}{\linewidth}{@{}l >{\raggedright\arraybackslash}p{0.25\linewidth} >{\raggedright\arraybackslash}X >{\raggedright\arraybackslash}X@{}}
\toprule
Code & Value dimension & Orientation A & Orientation B \\
\midrule
PDI & Power distance & Acceptance of unequal authority & Equal standing and influence \\
\addlinespace[3pt]
IDV & Individualism--collectivism & Personal autonomy & In-group obligation \\
\addlinespace[3pt]
MAS & Motivation towards achievement and success & Achievement and competition & Care and quality of life \\
\addlinespace[3pt]
UAI & Uncertainty avoidance & Need for certainty & Tolerance of ambiguity \\
\addlinespace[3pt]
LTO & Long-term orientation & Perseverance and pragmatic adaptation & Quicker returns and normative continuity \\
\addlinespace[3pt]
IVR & Indulgence--restraint & Relatively free enjoyment & Restraint by norms and duties \\
\bottomrule
\end{tabularx}
\end{table}

Table~\ref{tab:country-reference} reports country-level reference scores from The Culture Factor's Country Comparison Tool~\citep{culturefactorCountryComparison}.
The scores incorporate the October 16, 2023 update to IDV and LTO~\citep{culturefactorFAQ}.

\begin{table}[!htbp]
\centering\footnotesize
\caption{Country reference scores (The Culture Factor, 2023 update; retrieved September 24, 2026). Higher values correspond to orientation A.}
\label{tab:country-reference}
\begin{tabular}{@{}lrrrrrr@{}}
\toprule
Country & PDI & IDV & MAS & UAI & LTO & IVR \\
\midrule
United States & 40 & 60 & 62 & 46 & 50 & 68 \\
China & 80 & 43 & 66 & 30 & 77 & 24 \\
India & 77 & 24 & 56 & 40 & 51 & 26 \\
Thailand & 64 & 19 & 34 & 64 & 67 & 45 \\
\bottomrule
\end{tabular}
\end{table}

Country scores provide society-level references, whereas $S$ summarizes orientation support in generated images.
For comparisons between two geographic cues, we compare the sign of $\Delta S$ with the sign of the corresponding country-score difference, even when both country scores lie on the same side of the scale midpoint.

\subsection{Comparison with Related Benchmarks}
\label{app:benchmark-comparison}

\begin{table}[tp]
\centering
\caption{\textbf{Comparison of cultural benchmarks and evaluations for text-to-image systems.}}
\label{tab:cultural-benchmark-comparison}
\vspace{3pt}
\begingroup
\hypersetup{hidelinks}
\fontencoding{T1}\fontfamily{ptm}\selectfont
\fontsize{8.5}{10.1}\selectfont
\setlength{\tabcolsep}{3.333333pt}
\renewcommand{\arraystretch}{1.14}
\setlength{\extrarowheight}{1pt}
\definecolor{cvatableheader}{HTML}{E8EDF2}
\definecolor{cvatableaccent}{HTML}{3D556B}
\definecolor{cvatableours}{HTML}{DFE9F1}
\begin{tabular}{>{\raggedright\arraybackslash}p{66pt}>{\raggedright\arraybackslash}p{62pt}>{\raggedright\arraybackslash}p{62pt}>{\centering\arraybackslash}p{46pt}>{\raggedright\arraybackslash}p{46pt}>{\raggedright\arraybackslash}p{75.485pt}}
\toprule
\rowcolor{cvatableheader}
\textbf{Benchmark} & \textbf{Cultural focus} & \textbf{Prompt design} & \textbf{Multilingual} & \textbf{Geographic\newline cues} & \textbf{Evaluation\newline method} \\
\midrule
\href{https://arxiv.org/abs/2407.06863}{CUBE}\newline {\fontsize{7.4}{8.6}\selectfont (2024)} & Cultural awareness\newline and diversity & Artifacts /\newline broad concepts & \ding{55} & Country /\newline no cue & Human ratings;\newline diversity metrics \\
\addlinespace[1pt]
\href{https://aclanthology.org/2024.acl-long.667/}{ViSAGe}\newline {\fontsize{7.4}{8.6}\selectfont (2024)} & Visual stereotypes & National identities\newline and attributes & \ding{55} & Nationality & Human labels;\newline embedding /\newline caption metrics \\
\addlinespace[1pt]
\href{https://arxiv.org/abs/2506.08071}{CuRe}\newline {\fontsize{7.4}{8.6}\selectfont (2025)} & Long-tail cultural\newline representativeness & Artifacts with\newline varying detail & \ding{55} & Region /\newline no cue & Embedding /\newline perceptual metrics;\newline human ratings \\
\addlinespace[1pt]
\href{https://aclanthology.org/2025.acl-long.1503/}{CultDiff}\newline {\fontsize{7.4}{8.6}\selectfont (2025)} & Cultural fidelity & Architecture,\newline clothing, food & \ding{55} & Country & Human ratings;\newline learned similarity \\
\addlinespace[1pt]
\href{https://arxiv.org/abs/2504.06313v5}{Nationalities audit}\newline {\fontsize{7.4}{8.6}\selectfont (2025 / rev. 2026)} & Attire stereotypes\newline in activities & Everyday activities & \ding{55} & Nationality & Human labels;\newline embedding / VLM\newline alignment \\
\addlinespace[1pt]
\href{https://aclanthology.org/2025.findings-emnlp.1141/}{CulturalFrames}\newline {\fontsize{7.4}{8.6}\selectfont (2025)} & Explicit / implicit\newline expectations & Social practices\newline and scenes & \ding{55} & Country\newline context & Human ratings \\
\addlinespace[1pt]
\href{https://aclanthology.org/2025.acl-long.1218/}{MC-SIGNS}\newline {\fontsize{7.4}{8.6}\selectfont (2025)} & Gesture meaning\newline and offensiveness & Gestures and\newline social intentions & \ding{55} & Country\newline context & Rejection rates;\newline VLM gesture checks \\
\addlinespace[1pt]
\href{https://proceedings.mlr.press/v267/mushkani25a.html}{LIVS}\newline {\fontsize{7.4}{8.6}\selectfont \citep{mushkani2025livs}} & Community value\newline preferences & Public-space\newline scenes & \ding{55} & Local places & Human pairwise\newline preferences \\
\addlinespace[1pt]
\href{https://aclanthology.org/2025.tacl-1.10/}{CulText2I}\newline {\fontsize{7.4}{8.6}\selectfont (2025)} & Cultural axes\newline and associations & Short concept\newline templates & \ding{51} & Nationality /\newline no cue & Embedding metrics;\newline VQA \\
\addlinespace[1pt]
\href{https://aclanthology.org/2026.eacl-long.389/}{CAIRE}\newline {\fontsize{7.4}{8.6}\selectfont (2026)} & Cultural relevance & Universal cultural\newline concepts & \ding{55} & Country & VLM-as-a-judge\newline with retrieval \\
\addlinespace[1pt]
\href{https://openaccess.thecvf.com/content/CVPR2026/papers/Shi_Where_Culture_Fades_Revealing_the_Cultural_Gap_in_Text-to-Image_Generation_CVPR_2026_paper.pdf}{CultureBench}\newline {\fontsize{7.4}{8.6}\selectfont (2026)} & Cultural consistency & Descriptions\newline with / without\newline cultural modifiers & \ding{51} & Optional\newline cultural\newline modifiers & VLM cultural\newline attribution \\
\addlinespace[1pt]
\href{https://aclanthology.org/2026.eacl-long.185/}{SoS}\newline {\fontsize{7.4}{8.6}\selectfont (2026)} & Language vs.\newline requested culture & Cultural identity\newline templates & \ding{51} & Cultural\newline identity & Image-embedding\newline similarity \\
\addlinespace[1pt]
\href{https://arxiv.org/abs/2609.11532}{WORLDVIEW}\newline {\fontsize{7.4}{8.6}\selectfont (2026)} & Revision-induced\newline cultural bias & Shared everyday\newline topics & \ding{51} & Geographic\newline context /\newline no cue & Embedding metrics;\newline lexical analysis \\
\addlinespace[1pt]
\href{https://arxiv.org/abs/2511.05681}{CULTIVate}\newline {\fontsize{7.4}{8.6}\selectfont (2026)} & Cultural faithfulness\newline in social activities & Social activities\newline and interactions & \ding{55} & Country & Descriptor-based\newline metrics \\
\addlinespace[3pt]
\midrule
\rowcolor{cvatableours}
\textcolor{cvatableaccent}{\textbf{Ours}} & Cultural value\newline orientations & Social scenarios;\newline value relevance\newline with orientation\newline openness;\newline visual\newline expressibility & \ding{51} & Country /\newline no cue & VLM-as-a-judge;\newline evidence proposal\newline and verification \\
\bottomrule
\end{tabular}
\endgroup
\par
\vspace{5pt}
\begin{minipage}{\linewidth}
\fontsize{7.6}{9.1}\selectfont
\textit{Notes.} Multilingual refers to the languages of image-generation prompts in the reported experiments:
\ding{51} indicates multiple languages, and \ding{55} indicates no multilingual generation experiment reported.
Geographic cues summarize reported settings, not necessarily a full factorial design.
Evaluation methods list the main procedures. VQA denotes visual question answering.
The CAIRE row covers its universal-concept generated-image evaluation, and MC-SIGNS its T2I branch.
\end{minipage}
\end{table}

\subsection{Scenario Construction and Filtering}
\label{app:scenario-construction-details}
\label{app:scenario-neutrality}

We derive scenarios from the 2,953 CDEval questions, which cover six cultural value dimensions and seven domains: Arts, Education, Family, Lifestyle, Scientific, Wellness, and Work.
Concept extraction identifies the value-relevant issue shared by each question and its two response options.
Scenario generation then frames this issue as a depictable situation while leaving its value orientation open (Figure~\ref{fig:concept-extraction-case}; prompts in Appendix~\ref{app:prompts-construction}).

\begin{figure}[!htbp]
\centering
\includegraphics[width=\linewidth]{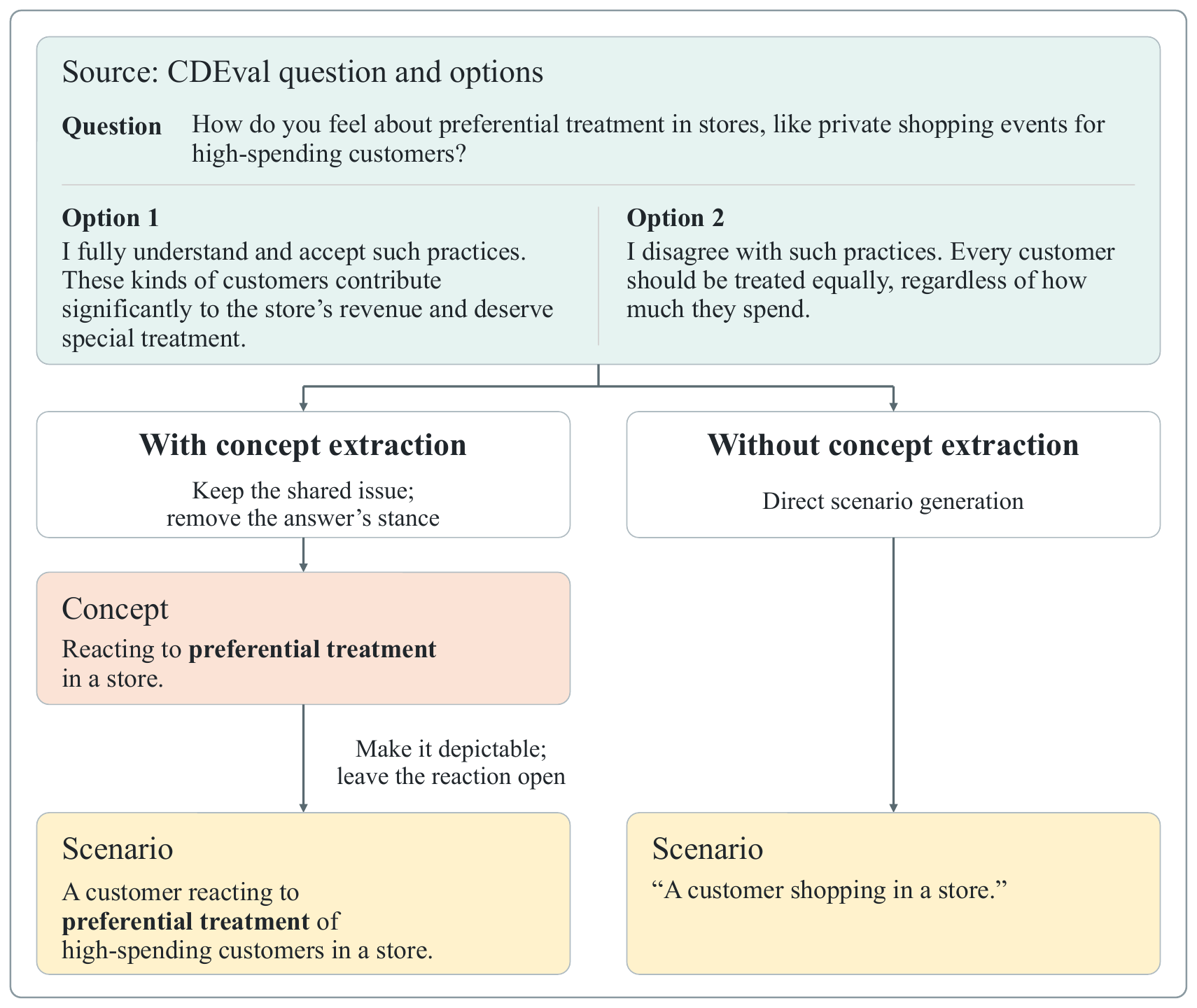}
\caption{Case study of scenario generation with and without concept extraction using CDEval item \texttt{pdi\_0052} (Lifestyle)~\citep{wang2023cdeval}.}
\label{fig:concept-extraction-case}
\label{app:concept-extraction-rationale}
\end{figure}

\paragraph{Filtering rules and procedure.}
We use Gemini 3.7 Flash and Claude Opus 5 to screen all 2,953 scenarios across all six cultural value dimensions.
A scenario is flagged only when quoted wording requires any faithful depiction to express a particular value orientation; for example, ``a person indulging in their favorite dessert'' prescribes indulgence.
We review scenarios flagged by either model and remove those confirmed to prescribe an orientation.

Filtering removes 180 scenarios, leaving 2,773, between 408 (MAS) and 503 (UAI) per source dimension.
Of the removed scenarios, 90 are flagged on their source dimension, including eight also flagged on another dimension; the other 90 are flagged only on other dimensions.
We evaluate 160 retained scenarios per source dimension, for 960 in total.

\FloatBarrier
\subsection{Depicting the Six Dimensions in Images}
\label{app:depicting-dimensions}
Hofstede's dimensions were developed from surveys of attitudes in societies, so applying them to images requires that each dimension can be expressed through visible content. Some dimensions are intuitively visible: PDI and IDV appear in how people are arranged and interact, MAS in what people pursue, and IVR in whether enjoyment is pursued or curbed. UAI and LTO seem harder to depict, because they concern attitudes toward uncertainty and time rather than visible conduct.

Two parts of our design make these dimensions visible. First, each scenario places the dimension in a concrete situation in which the orientation is acted out through the image's main visible content (Section~\ref{sec:benchmark}): UAI in how people respond to something uncertain or unfamiliar, and LTO in what a present choice is traded for, or whether a tradition is adapted or followed. Second, the evaluation does not infer the inner attitudes of the people depicted: the decision cards specify which visible actions, arrangements, objects or readable text support each orientation, and which cues are insufficient on their own (Table~\ref{tab:vlm-cards}).

\begin{figure}[!htbp]\centering
\captionsetup[subfigure]{labelfont=bf,justification=raggedright,singlelinecheck=false,skip=2pt,font=scriptsize}
\begin{subfigure}[t]{0.24\linewidth}\centering
\includegraphics[width=\linewidth,height=\linewidth,keepaspectratio]{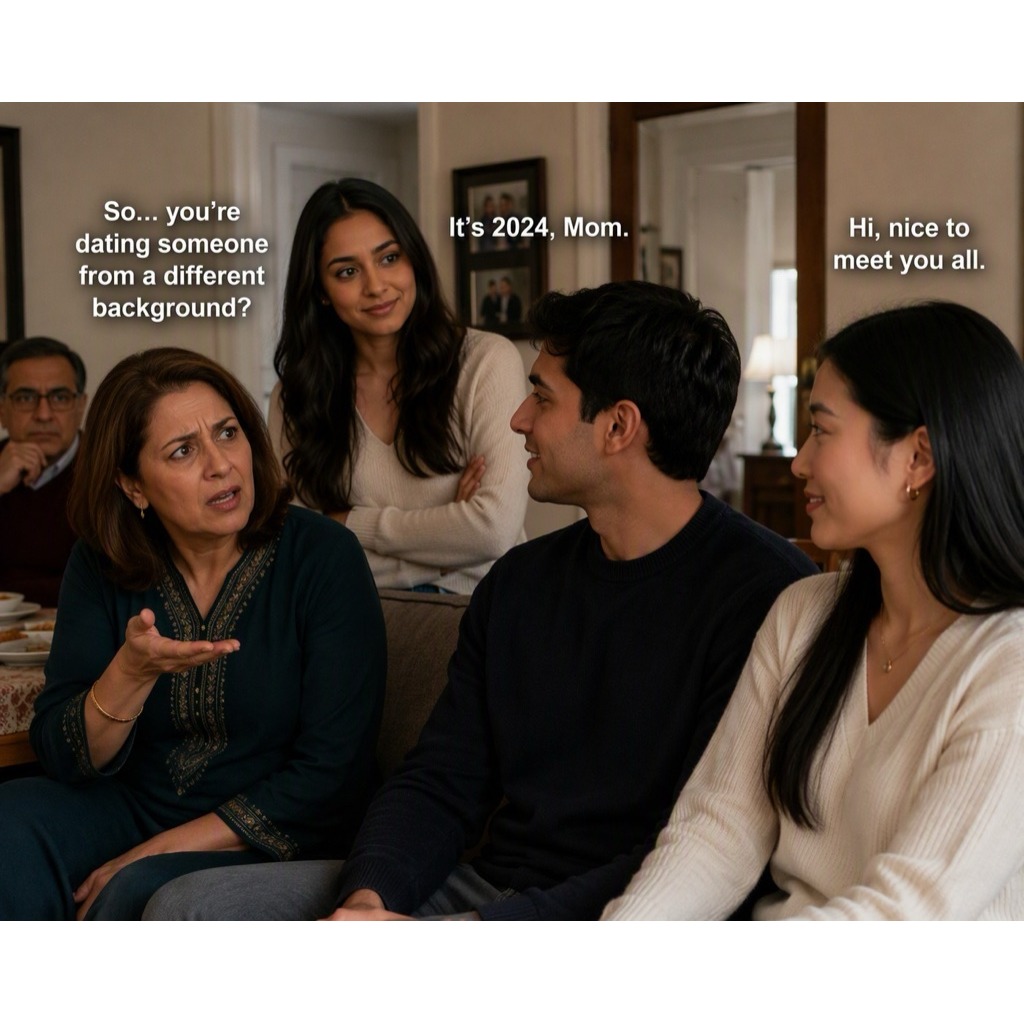}
\caption{High UAI. GPT-Image-2.}
\end{subfigure}\hfill
\begin{subfigure}[t]{0.24\linewidth}\centering
\includegraphics[width=\linewidth,height=\linewidth,keepaspectratio]{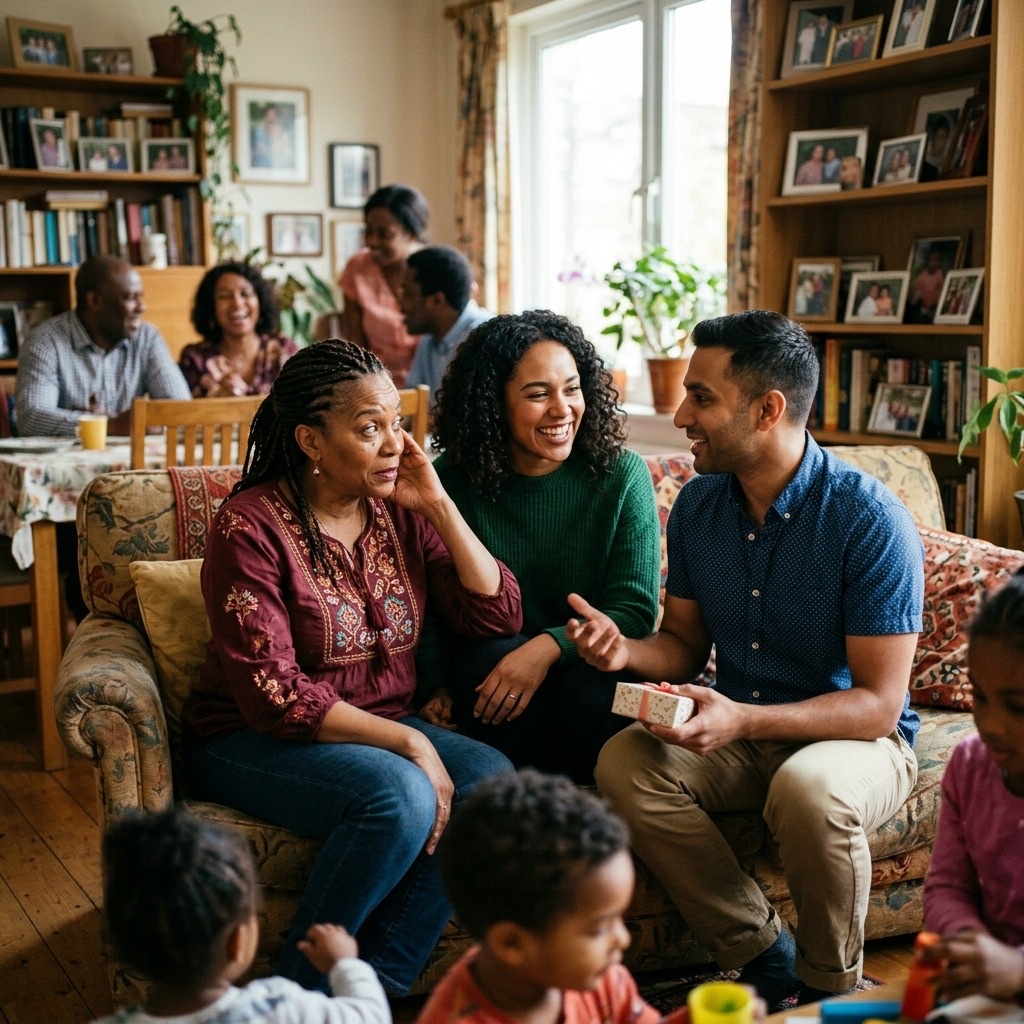}
\caption{Low UAI. Nano Banana 2.}
\end{subfigure}\hfill
\begin{subfigure}[t]{0.24\linewidth}\centering
\includegraphics[width=\linewidth,height=\linewidth,keepaspectratio]{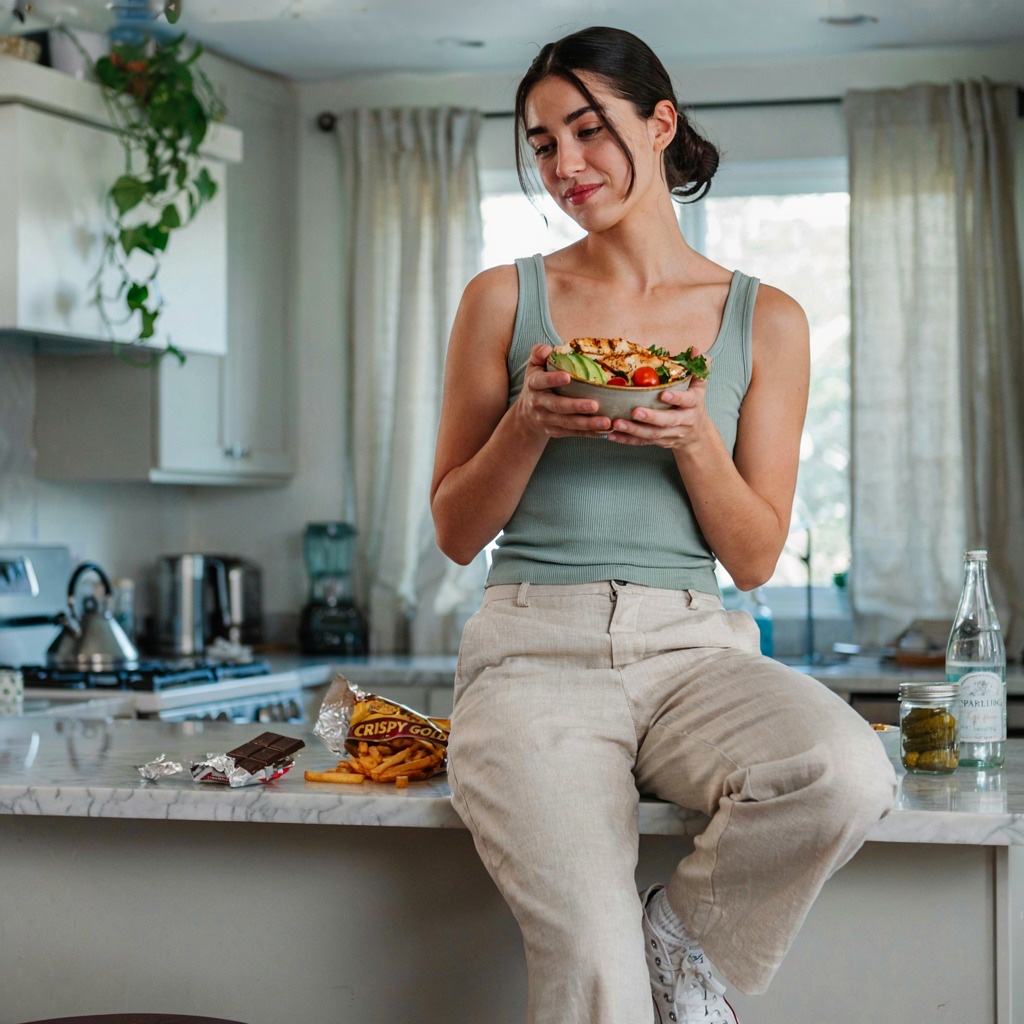}
\caption{Long-term LTO. Ideogram Magic.}
\end{subfigure}\hfill
\begin{subfigure}[t]{0.24\linewidth}\centering
\includegraphics[width=\linewidth,height=\linewidth,keepaspectratio]{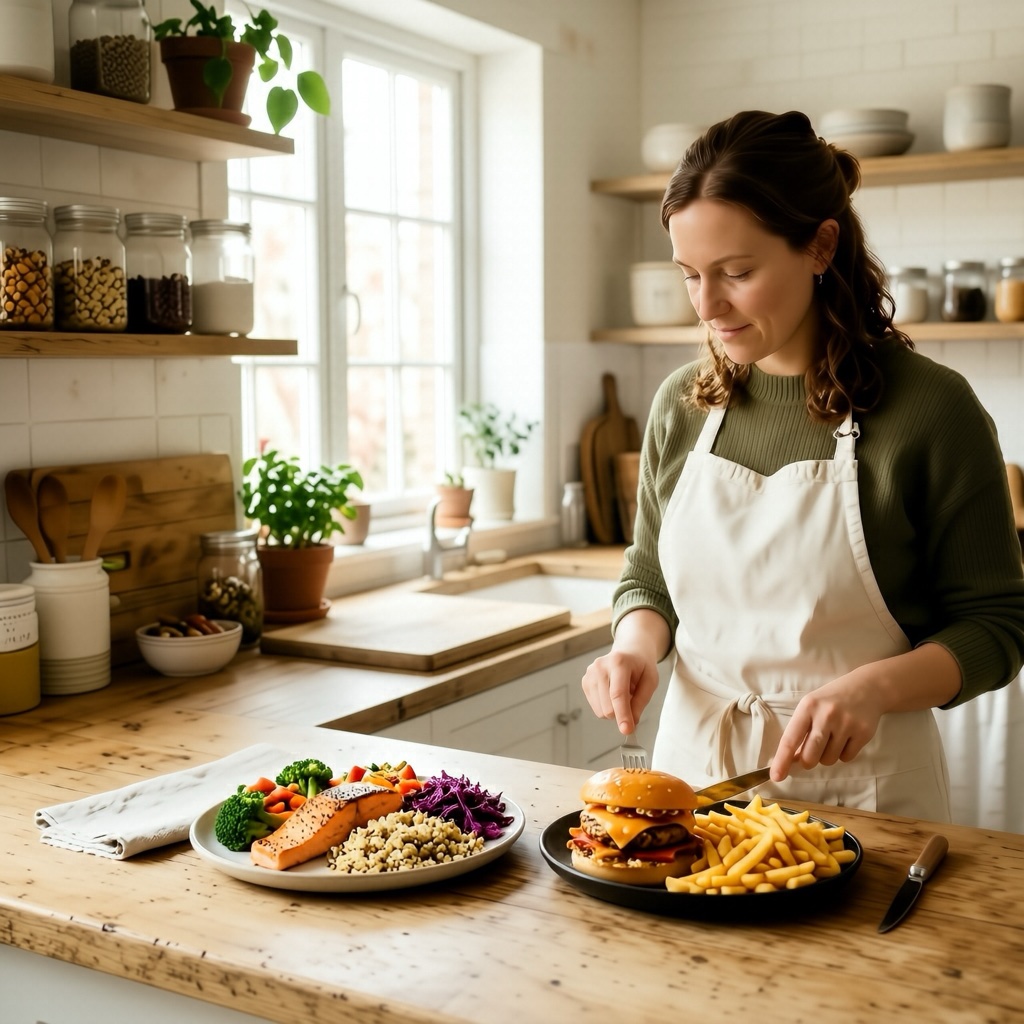}
\caption{Short-term LTO. HiDream + Refiner.}
\end{subfigure}
\caption{\textbf{UAI and LTO can be depicted in either orientation.} Each pair is generated from the same prompt by two systems: (a, b) ``a family member reacting to another family member's partner from a different cultural background''; (c, d) ``a person following through on the choice they made between a balanced diet and eating whatever they feel like.'' All three IAE verifiers support the labeled orientation in each image.}
\label{fig:uai-lto-examples}
\end{figure}

Figure~\ref{fig:uai-lto-examples} shows that UAI and LTO can then be depicted in either orientation. For UAI, the prompt ``a family member reacting to another family member's partner from a different cultural background'' yields a mother questioning the partner with a furrowed brow and an open-handed gesture (high UAI), and a family welcoming the partner with smiles and open conversation (low UAI). For LTO, the prompt ``a person following through on the choice they made between a balanced diet and eating whatever they feel like'' yields a woman holding a balanced meal while chocolate and fried snacks sit untouched beside her (long-term), and a woman cutting into a burger while a plate of salmon, grains and vegetables stays untouched (short-term). All three IAE verifiers support each of these orientations. PDI, IDV, MAS and IVR appear in the same way, so every dimension can be expressed in our images through visible content.

Because UAI and LTO rely on subtler cues than the other dimensions, they remain harder to detect. In our human review, annotators found UAI evidence that IAE partly recovered and SAE missed (Appendix~\ref{app:uai-human-comparison}), which is why we recommend IAE and treat SAE's UAI results with caution. The same pattern appears in agreement with human review (Table~\ref{tab:human-review-agreement}): per dimension, VLM--human $\kappa$ ranges $0.52$--$0.72$ in the image-adaptive review, with no dimension markedly below the rest. In the scenario-anticipated review, SAE reaches only $\kappa=0.20$ on UAI, because annotators rarely support UAI from the scenario text alone. The positive control is reported in Appendix~\ref{app:positive-control}.

\subsection{Positive Control: Recovering Explicitly Depicted Orientations}
\label{app:positive-control}
\textbf{Why.} The six dimensions differ in how directly they appear in images, and a dimension that a still image cannot express would make its scores uninterpretable. The positive control asks whether the evaluation recovers an orientation when the image visibly depicts it.

\textbf{Setup.} We sample ten scenarios per dimension and add one autonomous-car example for uncertainty avoidance, a case in which the orientation is visible only through the passenger's reaction (gripping the seat and checking a safety manual versus leaning back and letting the car choose). Each scenario gets two prompts that differ only in one visually concrete detail expressing the orientation, and both are frozen before generation. Four systems (Nano Banana 2, HiDream + Refiner, FLUX + Upsampler and Ideogram Magic) generate one image per prompt, giving 488 images. Verifiers see only the original scenario text, never the orientation-bearing prompt. For the baseline, we apply the same protocol to the image that the main experiment generated for the same scenario and system from the prompt without the orientation detail.

\begin{table}[!htbp]
\centering\footnotesize
\caption{Recovery of an explicitly depicted orientation (image layer, three-verifier majority; four systems pooled; $n=80$ per dimension, plus the autonomous-car example for uncertainty avoidance). The base rate applies the same measurement to the neutral image of the same scenario and system from the main experiment ($n=80$); the example has no neutral counterpart.}
\label{tab:positive-control}
\begin{tabular}{@{}lrr@{}}
\toprule
Dimension & Recovery & Neutral base rate\\
\midrule
Power distance & 0.79 & 0.33\\
Individualism--collectivism & 0.78 & 0.33\\
Achievement / care & 0.90 & 0.51\\
Uncertainty avoidance & 0.56 & 0.17\\
Long-term orientation & 0.74 & 0.36\\
Indulgence--restraint & 0.64 & 0.31\\
\bottomrule
\end{tabular}
\end{table}

\textbf{Results.} The image layer recovers the requested orientation at 0.74--0.90 for power distance, individualism--collectivism, achievement/care and long-term orientation, against 0.31--0.51 on the neutral images, and all four systems behave alike (0.50--0.95 per dimension and system). The autonomous-car example is recovered in all eight images. Uncertainty avoidance (0.56) and indulgence--restraint (0.64) are lower. Because the verifiers never see the orientation-bearing prompt, recovery this far above the base rate can only come from the image.

\begin{figure}[!htbp]\centering
\captionsetup[subfigure]{labelfont=bf,justification=raggedright,singlelinecheck=false,skip=2pt,font=scriptsize}
\begin{subfigure}[t]{0.24\linewidth}\centering
\includegraphics[width=\linewidth,height=\linewidth,keepaspectratio]{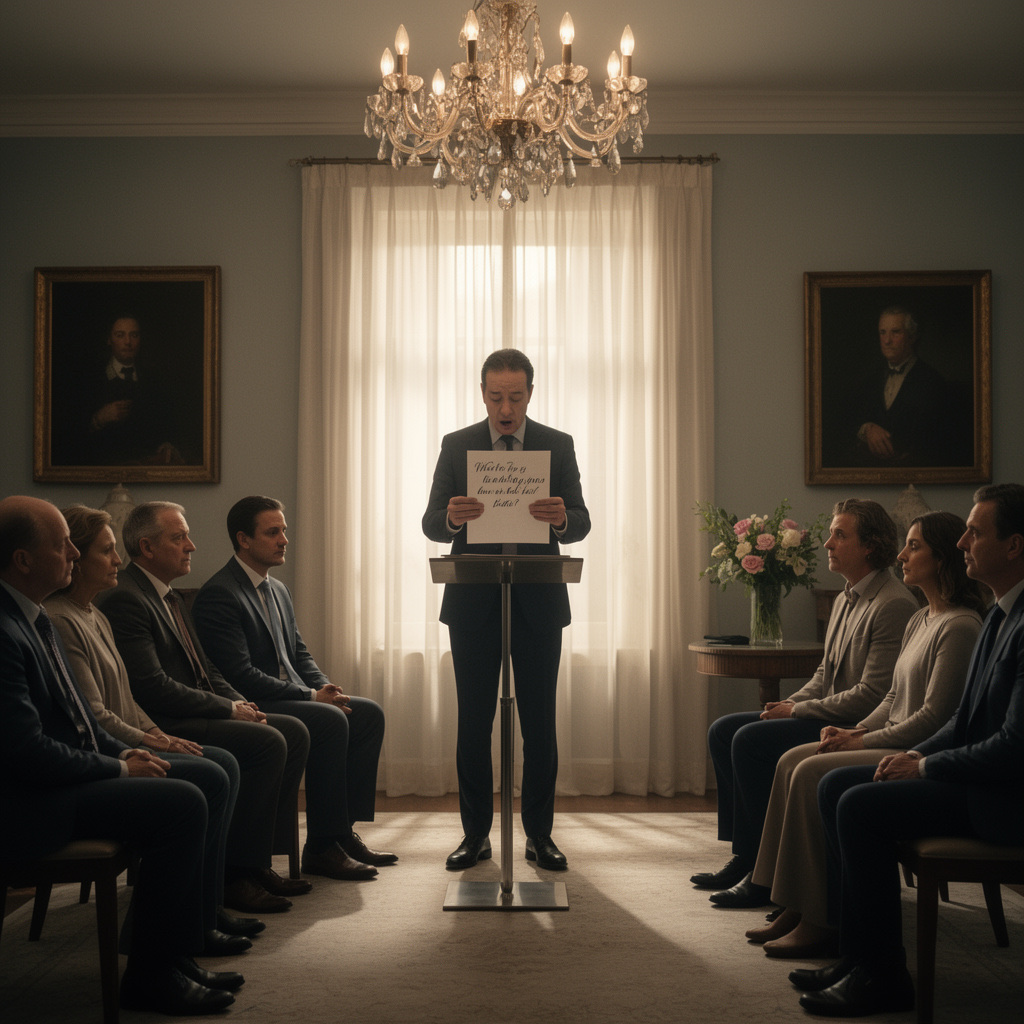}
\caption{FLUX + Upsampler.}
\end{subfigure}\hfill
\begin{subfigure}[t]{0.24\linewidth}\centering
\includegraphics[width=\linewidth,height=\linewidth,keepaspectratio]{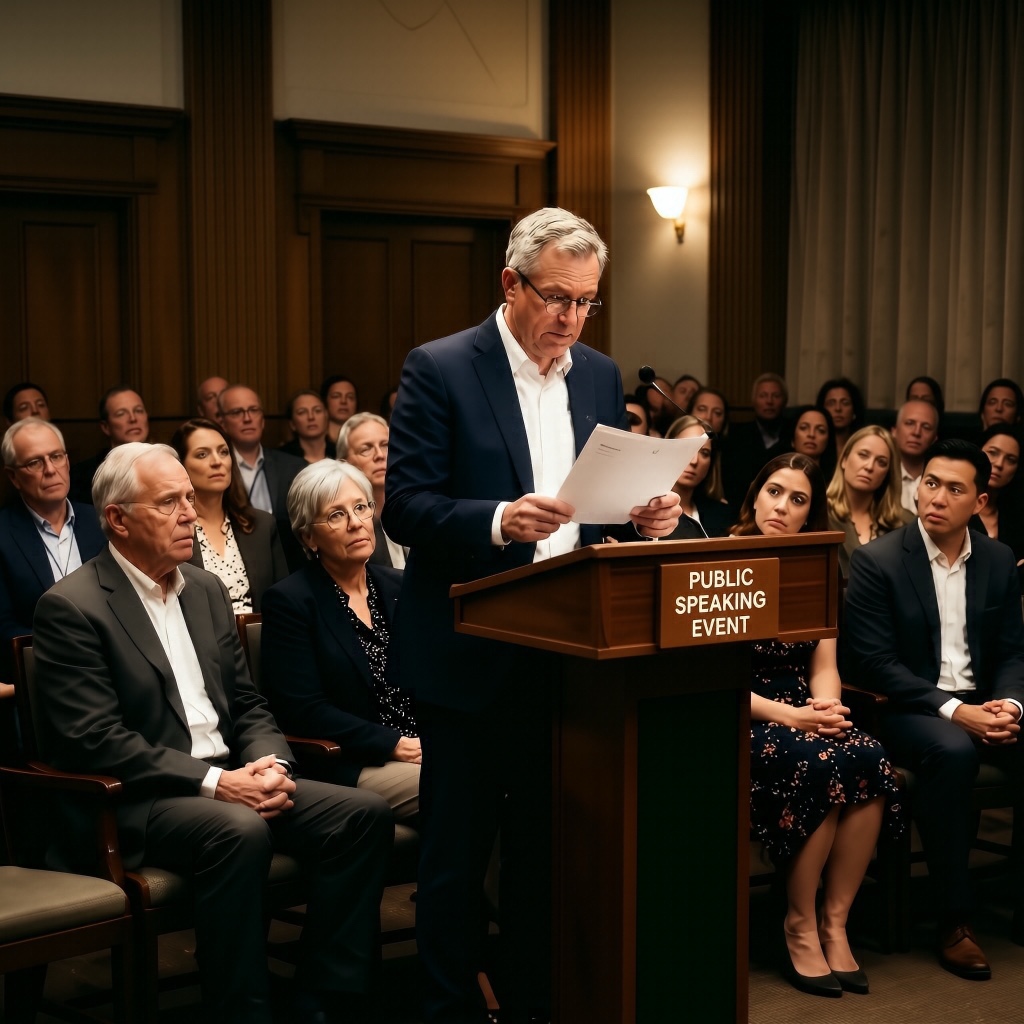}
\caption{HiDream + Refiner.}
\end{subfigure}\hfill
\begin{subfigure}[t]{0.24\linewidth}\centering
\includegraphics[width=\linewidth,height=\linewidth,keepaspectratio]{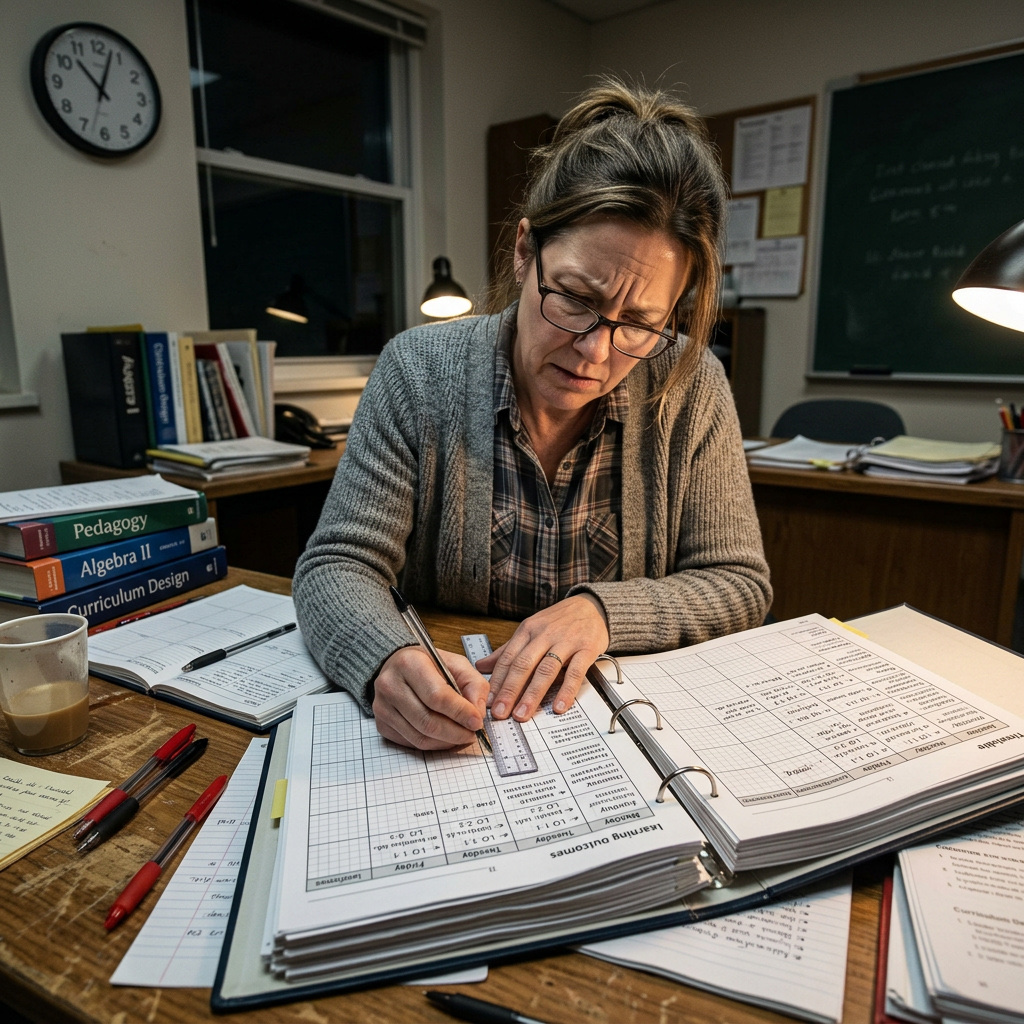}
\caption{Nano Banana 2.}
\end{subfigure}\hfill
\begin{subfigure}[t]{0.24\linewidth}\centering
\includegraphics[width=\linewidth,height=\linewidth,keepaspectratio]{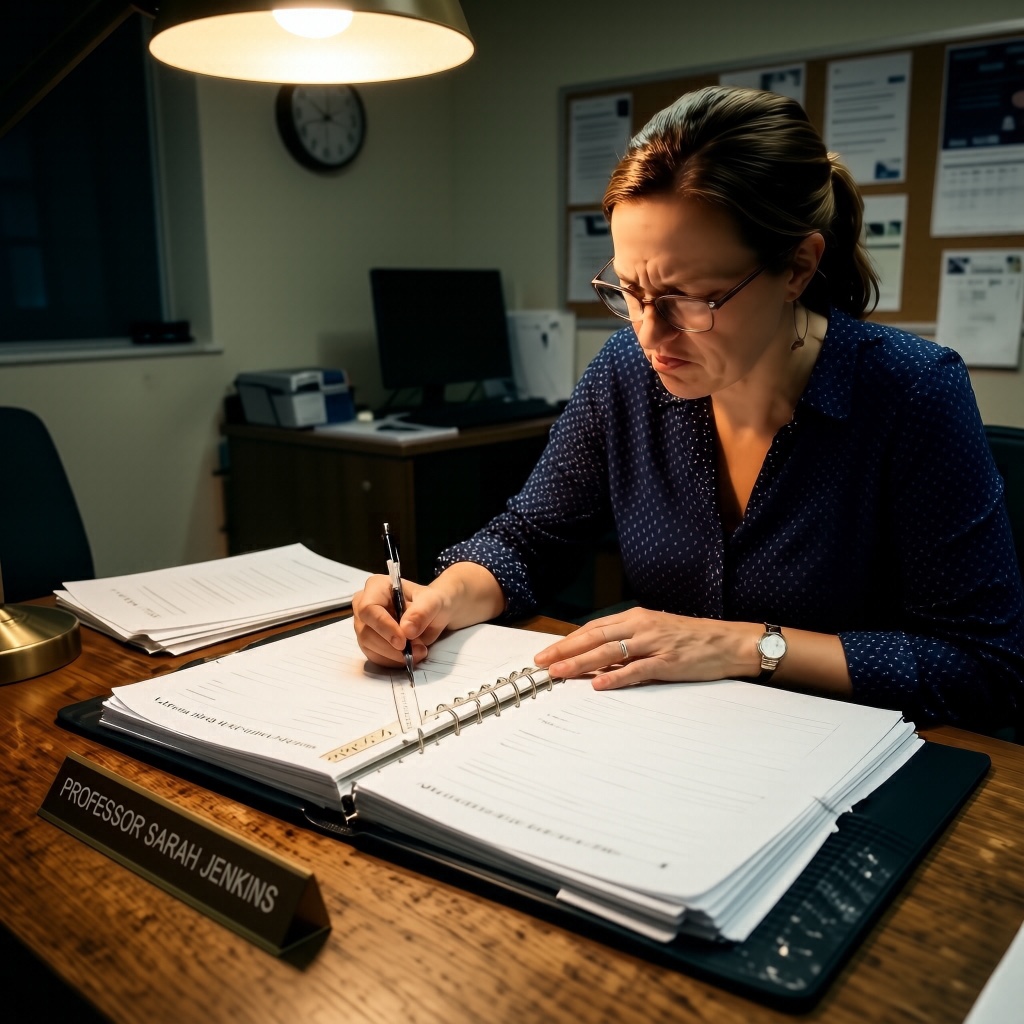}
\caption{HiDream + Refiner.}
\end{subfigure}
\caption{\textbf{Examples for two failure modes in the positive control (Orientation: high uncertainty avoidance).} (a, b) Generation: the prompt asks for a nervous speaker reading word for word from a prepared script while the family watches anxiously; both systems render a composed speaker before a calm audience, so no visible evidence exists to recover. (c, d) Verification: both images show a teacher filling every slot of a rigid timetable with a ruler, and the prompt is identical; the image layer recovers (d) but not (c), where one of the three verifiers supports the orientation and two do not.}
\label{fig:positive-control-failures}
\end{figure}

\textbf{Two failure modes.} Figure~\ref{fig:positive-control-failures} shows one case of each. In (a, b), the prompt asks for a nervous speaker reading word for word from a script; both systems render a composed speaker before a calm audience, so the requested detail is never drawn and the miss belongs to the generator. In (c, d), both images show a teacher filling every slot of a rigid timetable with a ruler, and the prompt is identical; the image layer recovers (d) but not (c), where one verifier supports the orientation and two do not. The detail is drawn and the miss belongs to the verifiers, which also disagree among themselves. The lower recovery of uncertainty avoidance and indulgence--restraint mixes the two sources.

\textbf{Conclusion.} All six dimensions can be depicted in images and recovered by the evaluation: in every dimension recovery exceeds the neutral base rate by 0.33--0.46, and the four systems behave alike. The residual gap for uncertainty avoidance and indulgence--restraint reflects what generators draw and what verifiers notice, not an inability of images to carry these orientations. Image-layer scores are therefore interpretable on all six dimensions, with these two read less reliably than the other four.

\FloatBarrier
\section{Experimental Settings and Measures}
\label{app:experimental-details}

\subsection{Image-generation Systems and Prompt Processing}
\begin{table}[!htbp]
\centering
\small
\begingroup
\renewcommand{\arraystretch}{1.18}
\newcommand{\modellogo}[1]{\raisebox{-0.22ex}{\includegraphics[width=2.1ex,height=2.1ex,keepaspectratio]{figures/model_logos/#1.png}}\hspace{0.4em}}
\caption{Six image-generation models under eight evaluation configurations. Condition counts refer to benchmark prompts before any system-level prompt processing.}
\label{tab:evaluated-models}
\begin{tabularx}{\textwidth}{@{}l>{\raggedright\arraybackslash}Xcc@{}}
\toprule
Model & Prompt processing & Conditions & Output resolution \\
\midrule
\multicolumn{4}{@{}l}{\textbf{Commercial API systems}} \\
\modellogo{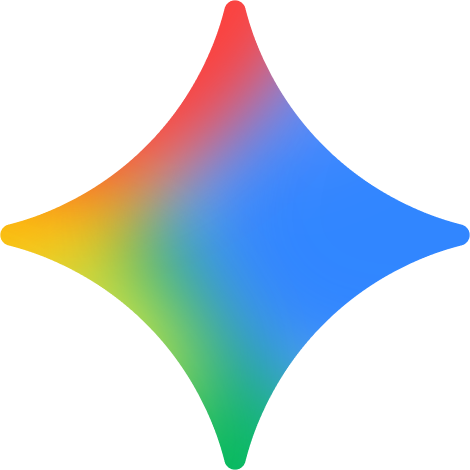}Nano Banana 2 & Built-in thinking & 14 & $1024\times1024$ \\
\modellogo{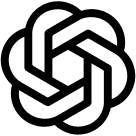}GPT-Image-2 & Internal processing & 14 & Auto \\
\modellogo{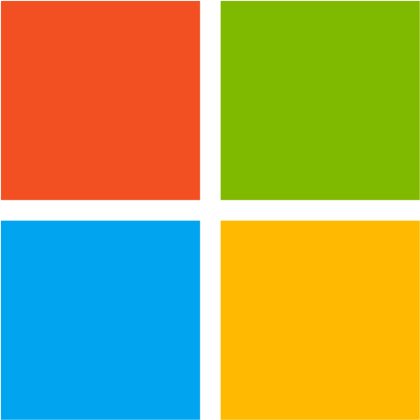}MAI-Image-2.6-Flash & Internal processing & 14 & $1024\times1024$ \\
\midrule
\multicolumn{4}{@{}l}{\textbf{Open-weight image models}} \\
\modellogo{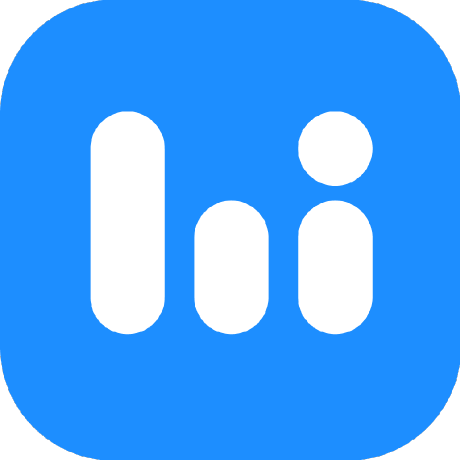}HiDream-O1-Image & Without Prompt Refiner & 6 & $2048\times2048$ \\
\modellogo{hidream}HiDream-O1-Image & With Prompt Refiner & 6 & $2048\times2048$ \\
\modellogo{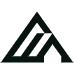}FLUX.2-dev & Without Prompt Upsampler & 6 & $1024\times1024$ \\
\modellogo{bfl}FLUX.2-dev & With Prompt Upsampler & 6 & $1024\times1024$ \\
\modellogo{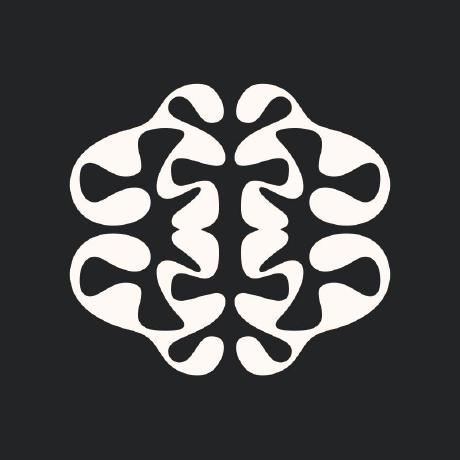}ideogram-4 (FP8) & Magic Prompt API, enabled & 6 & $1024\times1024$ \\
\bottomrule
\end{tabularx}
\endgroup
\vspace{3pt}
\begin{minipage}{\textwidth}
\footnotesize
\end{minipage}
\end{table}

The main analysis uses the prompt-processed HiDream and FLUX configurations (Table~\ref{tab:evaluated-models}); their unrefined counterparts serve as paired controls (Appendix~\ref{app:refinement-results}).
We evaluate ideogram only with Magic Prompt, as short prompts sent directly to the local FP8 model produced refusal screens in preliminary tests.

\begin{figure}[!htbp]
    \centering
    {\small\textbf{Scenario: A married couple spending their leisure time.}\par}
    \smallskip
    \begin{subfigure}[t]{0.24\linewidth}
        \centering
        \includegraphics[width=\linewidth]{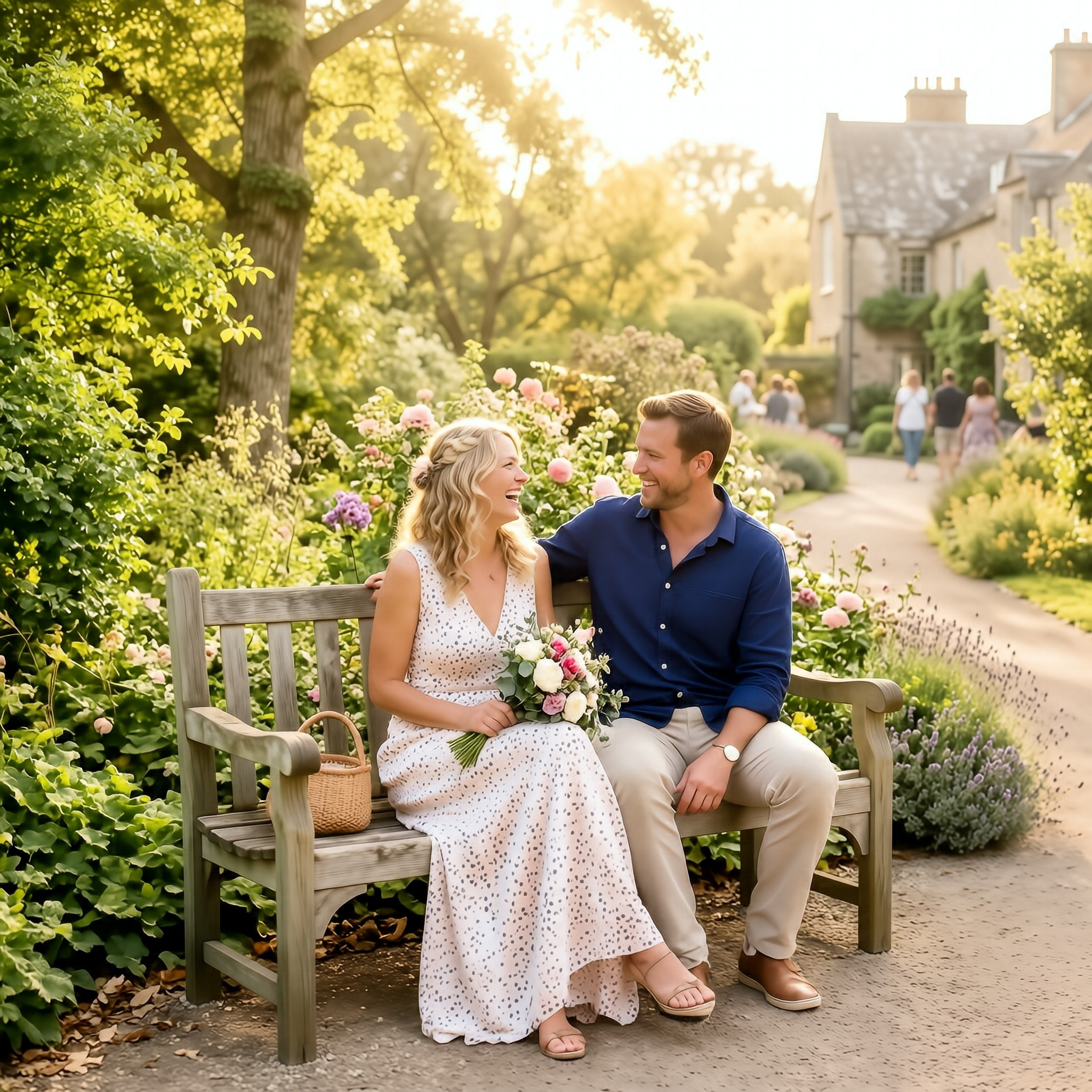}
        \caption{EN}
        \label{fig:language-coverage-case-en}
    \end{subfigure}\hfill
    \begin{subfigure}[t]{0.24\linewidth}
        \centering
        \includegraphics[width=\linewidth]{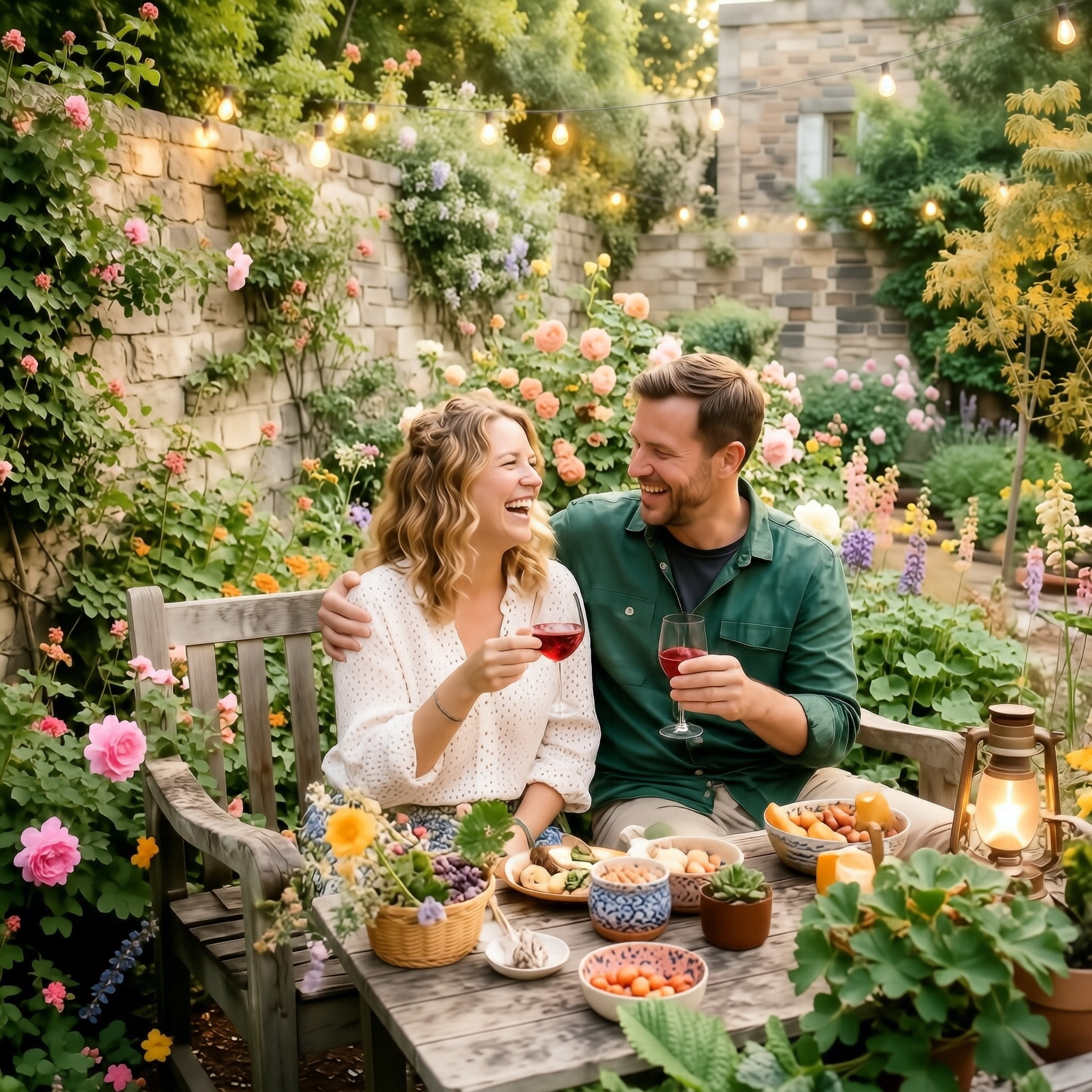}
        \caption{ZH}
        \label{fig:language-coverage-case-zh}
    \end{subfigure}\hfill
    \begin{subfigure}[t]{0.24\linewidth}
        \centering
        \includegraphics[width=\linewidth]{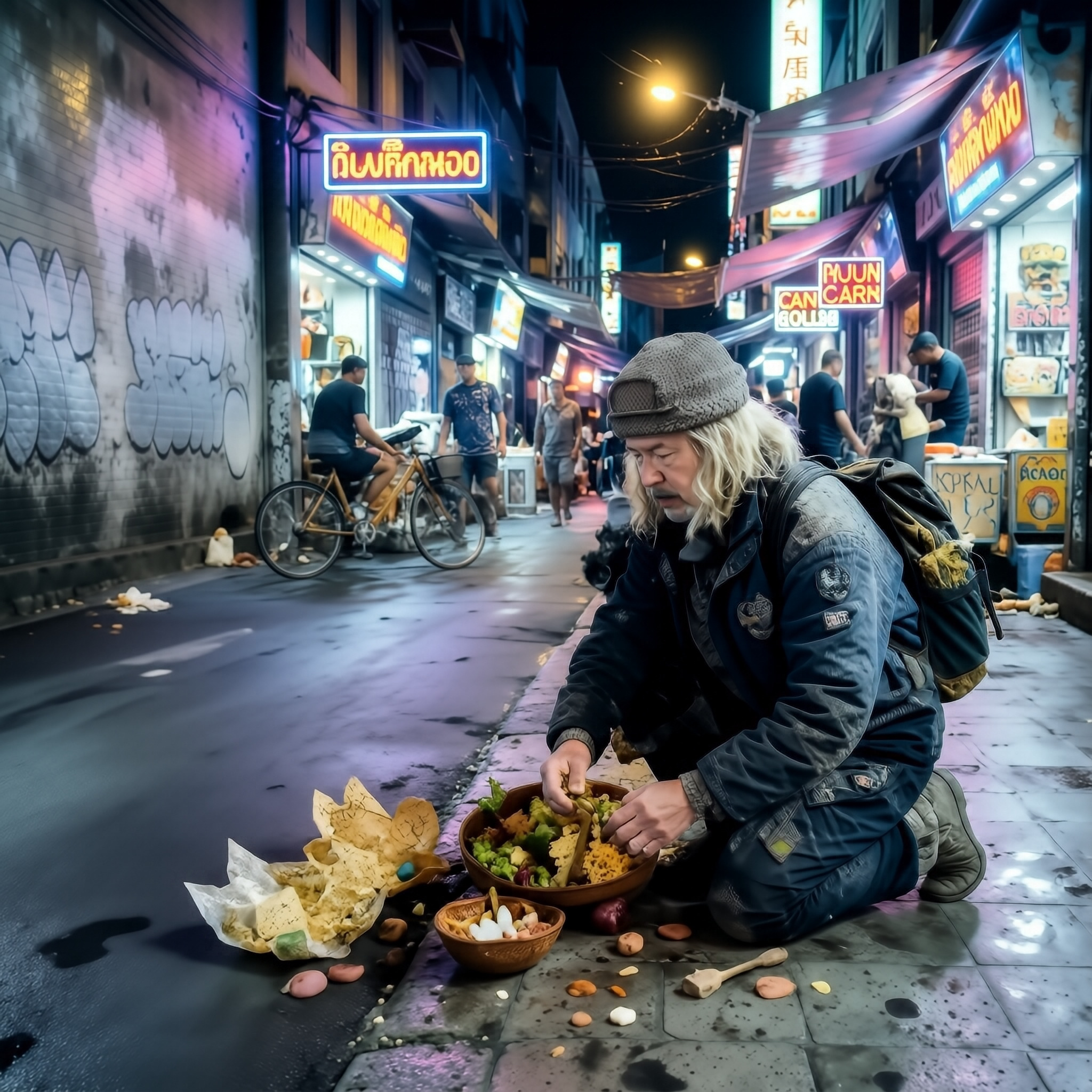}
        \caption{TH}
        \label{fig:language-coverage-case-th}
    \end{subfigure}\hfill
    \begin{subfigure}[t]{0.24\linewidth}
        \centering
        \includegraphics[width=\linewidth]{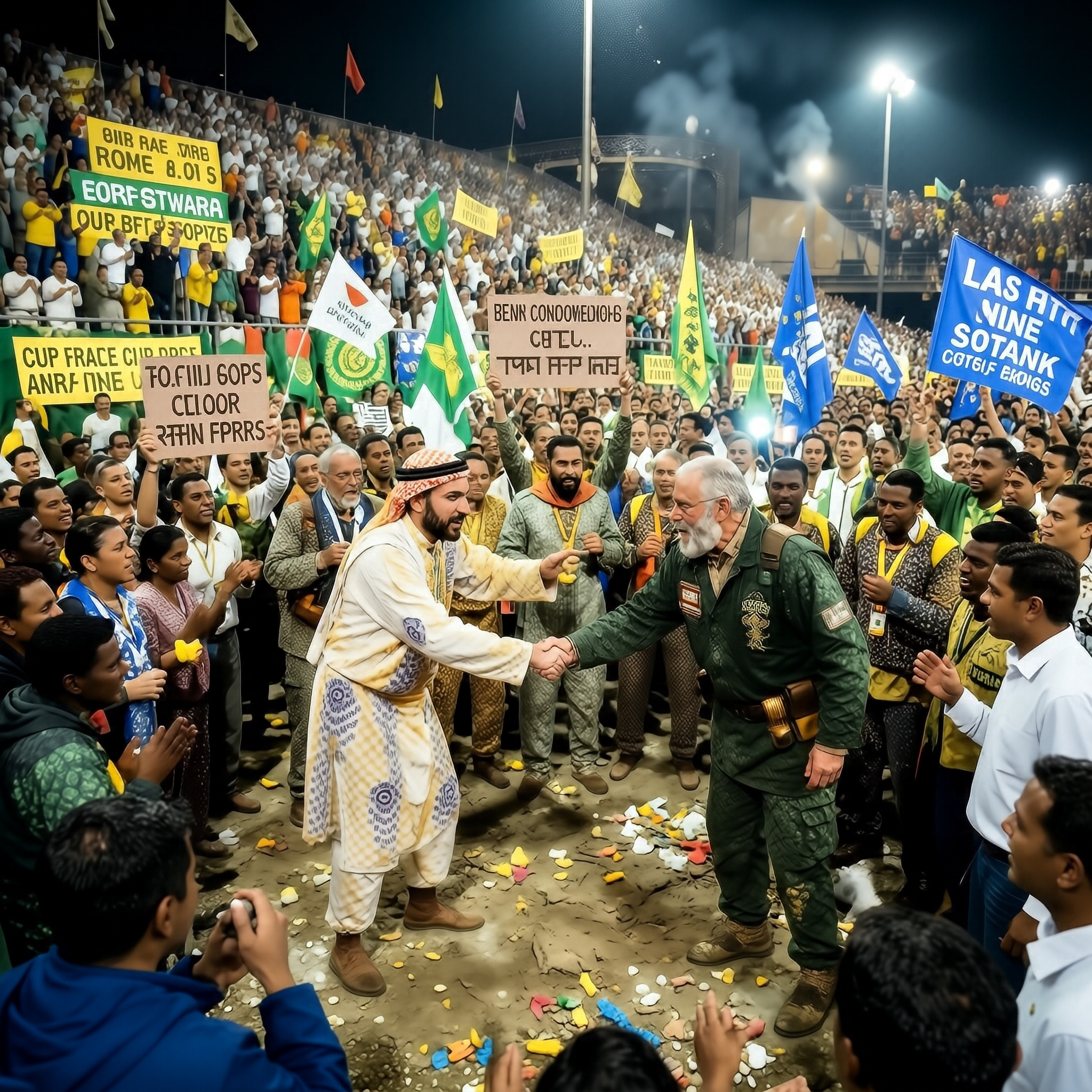}
        \caption{HI}
        \label{fig:language-coverage-case-hi}
    \end{subfigure}
    \caption{\textbf{Scenario adherence in a multilingual HiDream pilot.} English and Chinese prompts yield a couple at leisure, while Thai and Hindi prompts do not. HiDream-O1 without refinement, no cue.}
    \label{fig:language-coverage-case}
\end{figure}

\paragraph{Language coverage.}
\label{app:language-coverage}
In preliminary tests, the open-weight models followed Hindi and Thai prompts less reliably than English and Chinese ones (Figure~\ref{fig:language-coverage-case}). We therefore evaluate them only under the six shared English and Chinese conditions; the three commercial systems cover all 14 conditions (Table~\ref{tab:prompt-conditions}).

\begin{table}[ht]
    \caption{Language and geographic-cue conditions in \method. \checkmark{} marks configured conditions and a dash unconfigured ones; $^{\ast}$ marks the six conditions shared by all configurations.}
    \label{tab:prompt-conditions}
    \centering
    \small
    \begin{tabular}{@{}lccccc@{}}
        \toprule
        Prompt language & No cue & US cue & China cue & Thailand cue & India cue \\
        \midrule
        English & \checkmark$^{\ast}$ & \checkmark$^{\ast}$ & \checkmark$^{\ast}$ & \checkmark & \checkmark \\
        Chinese & \checkmark$^{\ast}$ & \checkmark$^{\ast}$ & \checkmark$^{\ast}$ & -- & -- \\
        Thai & \checkmark & \checkmark & -- & \checkmark & -- \\
        Hindi & \checkmark & \checkmark & -- & -- & \checkmark \\
        \bottomrule
    \end{tabular}
\end{table}

\paragraph{Generation settings and prompt processing.}
We submit benchmark prompts unchanged. Any rewriting occurs within the system's pipeline: HiDream's Prompt Refiner, FLUX's Prompt Upsampler, or ideogram's Magic Prompt. We keep original and processed prompts separately; failed HiDream refinements do not fall back to unrefined prompts. Local generations use seed 20260728 (Table~\ref{tab:generation-settings}).

\begin{table}[htbp]
\centering\footnotesize
\setlength{\tabcolsep}{3pt}
\caption{Generation and prompt-processing settings. ``Unspecified'' marks parameters absent from the recorded API requests.}
\label{tab:generation-settings}
\begin{tabularx}{\linewidth}{@{}p{0.20\linewidth}>{\raggedright\arraybackslash}p{0.39\linewidth}>{\raggedright\arraybackslash}X@{}}
\toprule
Component & Model identifier & Settings \\
\midrule
Nano Banana 2 & \texttt{gemini-3.1-flash-image} & Minimal thinking; seed 20260728 \\
GPT-Image-2 & \texttt{gpt-image-2} & Auto size \\
MAI & \texttt{MAI-Image-2.6-Flash} & $n=1$; seed unspecified \\
HiDream & \texttt{HiDream-O1-Image-Dev-2604} & With Refiner: flash scheduler, 28 image steps \\
HiDream Refiner & \texttt{HiDream-ai/Prompt-Refine} & $T=1$, top-$p=.95$, top-$k=64$; 2,048 output tokens \\
FLUX & \texttt{black-forest-labs/FLUX.2-dev} & 50 steps; guidance 4; bfloat16 \\
FLUX Upsampler & \texttt{Mistral-Small-3.2-}\newline\texttt{24B-Instruct-2506} & $T=.15$; 512 output tokens \\
ideogram Magic & \texttt{ideogram-4-fp8} & Magic Prompt enabled; 48 steps \\
\bottomrule
\end{tabularx}
\end{table}

\subsection{Evaluation Sample and Completion}
\label{app:sample-accounting}

Every image is evaluated on all six dimensions and scored only on its scenario's source dimension (Section~\ref{sec:orientation-aggregation}). We retain one image per available scenario--condition--configuration with complete SAE and IAE judgments from all three verifiers. Every configuration has complete judgments for at least 98\% of its planned images; the lowest is GPT-Image-2, with 5,650 of 5,760 under the six shared conditions and 13,243 of 13,440 over all configured conditions.

\subsection{Four-state Outcomes and Directional Scores}
\label{app:score-definition}

Separate support judgments for orientations A and B yield four states per dimension: A only, B only, both, and neither. Neither means that support was not established, not that the image is value-neutral. Alongside $S$ (Equation~\ref{eq:directional-score}), we report support prevalence,
\begin{equation}
C=\frac{n_{\mathrm{A\ only}}+n_{\mathrm{B\ only}}+n_{\mathrm{both}}}{N},
\label{eq:support-prevalence}
\end{equation}
where $N$ counts images in all four states. $S$ measures direction among supported images, and $C$ measures how often support is established. Images supporting both orientations enter $S$ with weight $0$. Identical $S$ and $C$ can still arise from orientations co-occurring within images or opposing each other across images.

\paragraph{Scoring on the source dimension.}
\label{app:source-dimension-scoring}
Each scenario is constructed from a question about its source dimension, so the requirements in Section~\ref{sec:scenarios} are designed for that dimension only. On another dimension, a scenario may raise no relevant issue, or its situation may favor one orientation even though its wording does not prescribe it; the filter in Section~\ref{sec:scenarios} removes only scenarios whose wording prescribes an orientation. For example, the IDV scenario ``a person at the vacation destination they chose'' leaves open whether the destination reflects personal or shared preferences, but a vacation is itself an occasion for leisure and enjoyment: under IAE, all 36 of its images from the six main systems under the six shared conditions support indulgence only. Scoring these images on IVR would reflect the scenario rather than the generation system. We therefore score each image only on its scenario's source dimension.

\subsection{Aggregation, Pairing, and Uncertainty}
\label{app:paired-analysis}

We compute profile scores $S$ from state counts pooled over the six shared conditions.

Language and cue comparisons pair scenarios within each system, hold the other prompt factor fixed, and include only scenarios available in both conditions. Each condition uses its own supported-image denominator. We report $\Delta S=S_{\mathrm{target}}-S_{\mathrm{baseline}}$.

We estimate 95\% confidence intervals for profile scores and paired score differences using 5,000 bootstrap replicates. Each replicate samples 160 scenarios with replacement per source dimension, keeping their conditions, systems, dimensions, and verifier judgments together. We then recompute both condition scores and their difference. With one image per prompt setting, the intervals reflect variation across scenarios. Pointwise intervals use bootstrap percentiles.\footnote{The bootstrap seed is 20260925. Intervals are reported only when at least 95\% of replicates yield defined estimates.}

We use approximate 95\% simultaneous confidence intervals to account for multiple comparisons in Tables~\ref{tab:full-profiles} and~\ref{tab:prompt-change-summary}. Within each method (SAE or IAE), all 36 profile estimates form one group, and each prompt comparison forms another (six main configurations $\times$ six dimensions).\footnote{The critical value is the 95th percentile of the bootstrap maximum standardized absolute deviation.}

\FloatBarrier
\section{Human Validation and Annotation Diagnostics}
\label{app:annotation-validity}
\subsection{Human Evidence-review Protocol}
\label{app:model-human-viewing}
Nine annotators reviewed the images, with cultural backgrounds spanning China, India, the United States and Korea. They participated voluntarily without compensation, gave informed consent, and could ask the authors about the instructions (Appendix~\ref{app:human-rubric}). All followed the same introduction and guided training. Annotation ran in two rounds: the first covered 20 images, each reviewed by five annotators; the second added 40 images, each reviewed by three, and images with Chinese or Hindi prompts or cues were assigned to annotators familiar with that language and culture. The main analysis uses three annotators per image, and for the first-round images the five-reviewer labels are reported as a robustness check. Reviewers saw the frozen candidate bank from the three production proposers and assessed SAE and IAE candidates separately. On each review page, candidates were grouped by proposer, with proposer identities and automatic decisions hidden, and reviewers could add their own evidence.

A reviewer supports an orientation through an accepted candidate or complete added evidence; the human majority is two of the image's three annotators. Automatic labels are recomputed from each verifier's accepted candidates in the same bank plus its added evidence, then aggregated two-of-three, so agreement measures evidence review given this bank.

\paragraph{Sample selection.}\label{app:recovered-r05}
The 60 images are selected so that each source dimension and each system family appears ten times.

\subsection{Agreement with Human Review}
\label{app:human-agreement}\label{app:recovered-r06}
Human review yields two references, because evidence for an orientation can be read either from the scenario text (SAE-H) or from the image itself (IAE-H). The two reviews rest on different information, so we make the primary comparison each method against the reference from the same candidates. Table~\ref{tab:human-review-agreement} reports this comparison on all 60 images, overall and per dimension, with 95\% intervals from 10,000 image-level bootstrap resamples; the last column gives human--human Krippendorff's $\alpha$ in the image-adaptive review.

\begin{table}[!htbp]
\centering\footnotesize
\setlength{\tabcolsep}{4pt}
\caption{Orientation agreement (\%) on the 60-image cohort, with 95\% intervals on the 0--1 scale. Each dimension has 120 judgments. The last column is human--human Krippendorff's $\alpha$ in the image-adaptive review.}\label{tab:human-review-agreement}
\begin{tabular}{@{}lrlrrlrr@{}}\toprule
& \multicolumn{3}{c}{SAE versus SAE-H} & \multicolumn{3}{c}{IAE versus IAE-H} & Human\\\cmidrule(lr){2-4}\cmidrule(l){5-7}\cmidrule(l){8-8}
Dim. & Agree. & 95\% CI & $\kappa$ & Agree. & 95\% CI & $\kappa$ & $\alpha$\\\midrule
All & 89.9 & [0.878, 0.919] & 0.584 & 84.2 & [0.807, 0.875] & 0.652 & 0.623\\
PDI & 92.5 & [0.875, 0.967] & 0.650 & 86.7 & [0.800, 0.933] & 0.658 & 0.681\\
IDV & 77.5 & [0.708, 0.833] & 0.460 & 75.0 & [0.667, 0.833] & 0.518 & 0.537\\
MAS & 91.7 & [0.867, 0.958] & 0.659 & 85.8 & [0.792, 0.917] & 0.718 & 0.565\\
UAI & 88.3 & [0.825, 0.942] & 0.198 & 84.2 & [0.775, 0.900] & 0.527 & 0.593\\
LTO & 92.5 & [0.875, 0.967] & 0.650 & 82.5 & [0.750, 0.892] & 0.624 & 0.503\\
IVR & 96.7 & [0.933, 0.992] & 0.827 & 90.8 & [0.858, 0.958] & 0.720 & 0.756\\
\bottomrule\end{tabular}

\vspace{2pt}
\begin{minipage}{\linewidth}\footnotesize\raggedright
Agreement can be high while $\kappa$ is low when one answer dominates. In the scenario-anticipated review for UAI, 88.3\% of pairs agree ($\kappa=0.20$), but 104 of the 106 agreements are pairs in which both the annotators and SAE judge the orientation unsupported; SAE recovers only 2 of the 16 human-supported judgments.
\end{minipage}
\end{table}

IAE reaches $\kappa=0.65$ against IAE-H and SAE $\kappa=0.58$ against SAE-H. Human annotators themselves agree at $\alpha=0.62$ in the image-adaptive review and $0.44$ in the scenario-anticipated review; $\alpha$ is used here because each image has three annotators, whereas $\kappa$ on the model side compares a single label with one reference. Per dimension, VLM--human $\kappa$ ranges $0.52$--$0.72$. The two methods separate on UAI, where SAE reaches $\kappa=0.20$ and IAE $0.53$; IAE is therefore used for most subsequent analyses.

To separate method error from ambiguous evidence, Table~\ref{tab:pairwise-rater-kappa} places agreement among annotators next to each method's agreement with its own human reference. Annotator agreement is Cohen's $\kappa$ in the image-adaptive review, pooled over the three pairs of an image's annotators; method agreement is $\kappa$ against the human majority (Table~\ref{tab:human-review-agreement}). On PDI, annotators agree well ($\kappa=0.68$), and so do IAE ($0.66$) and SAE ($0.65$). IDV is harder for annotators themselves ($\kappa=0.54$), and both methods are at a similar level (IAE $0.52$, SAE $0.46$), so the lower agreement reflects ambiguous IDV evidence rather than a dimension-specific failure of the methods. The only clear method-specific gap is SAE on UAI ($\kappa=0.20$), where SAE recovers only 2 of the 16 human-supported judgments while IAE reaches $\kappa=0.53$ (Appendix~\ref{app:uai-human-comparison}).

\begin{table}[!htbp]
\centering\footnotesize
\setlength{\tabcolsep}{5pt}
\caption{Agreement per dimension as Cohen's $\kappa$ (60 images, 120 judgments per dimension). Annotator vs annotator pools the three pairs of an image's annotators in the image-adaptive review; each method is compared with the human-majority reference built from its own candidates.}\label{tab:pairwise-rater-kappa}
\begin{tabular}{@{}lrrrrrrr@{}}\toprule
& All & PDI & IDV & MAS & UAI & LTO & IVR\\\midrule
Annotator vs annotator & 0.62 & 0.68 & 0.54 & 0.57 & 0.56 & 0.49 & 0.74\\
IAE vs IAE-H & 0.65 & 0.66 & 0.52 & 0.72 & 0.53 & 0.62 & 0.72\\
SAE vs SAE-H & 0.58 & 0.65 & 0.46 & 0.66 & 0.20 & 0.65 & 0.83\\
\bottomrule\end{tabular}
\end{table}

Table~\ref{tab:human-verification-summary} crosses each method with both references. We report it only for completeness.

As a robustness check, the five-reviewer labels on the first-round images give the same ordering as the three-annotator labels on those 20 images: $\kappa=0.57$ for IAE against IAE-H and $0.33$ for SAE against SAE-H, against $0.56$ and $0.42$ with three annotators; across the ten three-annotator subsets, IAE--IAE-H agreement ranges 78.8--86.2\%.

\begin{table}[!htbp]
\centering\small
\setlength{\tabcolsep}{5pt}
\caption{Cohen's $\kappa$ between each automatic method and each human-majority reference (720 image--orientation judgments).}
\label{tab:human-verification-summary}\label{tab:human-cross-reference}
\begin{tabular}{@{}lrr@{}}
\toprule
& \multicolumn{2}{c}{Human reference}\\
\cmidrule(l){2-3}
Automatic method & SAE-H & IAE-H\\
\midrule
SAE & 0.58 & 0.29\\
IAE & 0.45 & 0.65\\
\bottomrule
\end{tabular}
\end{table}

\subsection{Where Automatic and Human Review Disagree}
\label{app:annotation-disagreements}\label{app:recovered-r07}
Section~\ref{app:human-agreement} reports how often automatic and human labels agree; here we ask how they disagree. Automatic review rarely accepts evidence that annotators reject; most disagreements are evidence it misses. IAE misses less than SAE, most clearly for UAI. All counts use the 60 images and the three-annotator majority of Section~\ref{app:human-agreement}.

\paragraph{Disagreements are mostly omissions.}
Of the 529 IAE candidates that the automatic majority retains, annotators accept 489; of the 248 it rejects, they still accept 170 (SAE: 110 of 133 retained, 150 of 999 rejected). At the orientation level, 187 of the 202 IAE-supported pairs are human-supported, but IAE recovers only 187 of the 286 human-supported pairs. Orientation agreement exceeds candidate agreement because one retained candidate suffices to support an orientation. Annotators also added 17 evidence items that no proposer offered;

\paragraph{IAE's extra support is corroborated.}\label{app:recovered-r08}
IAE alone supports 132 of the 720 orientation pairs and SAE alone 1; the IAE human reference supports 117 of the 132.

\paragraph{UAI shows the largest gap.}\label{app:uai-human-comparison}
Each method compared with its own human reference, IAE recovers 15 of the 33 human-supported UAI judgments, below its overall 65\%, and SAE recovers 2 of 16. Both rarely accept unsupported UAI evidence (one false positive for IAE, none for SAE). These results concern orientation support, not the direction of individual systems' production UAI scores.

\subsection{Pairwise Comparison of Evidence Strength}
\label{app:pairwise-strength}
Our score records whether an image supports an orientation, not how strongly. We test whether VLM judges can compare the strength of evidence.

\textbf{Setup.} We sample 60 image pairs that cover every dimension, every system and four prompt languages; each pair shows two images generated from the same prompt by different systems. Three annotators from our human study and the three VLM judges answer which image expresses a given orientation more strongly: Left, Right or About the same. Each VLM judge answers in both left--right orders, and its answer counts only if the two agree.

\par\medskip\noindent\begin{minipage}{\linewidth}\captionsetup{type=table}
\centering\footnotesize
\caption{Agreement on pairwise strength judgments (60 pairs, three answer options). Within-group rows average the three pairs of raters; VLM rows compare with the majority of the three annotators.}
\label{tab:pairwise-strength}
\begin{tabular}{@{}lrr@{}}
\toprule
Comparison & Agreement & Cohen's $\kappa$\\
\midrule
Annotator vs.\ annotator & 54\% & 0.30\\
VLM judge vs.\ VLM judge & 74\% & 0.55\\
\midrule
Gemini vs.\ human majority & 50\% & 0.30\\
Muse vs.\ human majority & 37\% & 0.14\\
Luna vs.\ human majority & 37\% & 0.12\\
VLM majority vs.\ human majority & 38\% & 0.15\\
\bottomrule
\end{tabular}
\end{minipage}\par\medskip

\textbf{Results.}
\begin{itemize}[leftmargin=*,itemsep=2pt,topsep=2pt]
\item \textit{Annotators agree only moderately with each other.} Their pairwise agreement is 54\% ($\kappa=0.30$).
\item \textit{VLM judges agree with each other more closely than annotators do.} Their pairwise agreement is 74\% ($\kappa=0.55$).
\item \textit{But VLM judges agree far less with humans.} The best judge, Gemini, reaches $\kappa=0.30$ with the human majority, the other two judges $0.12$--$0.14$, and the VLM majority $0.15$.
\item \textit{The disagreement lies in judging whether two images differ.} The human majority answered About the same in 30 of the 60 pairs, the VLM majority in only 5. When the human majority chose an image, the VLM majority chose the same image in 22 of 30 pairs.
\end{itemize}
These results show that \textbf{VLM judges are internally consistent but share a criterion that differs from human judgment}: they rarely treat two images as equally strong. They therefore \textbf{cannot yet measure the strength of value evidence reliably}. This is why our score records whether an orientation is present rather than how strongly; on that task, IAE reaches $\kappa=0.65$ with human reviewers (Table~\ref{tab:human-review-agreement}). A strength-aware score would first require calibrating VLM judges against human judgments (Appendix~\ref{sec:limitations}).

\FloatBarrier
\newenvironment{AppendixResultBlock}[1]{\par\medskip\noindent\begin{minipage}{\linewidth}\captionsetup{type=#1}}{\end{minipage}\par\medskip}
\section{Extended Results}
\label{app:extended-results}

\subsection{Complete Value Profiles}
\label{app:full-profiles}
\label{app:recovered-r09}

Table~\ref{tab:full-profiles} reports pooled orientation scores for each main configuration under the six shared conditions, scoring each dimension on its scenarios only. SAE and IAE agree on score signs in all dimensions other than UAI; the only other discrepancy is MAI's near-zero IDV score ($-0.02$ under IAE, $+0.07$ under SAE).

\begin{AppendixResultBlock}{table}
\centering\small
\setlength{\tabcolsep}{4pt}
\caption{Pooled orientation scores under the six shared conditions, with each dimension scored on its scenarios only. Bold: within-method simultaneous 95\% interval excludes zero.}\label{tab:full-profiles}
\begin{tabular}{@{}llrrrrrr@{}}
\toprule
Method & System & PDI & IDV & MAS & UAI & LTO & IVR\\\midrule
SAE & MAI & $\mathbf{-0.822}$ & $0.067$ & $\mathbf{-0.467}$ & $-0.157$ & $\mathbf{0.780}$ & $\mathbf{0.692}$\\
SAE & GPT Image 2 & $\mathbf{-0.827}$ & $0.066$ & $-0.246$ & $0.062$ & $\mathbf{0.597}$ & $\mathbf{0.720}$\\
SAE & Nano Banana 2 & $\mathbf{-0.904}$ & $0.204$ & $\mathbf{-0.329}$ & $-0.275$ & $\mathbf{0.524}$ & $\mathbf{0.804}$\\
SAE & FLUX + Upsampler & $\mathbf{-0.711}$ & $\mathbf{0.483}$ & $-0.194$ & $-0.225$ & $\mathbf{0.545}$ & $\mathbf{0.876}$\\
SAE & HiDream + Refiner & $\mathbf{-0.711}$ & $\mathbf{0.344}$ & $-0.246$ & $-0.160$ & $\mathbf{0.457}$ & $\mathbf{0.880}$\\
SAE & Ideogram Magic & $\mathbf{-0.836}$ & $0.187$ & $-0.288$ & $-0.236$ & $\mathbf{0.588}$ & $\mathbf{0.822}$\\
\midrule
IAE & MAI & $\mathbf{-0.793}$ & $-0.021$ & $\mathbf{-0.210}$ & $-0.065$ & $\mathbf{0.718}$ & $\mathbf{0.532}$\\
IAE & GPT Image 2 & $\mathbf{-0.701}$ & $0.077$ & $-0.111$ & $0.203$ & $\mathbf{0.601}$ & $\mathbf{0.585}$\\
IAE & Nano Banana 2 & $\mathbf{-0.741}$ & $0.148$ & $-0.165$ & $0.187$ & $\mathbf{0.670}$ & $\mathbf{0.735}$\\
IAE & FLUX + Upsampler & $\mathbf{-0.602}$ & $\mathbf{0.337}$ & $-0.064$ & $\mathbf{0.242}$ & $\mathbf{0.639}$ & $\mathbf{0.800}$\\
IAE & HiDream + Refiner & $\mathbf{-0.691}$ & $\mathbf{0.285}$ & $-0.148$ & $\mathbf{0.258}$ & $\mathbf{0.638}$ & $\mathbf{0.795}$\\
IAE & Ideogram Magic & $\mathbf{-0.715}$ & $0.199$ & $\mathbf{-0.193}$ & $\mathbf{0.284}$ & $\mathbf{0.651}$ & $\mathbf{0.708}$\\
\bottomrule
\end{tabular}
\end{AppendixResultBlock}

\paragraph{Why MAI differs on UAI.}
\label{app:mai-uai-case}
We examine why MAI-2.6-Flash has a slightly negative IAE UAI score, while the other five main configurations have positive scores. Across the six shared conditions, 944 MAI-2.6-Flash images of UAI scenarios have complete IAE judgments: 269 support only high UAI, 316 only low UAI, 138 both, and 221 neither. The score is based on 723 supported images ($S=-0.065$; 95\% CI $[-0.192,0.062]$). Images supporting both orientations make up 19\% of MAI-2.6-Flash's supported images, compared with 4--12\% for the other five configurations.

\begin{figure}[!htbp]
\centering
\captionsetup[subfigure]{justification=centering,skip=2pt,font=small}
\begin{subfigure}[t]{0.24\linewidth}\centering
\includegraphics[width=\linewidth]{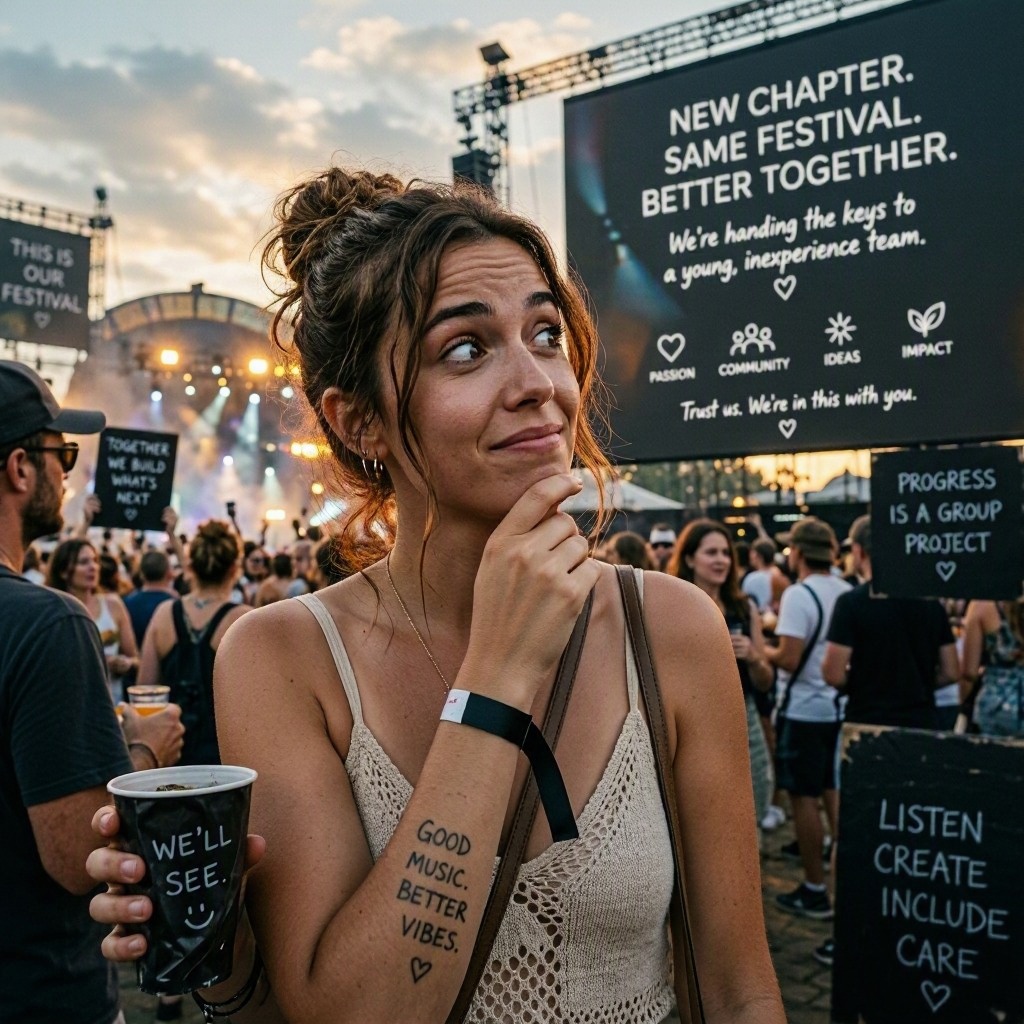}
\caption{a festival attendee reacting to the young, inexperienced team taking over the festival's management}
\label{fig:mai-artwork}
\end{subfigure}\hfill
\begin{subfigure}[t]{0.24\linewidth}\centering
\includegraphics[width=\linewidth]{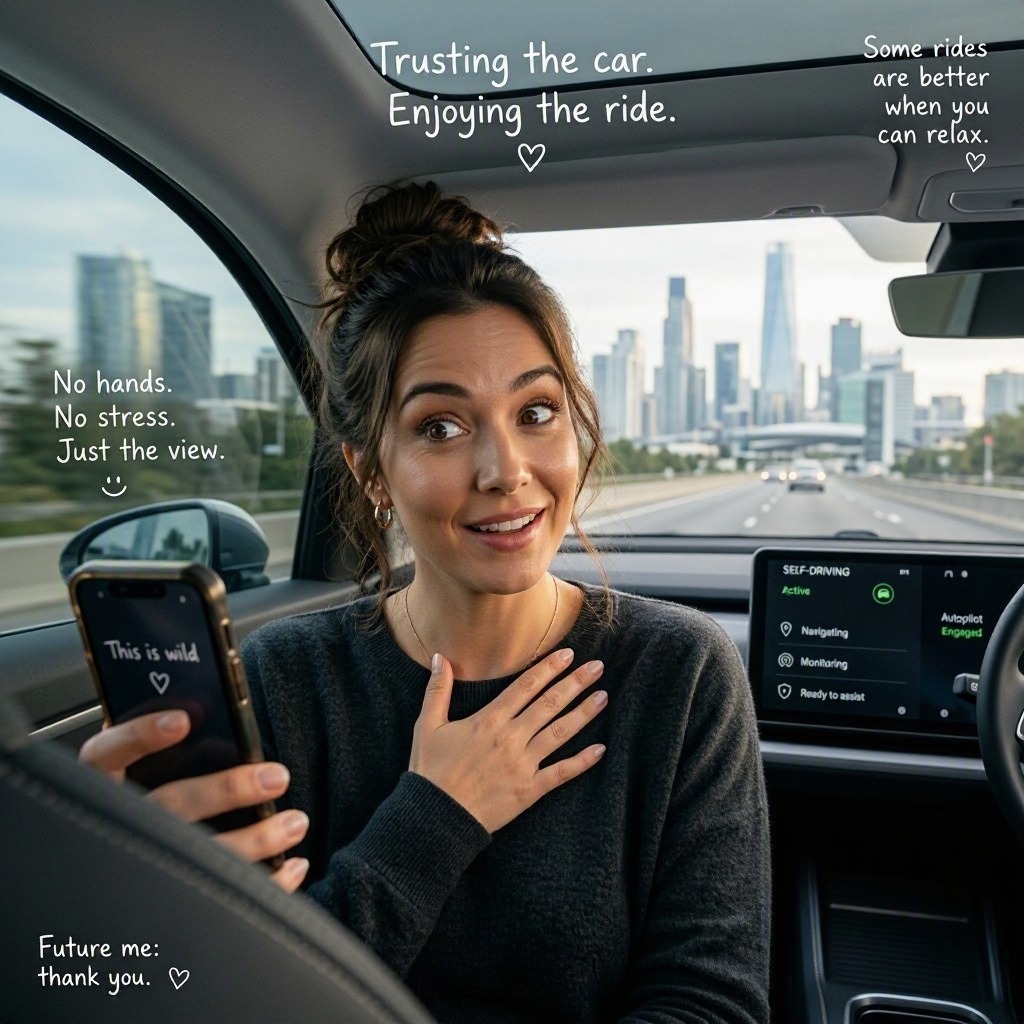}
\caption{a person reacting to riding in a self-driving car}
\label{fig:mai-car}
\end{subfigure}\hfill
\begin{subfigure}[t]{0.24\linewidth}\centering
\includegraphics[width=\linewidth]{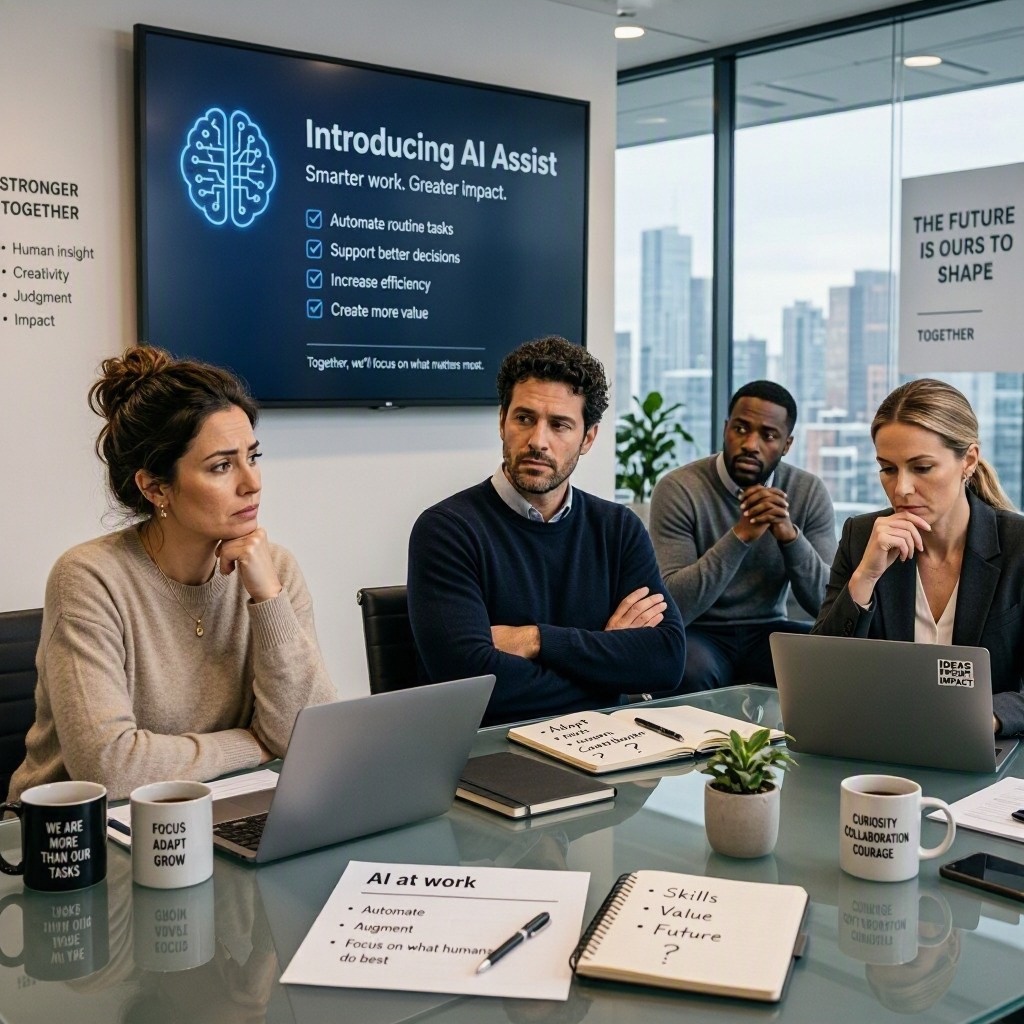}
\caption{employees reacting to the introduction of an AI system that may replace some of their duties}
\label{fig:mai-ai-work}
\end{subfigure}\hfill
\begin{subfigure}[t]{0.24\linewidth}\centering
\includegraphics[width=\linewidth]{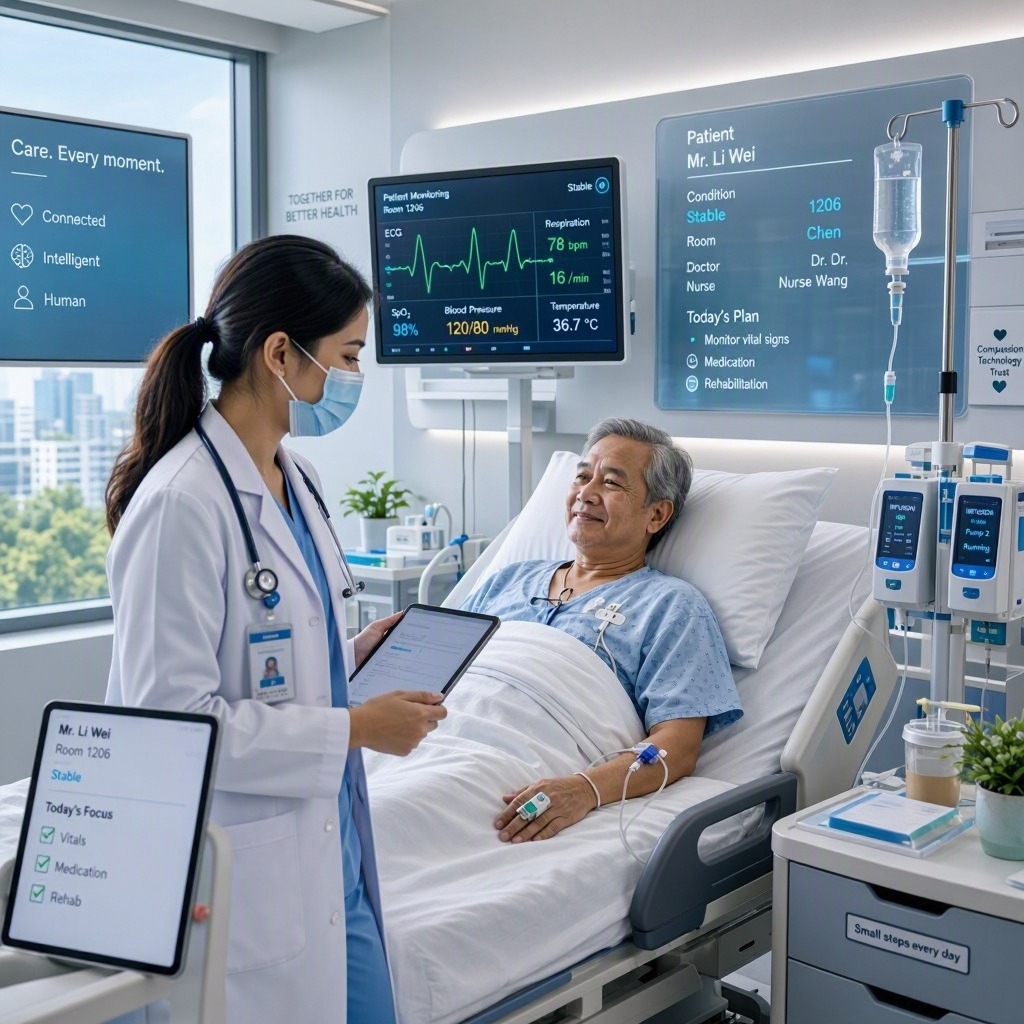}
\caption{a healthcare professional attending to a patient in a hospital room equipped with advanced medical technology}
\label{fig:mai-hospital}
\end{subfigure}
\caption{\textbf{Poster-like images generated by MAI-2.6-Flash.} Prompts are in English for (a)--(c) and Chinese for (d), all without a geographic cue.}
\label{fig:mai-uai-text}
\end{figure}

We next inspect a selection of MAI-generated images. In most outputs, added slogans, dialogue, and explanatory text make attitudes explicit and give the scene a poster-like presentation. Figure~\ref{fig:mai-uai-text} shows four examples of this style.

In panel~(\subref{fig:mai-artwork}), a screen announces that the festival is being handed to a young, inexperienced team and asks attendees to trust it, while the attendee's skeptical expression and her cup reading ``We'll see.'' express doubt, introducing competing UAI cues. All three IAE verifiers support both orientations for this image. In panel~(\subref{fig:mai-car}), slogans promote trust in the car, while the foreground hand holding a phone creates an awkward composition. In panel~(\subref{fig:mai-ai-work}), positive promotional slogans about AI contrast with the employees' concerned expressions. In panel~(\subref{fig:mai-hospital}), the information board lists ``Chen'' as the room and ``Dr. Dr.'' as the doctor.

MAI's poster-like presentation introduces attitudes through added text. However, these attitudes may conflict with one another or with those depicted in the focal situation.

\subsection{Prompt-Language and Geographic-Cue Effects}
\label{app:language-no-cue}
\label{app:recovered-r10}
\label{app:recovered-r11}
\label{app:recovered-r12}
\label{app:recovered-r13}

Table~\ref{tab:prompt-change-summary} summarizes paired score changes for prompt-language and geographic-cue contrasts under the six shared conditions (Appendix~\ref{app:paired-analysis}), with each dimension scored on its scenarios only.

\begin{AppendixResultBlock}{table}
\centering\footnotesize
\setlength{\tabcolsep}{4pt}
\caption{Paired score changes with each dimension scored on its scenarios only. Med./Max.: median/maximum $|\Delta S|$; Keep: number of unchanged score signs out of 36; Sim.: shifts with simultaneous 95\% intervals excluding zero.}\label{tab:prompt-change-summary}\label{tab:language-summary}\label{tab:cue-summary}
\begin{tabular}{@{}lrrrrrrrr@{}}
\toprule
& \multicolumn{4}{c}{SAE} & \multicolumn{4}{c}{IAE}\\
\cmidrule(lr){2-5}\cmidrule(l){6-9}
Contrast; fixed condition & Med. & Max. & Keep & Sim. & Med. & Max. & Keep & Sim.\\\midrule
EN $\to$ ZH; no cue & 0.084 & 0.286 & 35 & 0 & 0.072 & 0.204 & 36 & 0\\
EN $\to$ ZH; US cue & 0.063 & 0.324 & 36 & 0 & 0.066 & 0.194 & 34 & 0\\
EN $\to$ ZH; China cue & 0.070 & 0.320 & 32 & 0 & 0.074 & 0.320 & 34 & 1\\
\midrule
none $\to$ US; EN & 0.060 & 0.280 & 36 & 0 & 0.053 & 0.182 & 35 & 0\\
none $\to$ China; EN & 0.074 & 0.553 & 34 & 2 & 0.137 & 0.362 & 32 & 7\\
US $\to$ China; EN & 0.108 & 0.493 & 33 & 2 & 0.075 & 0.409 & 32 & 9\\
\midrule
none $\to$ US; ZH & 0.069 & 0.335 & 36 & 0 & 0.060 & 0.223 & 35 & 1\\
none $\to$ China; ZH & 0.114 & 0.319 & 33 & 0 & 0.092 & 0.327 & 33 & 1\\
US $\to$ China; ZH & 0.124 & 0.414 & 34 & 3 & 0.088 & 0.440 & 33 & 6\\
\bottomrule
\end{tabular}
\end{AppendixResultBlock}

\begin{figure}[!tbp]
    \centering
    \includegraphics[width=\linewidth,trim=0 3bp 0 0,clip]{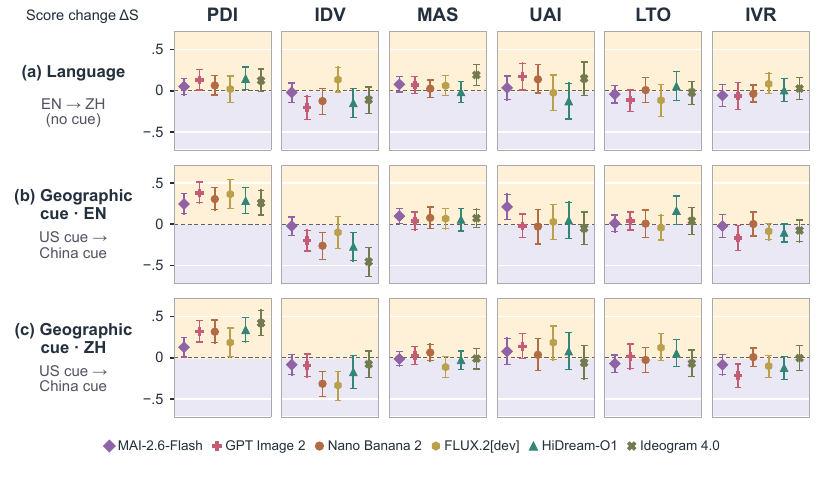}
    \caption{\textbf{China cues shift PDI and IDV more than switching prompt language.} Paired IAE score changes (target minus baseline) for the comparisons in Figure~\ref{fig:main-prompt-effects}, computed on each dimension's scenarios; bars show pointwise 95\% paired-bootstrap intervals.}
    \label{fig:main-prompt-effects-deltas}
\end{figure}

Figure~\ref{fig:main-prompt-effects-deltas} shows paired IAE score changes and their pointwise 95\% confidence intervals.

\subsection{Effects of Prompt Refinement}
\label{app:refinement-results}
\label{sec:prompt-refinement-effects}

\begin{AppendixResultBlock}{table}
\centering\small
\setlength{\tabcolsep}{6pt}
\caption{Score changes from prompt refinement, $\Delta S$ [pointwise 95\% CI], on 960 paired scenarios across six conditions (5,760 images per arm), under the six-dimension scoring of Appendix~\ref{app:six-dimension-scoring}.}\label{tab:refinement}
\begin{tabular}{@{}llrr@{}}
\toprule
Method & Dimension & HiDream Refiner & FLUX Upsampler\\\midrule
SAE & PDI & $+0.128$ [+0.057, +0.203] & $-0.136$ [-0.232, -0.046]\\
SAE & IDV & $+0.248$ [+0.166, +0.330] & $+0.166$ [+0.097, +0.234]\\
SAE & MAS & $+0.097$ [+0.008, +0.186] & $-0.005$ [-0.099, +0.088]\\
SAE & UAI & $+0.602$ [+0.418, +0.786] & $-0.246$ [-0.447, -0.041]\\
SAE & LTO & $+0.182$ [+0.021, +0.339] & $+0.230$ [+0.075, +0.385]\\
SAE & IVR & $-0.050$ [-0.086, -0.017] & $+0.057$ [-0.012, +0.129]\\
\midrule
IAE & PDI & $+0.121$ [+0.063, +0.181] & $-0.080$ [-0.152, -0.012]\\
IAE & IDV & $+0.131$ [+0.087, +0.176] & $+0.068$ [+0.030, +0.106]\\
IAE & MAS & $+0.122$ [+0.089, +0.156] & $+0.037$ [+0.003, +0.071]\\
IAE & UAI & $+0.539$ [+0.422, +0.649] & $-0.269$ [-0.353, -0.189]\\
IAE & LTO & $+0.007$ [-0.057, +0.070] & $+0.063$ [+0.009, +0.120]\\
IAE & IVR & $-0.083$ [-0.121, -0.048] & $+0.024$ [-0.016, +0.068]\\
\bottomrule
\end{tabular}
\end{AppendixResultBlock}

Value expression differs in strength and direction between configurations with and without prompt processing (Table~\ref{tab:refinement}). Under IAE and six-dimension scoring, HiDream's UAI reverses from $-0.357$ to $+0.182$, while FLUX's falls from $+0.411$ to $+0.142$; under source-dimension scoring, the refined HiDream configuration's UAI score is $+0.258$.

\begin{figure}[!htbp]
\centering
\captionsetup[subfigure]{justification=centering,skip=2pt,font=small}
\begin{subfigure}[t]{0.3\linewidth}\centering
\includegraphics[width=\linewidth]{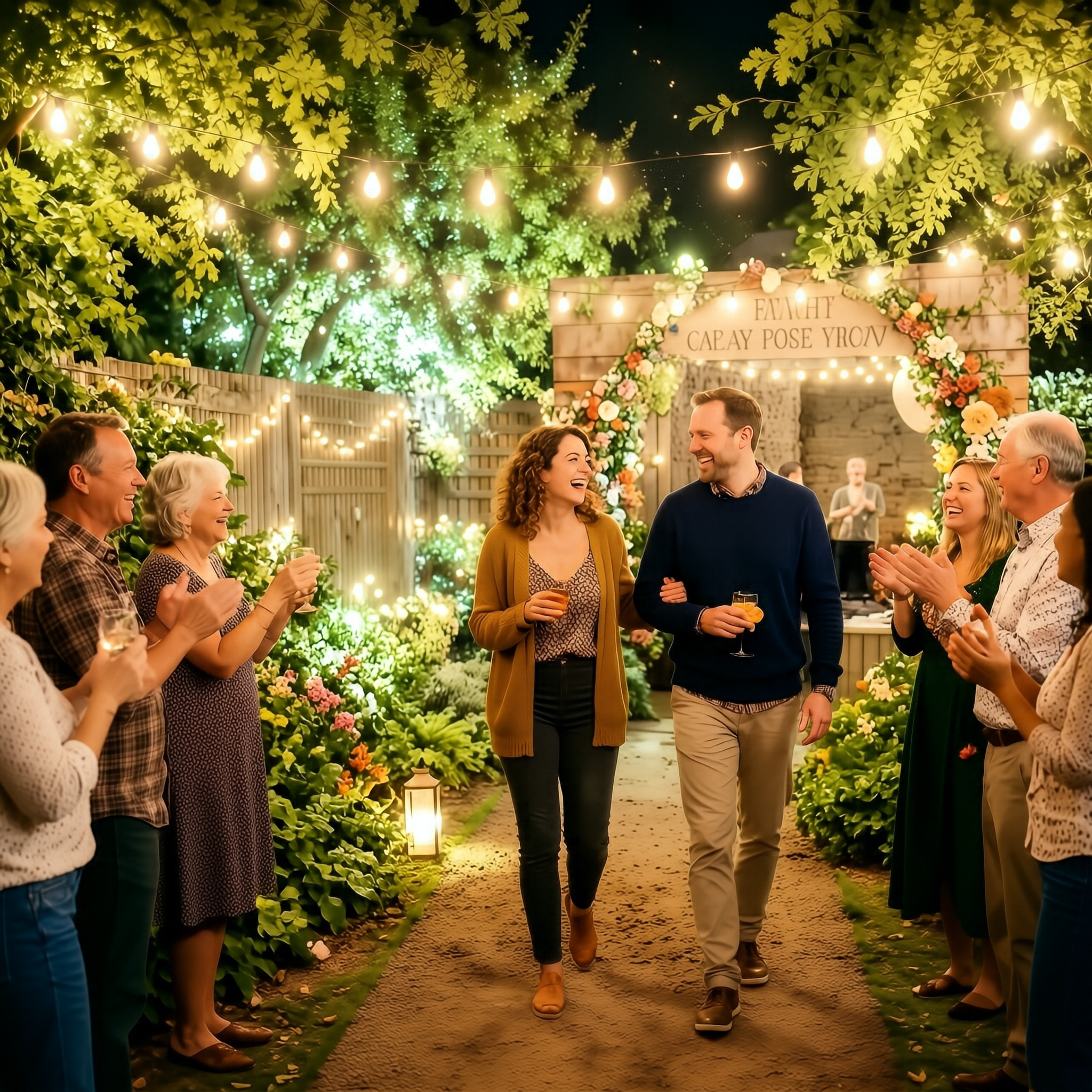}
\caption{Without refiner: collectivism only.}
\end{subfigure}\hspace{0.06\linewidth}
\begin{subfigure}[t]{0.3\linewidth}\centering
\includegraphics[width=\linewidth]{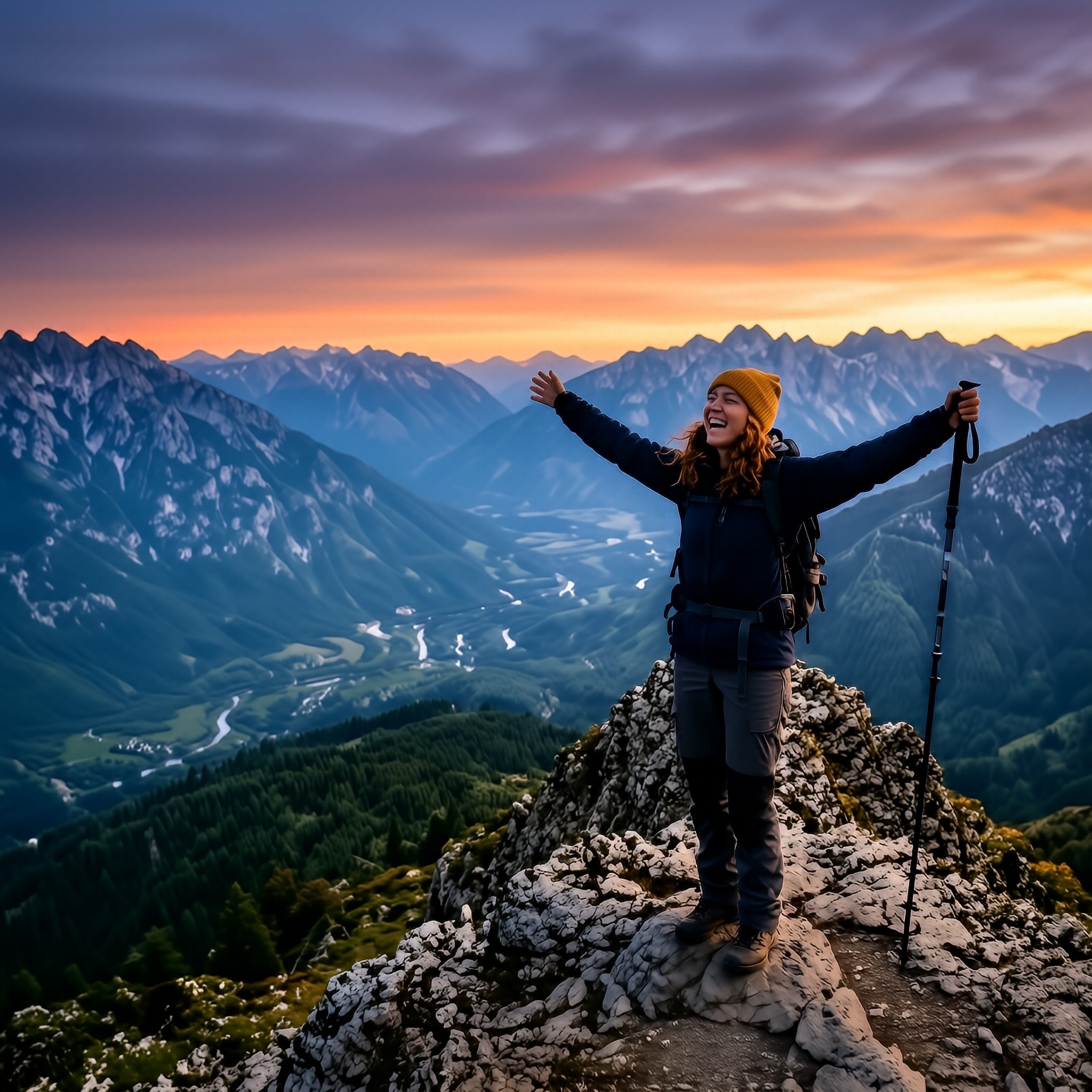}
\caption{With refiner: individualism only.}
\end{subfigure}
\caption{\textbf{HiDream-O1 with and without prompt refinement} (English, no cue; CDEval \texttt{ivr\_0356}; matched prompt and seed).}
\label{fig:refinement-family-case}
\end{figure}

\paragraph{Case study: from a family celebration to a personal pursuit.}
Value orientations in an image-generation system may arise from the prompt refiner, the diffusion model, or both (Figure~\ref{fig:refinement-family-case}). The prompt describes ``a person following through on the choice they made between attending a family celebration and pursuing a personal plan that brings enjoyment.'' HiDream's refiner rewrites it as ``a person with a joyful expression [who] stands on a mountain summit with arms outstretched,'' dropping the family alternative. Our main analysis therefore characterizes visual value leakage in complete image-generation systems. Future research can disentangle the contributions of prompt refiners and diffusion models.

\subsection{Additional Language and Geographic-Cue Shifts}
\label{app:additional-languages}
\label{app:expanded-shifts}

\begin{figure}[!tbp]
    \centering
    \includegraphics[width=\linewidth]{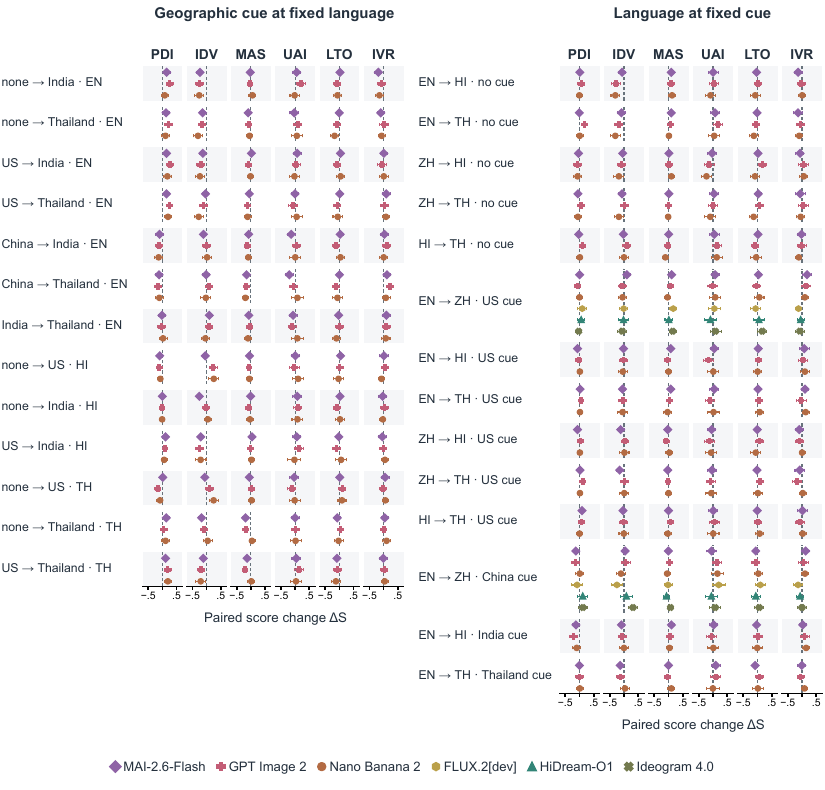}
    \caption{\textbf{Additional IAE contrasts.} Left: geographic cue at fixed language; right: language at fixed cue. Rows give baseline $\rightarrow$ target and the fixed factor. Bars are pointwise 95\% bootstrap intervals.}
    \label{fig:expanded-cue-matrix}
    \label{fig:expanded-cue-english-1}
    \label{fig:expanded-cue-english-2}
    \label{fig:expanded-cue-hindi}
    \label{fig:expanded-cue-thai}
    \label{fig:expanded-language-matrix}
    \label{fig:expanded-shared-language}
    \label{fig:expanded-language-none-1}
    \label{fig:expanded-language-none-2}
    \label{fig:expanded-language-us-1}
    \label{fig:expanded-language-us-2}
    \label{fig:expanded-language-local}
\end{figure}

Across the additional languages and geographic cues, value expression varies in strength while its direction largely persists (Figure~\ref{fig:expanded-cue-matrix}). As with China cues, India and Thailand cues raise PDI and lower IDV relative to a US cue, under both English and the local-language prompts. Switching from English to Hindi or Thai without a cue changes scores by a median of $0.07$ and $0.08$ in absolute value, comparable to $0.07$ for switching to Chinese. Most score-sign changes occur on IDV, where the commercial systems' scores are close to zero. The 27 IAE contrasts use English, Chinese, Hindi, and Thai prompts (EN, ZH, HI, TH). EN-to-ZH contrasts under US and China cues cover all six systems; the remaining contrasts cover the three commercial systems. All scores are computed on each dimension's scenarios.

\FloatBarrier
\section{Robustness and Sensitivity Analyses}\label{app:robustness}
We assess sensitivity to verifier aggregation and scenario partitions across six image-generation systems under the same prompt conditions.
We also compare individual verifiers.

\subsection{Verifier Aggregation}\label{app:aggregation-sensitivity}
\begin{figure}[!htbp]
\centering
\includegraphics[width=\linewidth]{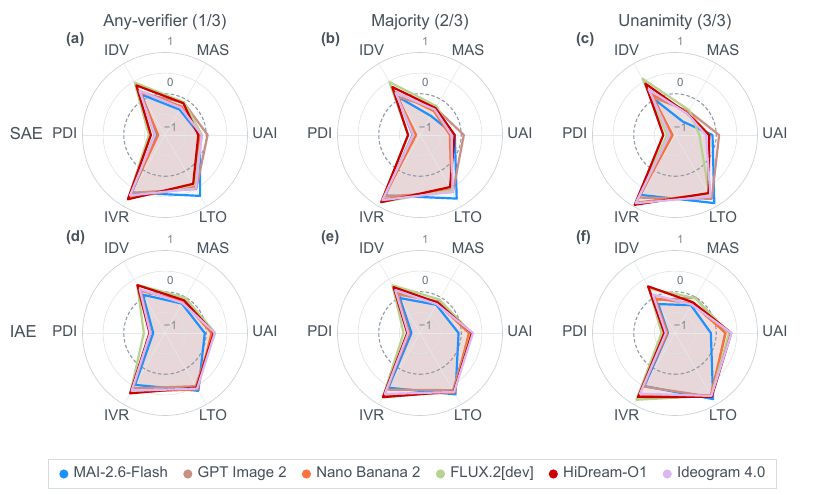}
\caption{\textbf{Score directions in four dimensions persist across aggregation rules.} Columns show any-verifier (1/3), majority (2/3), and unanimity (3/3); rows show SAE and IAE. Each profile pools the six shared conditions. Dashed circles mark zero.}\label{fig:aggregation-profiles}
\end{figure}

We recompute SAE and IAE scores under any-verifier (1/3), majority (2/3), and unanimity (3/3) aggregation, keeping the three verifiers, images, and judgments fixed (Figure~\ref{fig:aggregation-profiles}).

Score directions in PDI, MAS, LTO, and IVR persist across all three aggregation rules. SAE preserves all 36 directions. Under IAE, unanimity turns IDV negative for MAI, GPT Image 2, and Nano Banana 2; the UAI direction is stable within each method but remains weak for MAI. Unanimity reduces the share of IAE-supported images from 73.0\% to 36.3\% for IDV and from 89.4\% to 59.9\% for MAS.

\begin{figure}[!htbp]
\centering
\includegraphics[width=\linewidth]{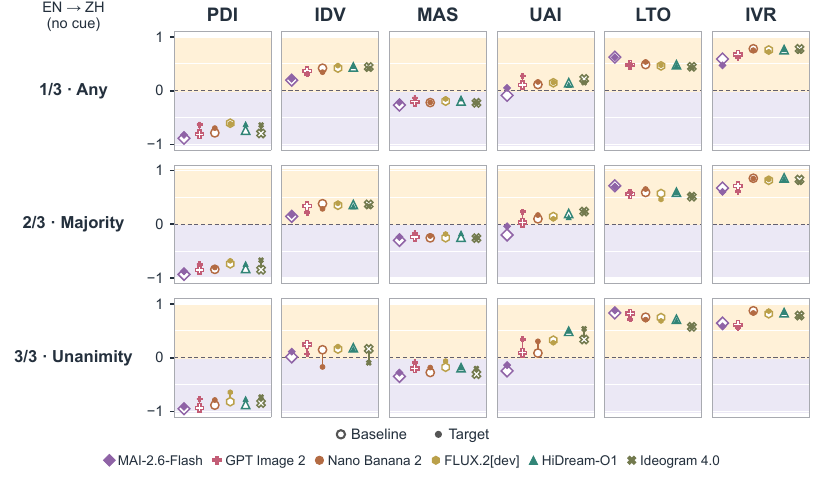}
\caption{\textbf{Median absolute language effects remain small under all three aggregation rules.} IAE scores before and after switching from English to Chinese without cues. Rows show the aggregation rules; open and filled markers show baseline and target scores on matched scenarios.}\label{fig:aggregation-language}
\end{figure}

\begin{figure}[!htbp]
\centering
\includegraphics[width=\linewidth]{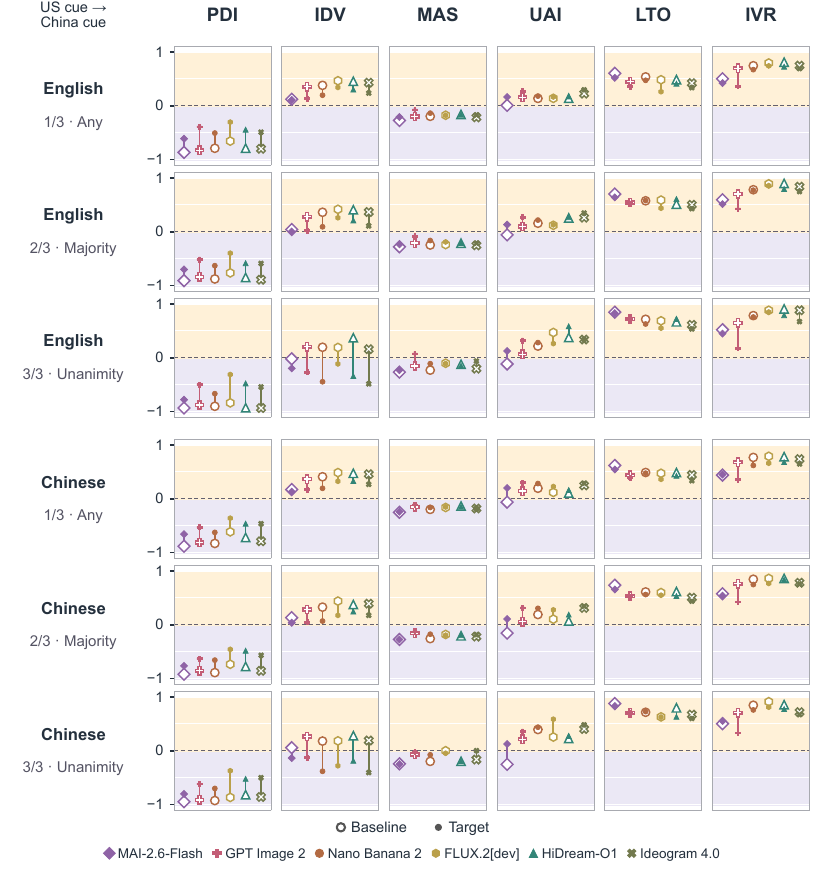}
\caption{\textbf{US-to-China cues raise PDI and lower IDV under all three aggregation rules.} IAE comparisons with English and Chinese prompts. Open and filled markers show US-cue and China-cue scores on matched scenarios; lines connect scores from the same system.}\label{fig:aggregation-geographic-cue}
\end{figure}

\paragraph{Paired language and geographic-cue effects.}\label{app:recovered-r16}
To test whether the main prompt findings depend on verifier aggregation, we repeat the main IAE language and geographic-cue comparisons (Figures~\ref{fig:aggregation-language} and~\ref{fig:aggregation-geographic-cue}). Switching from English to Chinese without cues yields small median absolute score changes under any-verifier and majority rules ($0.064$ and $0.067$) and a larger one under unanimity ($0.120$), where fewer images are supported. Switching from US to China cues raises PDI in all six systems under every rule, and lowers IDV in all six systems under majority and unanimity and in five of six under any-verifier, in both English and Chinese.

\subsection{Verifier Differences and Human Agreement}\label{app:verifier-differences}
Within each method, the three verifiers receive the same candidate bank. We compare their individual support, candidate acceptance, and human agreement (Table~\ref{tab:verifier-diagnostics}).

\begingroup\footnotesize\setlength{\tabcolsep}{3pt}
\begin{table}[!htbp]\centering
\caption{Verifier diagnostics (\%, except $\kappa$). Support and acceptance use the production images (412,524 image--orientation slots per method; 662,844 SAE and 434,932 IAE candidate appearances per verifier); agreement uses the 720 image--orientation judgments per method on the 60 human-reviewed images.}\label{tab:verifier-diagnostics}
\renewcommand{\arraystretch}{1.15}
\begin{tabularx}{\linewidth}{l*{8}{>{\raggedleft\arraybackslash}X}}
\toprule
\rowcolor[HTML]{E8EBEF}
& \multicolumn{4}{c}{\textbf{SAE}} & \multicolumn{4}{c}{\textbf{IAE}} \\
\cmidrule(lr){2-5}\cmidrule(l){6-9}
Verifier & Support & Accept & Agree. & $\kappa$ & Support & Accept & Agree. & $\kappa$ \\
\midrule
Luna & 13.0 & 13.7 & 89.3 & 0.582 & 29.4 & 69.3 & 84.9 & 0.672\\
\rowcolor[HTML]{F5F6F8}
Gemini & 6.7 & 9.0 & 86.5 & 0.399 & 13.2 & 36.7 & 75.4 & 0.431\\
Muse & 12.4 & 14.5 & 89.7 & 0.597 & 30.7 & 72.1 & 85.8 & 0.695\\
\bottomrule\end{tabularx}\end{table}
\endgroup

The three verifiers show different judgment preferences even with a shared candidate bank. Gemini supports fewer orientations and accepts fewer candidates than Luna and Muse under both methods.

\paragraph{Comparison with simpler evaluation.} To test whether pooling candidates and voting is needed, we compare the production pipeline with simpler variants on the 60 human-reviewed images (Table~\ref{tab:simple-eval}). In the single-model variant, each model proposes and verifies evidence on its own with the same prompts, so it supports an orientation only through candidates it proposed itself or evidence it added. The pipeline agrees with the human majority more closely than the vote over the three single-model labels and than two of the three single models; Muse alone is slightly higher, but single models vary widely ($\kappa=0.42$--$0.68$), and relying on one evaluation model would carry that model's biases into the results. Pooling candidates and voting keep agreement close to the best single model without depending on any one model.

\begin{table}[!htbp]
\centering\footnotesize
\setlength{\tabcolsep}{5pt}
\caption{Cohen's $\kappa$ between each evaluation variant and the human majority in the image-adaptive review (IAE versus IAE-H), on the 60 human-reviewed images (720 image--orientation judgments).}\label{tab:simple-eval}
\begin{tabular}{@{}lr@{}}\toprule
Method & $\kappa$\\\midrule
Pooled candidates $\rightarrow$ three-verifier vote (production) & 0.65\\
Single model $\rightarrow$ three-model vote & 0.62\\
Single model: Gemini & 0.42\\
Single model: Luna & 0.60\\
Single model: Muse & 0.68\\
\bottomrule\end{tabular}
\end{table}

\subsection{Stability across Scenario Partitions}
\label{app:scenario-partition-stability}
To assess whether 960 scenarios are sufficient to recover stable value directions, we compare the full sample with four quarters of 240 (40 per source dimension) and two halves of 480 (H1=Q1+Q2; H2=Q3+Q4). The fixed random partition (seed 20260926) keeps repeated texts together without using outcomes. We recompute profiles within each group.

\begin{figure}[!tbp]
\centering
\includegraphics[width=\linewidth]{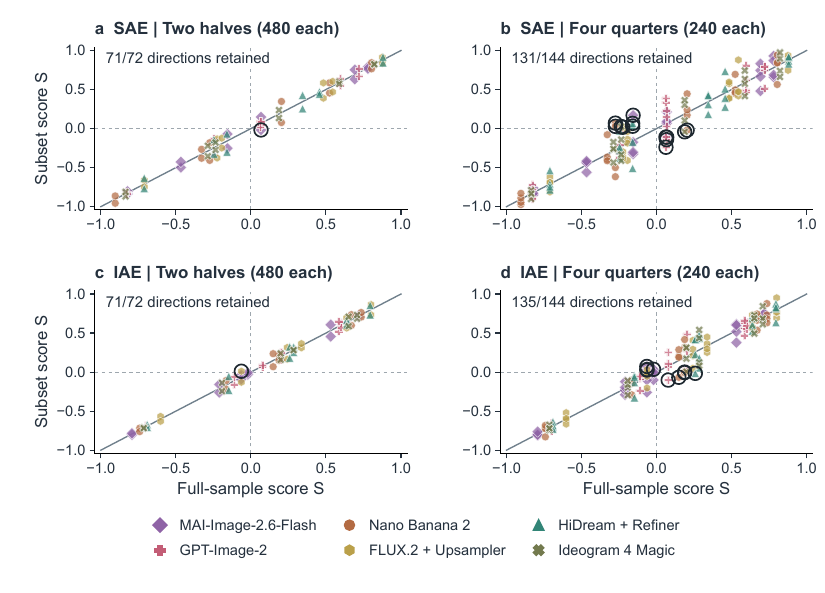}
\caption{\textbf{Value profiles across scenario partitions.} Each point compares a system--dimension score in a half or quarter with the full-sample score; open rings mark sign changes. Colors and markers follow Figure~\ref{fig:model-profiles}.}
\label{fig:scenario-partition-stability}
\end{figure}

The predominant value directions already emerge in the 240- and 480-scenario subsets, supporting the adequacy of the 960-scenario sample for characterizing broad value orientations. The two 480-scenario halves retain 142 of 144 directions across both methods, and the 240-scenario quarters retain 266 of 288 (Figure~\ref{fig:scenario-partition-stability}). The changes occur on IDV and UAI, where full-sample scores are close to zero.

\subsection{Stability across Repeated Samples}
\label{app:multi-sample}
The main analysis generates one image per prompt, so its results could depend on which image happens to be drawn. To test this, we generate more images for the same prompts.

\textbf{Setup.} Because generating with commercial systems is costly, we use the three open-weight systems that we run on our own GPUs: HiDream + Refiner, FLUX + Upsampler and Ideogram Magic. We use 240 scenarios (one fixed quarter from Appendix~\ref{app:scenario-partition-stability}, 40 per dimension) and English prompts with no cue and with the China cue. For each prompt, sample 1 is the original image and samples 2--5 are new. Each new sample changes two sources of randomness: the random seed of the diffusion model, and the output of the system's prompt rewriter, which is sampled again for every image. All other generation settings are unchanged. We score the new images with the same verifiers and protocol. Because this subset has only 40 scenarios per dimension, we score every image on all six dimensions (Appendix~\ref{app:six-dimension-scoring}), so that each score pools all 240 scenarios. Table~\ref{tab:multi-sample} reports each score's mean and range over the five samples.

\begin{table}[!tbp]
\centering\footnotesize
\setlength{\tabcolsep}{3pt}
\newcommand{\msc}[3]{\begin{tabular}[t]{@{}r@{}}#1\\[-2pt]{\tiny[#2, #3]}\end{tabular}}
\caption{Orientation scores across five samples per prompt (240 scenarios, English prompts; every image scored on all six dimensions). Each cell gives the mean over samples 1--5 and, in brackets, the minimum and maximum. Sample 1 is the image used in the main analysis.}\label{tab:multi-sample}
\begin{tabular*}{\linewidth}{@{\extracolsep{\fill}}lrrrrrr@{}}
\toprule
System & PDI & IDV & MAS & UAI & LTO & IVR\\
\midrule
\multicolumn{7}{@{}l}{\textit{IAE, no cue}}\\
HiDream + Refiner & \msc{$-$0.91}{$-$0.98}{$-$0.88} & \msc{$+$0.30}{$+$0.18}{$+$0.38} & \msc{$-$0.25}{$-$0.34}{$-$0.17} & \msc{$+$0.19}{$+$0.05}{$+$0.33} & \msc{$+$0.56}{$+$0.48}{$+$0.68} & \msc{$+$0.81}{$+$0.76}{$+$0.88}\\
FLUX + Upsampler & \msc{$-$0.87}{$-$0.93}{$-$0.76} & \msc{$+$0.37}{$+$0.28}{$+$0.48} & \msc{$-$0.24}{$-$0.28}{$-$0.18} & \msc{$+$0.02}{$-$0.03}{$+$0.10} & \msc{$+$0.66}{$+$0.53}{$+$0.73} & \msc{$+$0.73}{$+$0.63}{$+$0.87}\\
Ideogram Magic & \msc{$-$0.87}{$-$0.92}{$-$0.83} & \msc{$+$0.43}{$+$0.37}{$+$0.48} & \msc{$-$0.26}{$-$0.31}{$-$0.21} & \msc{$+$0.26}{\phantom{+}0.00}{$+$0.49} & \msc{$+$0.60}{$+$0.57}{$+$0.63} & \msc{$+$0.82}{$+$0.75}{$+$0.87}\\
\midrule
\multicolumn{7}{@{}l}{\textit{IAE, China cue}}\\
HiDream + Refiner & \msc{$-$0.63}{$-$0.70}{$-$0.55} & \msc{$+$0.09}{$+$0.02}{$+$0.21} & \msc{$-$0.19}{$-$0.25}{$-$0.14} & \msc{$+$0.13}{$-$0.03}{$+$0.22} & \msc{$+$0.67}{$+$0.61}{$+$0.73} & \msc{$+$0.79}{$+$0.75}{$+$0.83}\\
FLUX + Upsampler & \msc{$-$0.51}{$-$0.59}{$-$0.41} & \msc{$+$0.16}{$+$0.07}{$+$0.25} & \msc{$-$0.16}{$-$0.22}{$-$0.10} & \msc{$+$0.08}{$-$0.11}{$+$0.17} & \msc{$+$0.58}{$+$0.48}{$+$0.67} & \msc{$+$0.74}{$+$0.66}{$+$0.83}\\
Ideogram Magic & \msc{$-$0.60}{$-$0.84}{$-$0.37} & \msc{$+$0.01}{$-$0.10}{$+$0.14} & \msc{$-$0.22}{$-$0.29}{$-$0.16} & \msc{$+$0.26}{$+$0.16}{$+$0.31} & \msc{$+$0.53}{$+$0.47}{$+$0.57} & \msc{$+$0.72}{$+$0.64}{$+$0.78}\\
\midrule
\multicolumn{7}{@{}l}{\textit{SAE, no cue}}\\
HiDream + Refiner & \msc{$-$0.97}{$-$1.00}{$-$0.89} & \msc{$+$0.24}{$+$0.10}{$+$0.30} & \msc{$-$0.27}{$-$0.35}{$-$0.16} & \msc{$-$0.24}{$-$0.33}{$-$0.18} & \msc{$+$0.36}{$+$0.13}{$+$0.58} & \msc{$+$0.83}{$+$0.79}{$+$0.88}\\
FLUX + Upsampler & \msc{$-$0.95}{$-$1.00}{$-$0.83} & \msc{$+$0.48}{$+$0.31}{$+$0.65} & \msc{$-$0.31}{$-$0.38}{$-$0.11} & \msc{$-$0.49}{$-$0.78}{$-$0.33} & \msc{$+$0.40}{$+$0.30}{$+$0.48} & \msc{$+$0.63}{$+$0.44}{$+$0.82}\\
Ideogram Magic & \msc{$-$0.94}{$-$1.00}{$-$0.88} & \msc{$+$0.40}{$+$0.30}{$+$0.50} & \msc{$-$0.37}{$-$0.45}{$-$0.21} & \msc{$-$0.31}{$-$0.60}{\phantom{+}0.00} & \msc{$+$0.49}{$+$0.41}{$+$0.60} & \msc{$+$0.80}{$+$0.72}{$+$0.85}\\
\midrule
\multicolumn{7}{@{}l}{\textit{SAE, China cue}}\\
HiDream + Refiner & \msc{$-$0.69}{$-$0.88}{$-$0.54} & \msc{$+$0.09}{$-$0.02}{$+$0.22} & \msc{$-$0.30}{$-$0.36}{$-$0.17} & \msc{$-$0.26}{$-$0.38}{$-$0.14} & \msc{$+$0.44}{$+$0.27}{$+$0.50} & \msc{$+$0.81}{$+$0.72}{$+$0.89}\\
FLUX + Upsampler & \msc{$-$0.72}{$-$0.90}{$-$0.47} & \msc{$+$0.26}{$+$0.17}{$+$0.31} & \msc{$-$0.29}{$-$0.39}{$-$0.16} & \msc{$-$0.35}{$-$0.71}{\phantom{+}0.00} & \msc{$+$0.27}{$+$0.11}{$+$0.52} & \msc{$+$0.62}{$+$0.45}{$+$0.80}\\
Ideogram Magic & \msc{$-$0.88}{$-$1.00}{$-$0.76} & \msc{$+$0.06}{$-$0.06}{$+$0.18} & \msc{$-$0.20}{$-$0.28}{$-$0.05} & \msc{$-$0.29}{$-$0.38}{$-$0.14} & \msc{$+$0.44}{$+$0.26}{$+$0.60} & \msc{$+$0.78}{$+$0.63}{$+$0.93}\\
\bottomrule
\end{tabular*}
\end{table}

\textbf{Results.}
\begin{itemize}[leftmargin=*,itemsep=2pt,topsep=2pt]
\item \textit{Profile directions do not change.} Without a cue, 87 of 90 IAE scores and 89 of 90 SAE scores (3 systems $\times$ 6 dimensions $\times$ 5 samples) have the same sign as in the main results. The exceptions are UAI scores, which are already close to zero (within $0.04$ of zero), so a small change can flip their sign.
\item \textit{The China-cue effect does not change.} Under IAE, adding the China cue raises PDI and lowers IDV for all three systems in all five samples.
\item \textit{Individual images vary, but rarely in direction.} Under IAE, for the same prompt and dimension, the new samples express opposite orientations in only 3--4.5\% of cases; most differences are in whether an orientation is expressed at all.
\end{itemize}
In summary, our results are stable across repeated samples. Each score aggregates over many scenarios, so the variation of individual images largely averages out, and the conclusions do not depend on which image is drawn for each prompt.

\subsection{Graded Strength Weighting}
\label{app:graded-score}
Our score treats an orientation as supported when at least two verifiers accept it, which gives strong and weak evidence the same weight. As a robustness check, we instead weight each image by the share of verifiers that support the orientation ($1/3$, $2/3$ or $1$) and recompute every score on each dimension's own scenarios. Binary weights are the special case of $0/1$ weights and reproduce the pooled profiles in Table~\ref{tab:full-profiles} exactly. With graded weights, all 72 scores (six systems, six dimensions, both methods) keep their sign, and no score changes by more than $0.076$.

\subsection{Scoring All Six Dimensions}
\label{app:six-dimension-scoring}
The main analysis scores each image only on its scenario's source dimension, because scenarios are constructed to open that dimension and may favor an orientation on another one (Appendix~\ref{app:source-dimension-scoring}). Here we report the alternative used in the original design, in which every image is scored on all six dimensions with the same images, conditions, and votes.

The two scorings agree on the main conclusions. Under both, systems favor low power distance, individualism, care and quality of life, long-term orientation, and indulgence, UAI is the one dimension whose direction depends on the evaluation strategy, China cues raise PDI and lower IDV, and prompt language has a small effect relative to geographic cues. Tables~\ref{tab:full-profiles-six-dim} and~\ref{tab:prompt-change-summary-six-dim} give the six-dimension value profiles and paired score changes.

The scorings differ in two ways. First, scoring all six dimensions pools the images of all 960 scenarios into every cell, so each estimate uses roughly six times as many images and its interval is narrower; scoring the source dimension uses only that dimension's 160 scenarios. Second, non-source labels describe what a scenario invites rather than what the system contributes: a vacation scene, for instance, supports indulgence under every system (Appendix~\ref{app:source-dimension-scoring}). The source-dimension scores in the main text therefore isolate system behavior more cleanly, at the cost of wider intervals. The two scorings can differ on individual contrasts; we report source-dimension scores in the main text and use the six-dimension results as a robustness check.

\begin{AppendixResultBlock}{table}
\centering\small
\setlength{\tabcolsep}{4pt}
\caption{Pooled orientation scores under the six shared conditions. Bold: within-method simultaneous 95\% interval excludes zero.}\label{tab:full-profiles-six-dim}
\begin{tabular}{@{}llrrrrrr@{}}
\toprule
Method & System & PDI & IDV & MAS & UAI & LTO & IVR\\\midrule
SAE & MAI & $\mathbf{-0.886}$ & $\mathbf{0.176}$ & $\mathbf{-0.420}$ & $\mathbf{-0.239}$ & $\mathbf{0.733}$ & $\mathbf{0.777}$\\
SAE & GPT Image 2 & $\mathbf{-0.805}$ & $\mathbf{0.273}$ & $\mathbf{-0.250}$ & $-0.124$ & $\mathbf{0.539}$ & $\mathbf{0.787}$\\
SAE & Nano Banana 2 & $\mathbf{-0.887}$ & $\mathbf{0.297}$ & $\mathbf{-0.383}$ & $\mathbf{-0.276}$ & $\mathbf{0.487}$ & $\mathbf{0.885}$\\
SAE & FLUX + Upsampler & $\mathbf{-0.704}$ & $\mathbf{0.458}$ & $\mathbf{-0.304}$ & $-0.271$ & $\mathbf{0.398}$ & $\mathbf{0.890}$\\
SAE & HiDream + Refiner & $\mathbf{-0.741}$ & $\mathbf{0.374}$ & $\mathbf{-0.284}$ & $\mathbf{-0.275}$ & $\mathbf{0.376}$ & $\mathbf{0.919}$\\
SAE & Ideogram Magic & $\mathbf{-0.845}$ & $\mathbf{0.348}$ & $\mathbf{-0.345}$ & $-0.237$ & $\mathbf{0.374}$ & $\mathbf{0.868}$\\
\midrule
IAE & MAI & $\mathbf{-0.854}$ & $\mathbf{0.089}$ & $\mathbf{-0.267}$ & $-0.042$ & $\mathbf{0.688}$ & $\mathbf{0.581}$\\
IAE & GPT Image 2 & $\mathbf{-0.740}$ & $\mathbf{0.196}$ & $\mathbf{-0.164}$ & $\mathbf{0.165}$ & $\mathbf{0.546}$ & $\mathbf{0.596}$\\
IAE & Nano Banana 2 & $\mathbf{-0.785}$ & $\mathbf{0.256}$ & $\mathbf{-0.219}$ & $\mathbf{0.187}$ & $\mathbf{0.594}$ & $\mathbf{0.807}$\\
IAE & FLUX + Upsampler & $\mathbf{-0.631}$ & $\mathbf{0.341}$ & $\mathbf{-0.206}$ & $0.142$ & $\mathbf{0.531}$ & $\mathbf{0.838}$\\
IAE & HiDream + Refiner & $\mathbf{-0.709}$ & $\mathbf{0.323}$ & $\mathbf{-0.205}$ & $\mathbf{0.182}$ & $\mathbf{0.575}$ & $\mathbf{0.848}$\\
IAE & Ideogram Magic & $\mathbf{-0.740}$ & $\mathbf{0.293}$ & $\mathbf{-0.230}$ & $\mathbf{0.285}$ & $\mathbf{0.479}$ & $\mathbf{0.789}$\\
\bottomrule
\end{tabular}
\end{AppendixResultBlock}

\begin{AppendixResultBlock}{table}
\centering\footnotesize
\setlength{\tabcolsep}{4pt}
\caption{Paired score changes: 36 system--dimension comparisons per contrast and method. Med./Max.: median/maximum $|\Delta S|$; Keep: number of unchanged score signs; Sim.: number of shifts with simultaneous 95\% intervals excluding zero.}\label{tab:prompt-change-summary-six-dim}
\begin{tabular}{@{}lrrrrrrrr@{}}
\toprule
& \multicolumn{4}{c}{SAE} & \multicolumn{4}{c}{IAE}\\
\cmidrule(lr){2-5}\cmidrule(l){6-9}
Contrast; fixed condition & Med. & Max. & Keep & Sim. & Med. & Max. & Keep & Sim.\\\midrule
EN $\to$ ZH; no cue & 0.052 & 0.242 & 36 & 2 & 0.057 & 0.205 & 36 & 5\\
EN $\to$ ZH; US cue & 0.046 & 0.248 & 36 & 0 & 0.031 & 0.187 & 36 & 1\\
EN $\to$ ZH; China cue & 0.049 & 0.270 & 36 & 0 & 0.035 & 0.148 & 36 & 0\\
\midrule
none $\to$ US; EN & 0.045 & 0.193 & 36 & 0 & 0.027 & 0.140 & 36 & 1\\
none $\to$ China; EN & 0.078 & 0.372 & 36 & 11 & 0.117 & 0.337 & 35 & 18\\
US $\to$ China; EN & 0.074 & 0.409 & 36 & 10 & 0.092 & 0.366 & 35 & 16\\
none $\to$ US; ZH & 0.054 & 0.252 & 36 & 0 & 0.043 & 0.190 & 36 & 4\\
none $\to$ China; ZH & 0.075 & 0.327 & 36 & 3 & 0.090 & 0.253 & 35 & 9\\
US $\to$ China; ZH & 0.111 & 0.454 & 36 & 7 & 0.097 & 0.335 & 35 & 17\\
\bottomrule
\end{tabular}
\end{AppendixResultBlock}

\FloatBarrier
\section{Prompts and Annotation Materials}
\label{app:full-materials}

This appendix summarizes the prompts and decision criteria in manuscript terminology. The full prompts are available in our GitHub repository (\repourl{}).

\subsection{Pipeline Index}
\label{app:prompts}

Table~\ref{tab:prompt-overview} summarizes the inputs, outputs, and call counts for construction and annotation. Following the prompt design of \citet{lyu2026syntra}, each prompt is organized into labeled sections, such as task instructions, decision rules, an output contract, and the input, rather than written as free-form text.

\begin{table}[!hb]
\definecolor{promptHeader}{HTML}{EEEEEE}
\definecolor{promptConstruction}{HTML}{EDF3FA}
\definecolor{promptConstructionText}{HTML}{355779}
\definecolor{promptEvaluation}{HTML}{EAF4F1}
\definecolor{promptEvaluationText}{HTML}{346B60}
\definecolor{promptRule}{HTML}{66717D}
\hypersetup{hidelinks}
\arrayrulecolor{promptRule}
\caption{Prompts used to build and annotate \method{}, in pipeline order. Inputs and outputs are summarized; full prompts are in our repository (\repourl{}).}
\label{tab:prompt-overview}
\centering
\footnotesize
\setlength{\tabcolsep}{4pt}
\renewcommand{\arraystretch}{1.18}
\begin{tabularx}{\linewidth}{@{}c>{\raggedright\arraybackslash}p{0.24\linewidth}>{\raggedright\arraybackslash}X>{\raggedright\arraybackslash}p{0.26\linewidth}>{\raggedright\arraybackslash}p{0.125\linewidth}@{}}
\toprule
\rowcolor{promptHeader}
\multicolumn{1}{@{}>{\columncolor{promptHeader}[0pt][\tabcolsep]}c}{\rule{0pt}{2.5ex}\#} & \textbf{Prompt} & \textbf{Key inputs} & \textbf{Main outputs} & \multicolumn{1}{>{\columncolor{promptHeader}[\tabcolsep][0pt]}p{0.125\linewidth}@{}}{\textbf{Calls}} \\
\midrule
\multicolumn{5}{@{}>{\columncolor{promptConstruction}[0pt][0pt]}l@{}}{\rule[-0.7ex]{0pt}{3.2ex}\textcolor{promptConstructionText}{\textbf{Scenario construction} (Section~\ref{sec:benchmark})}} \\
\textendash & Concept extraction & CDEval item + metadata & Concept + rationale & 1/item \\
\addlinespace[3pt]
\textendash & Scenario generation & Same as concept extraction + concept & Scenario + rationale & 1/concept \\
\addlinespace[3pt]
\textendash & Translation & English scenario & Chinese, Thai, Hindi scenarios & 1/scenario \\
\addlinespace[4pt]
\multicolumn{5}{@{}>{\columncolor{promptEvaluation}[0pt][0pt]}l@{}}{\rule[-0.7ex]{0pt}{3.2ex}\textcolor{promptEvaluationText}{\textbf{Annotation} (Section~\ref{sec:annotation})}} \\
1 & SAE\newline proposal & Scenario & Potential orientation evidence & 3/scenario \\
\addlinespace[3pt]
2 & IAE\newline proposal & Image + scenario & Candidate orientation evidence & 3/image \\
\addlinespace[3pt]
3 & SAE\newline verification & Image + scenario + pooled candidates from 1 & Candidate reviews + 12 orientation decisions & 3/image \\
\addlinespace[3pt]
4 & IAE\newline verification & Image + scenario + pooled candidates from 2 & Candidate reviews + added evidence + 12 orientation decisions & 3/image \\
\bottomrule
\end{tabularx}
\par\vspace{4pt}
\begin{minipage}{\linewidth}
\scriptsize\raggedright
\textit{Inputs.} A CDEval item contains a question and two options; metadata includes its domain, dimension, and orientation descriptions. Highlighted fields such as \PVar{SCENARIO} are filled at run time.\par
\smallskip
\textit{Models and calls.} Concept extraction and scenario generation use DeepSeek-V4-Flash; translation uses Gemini 3.7 Flash. Each annotation prompt runs independently on GPT-5.6 Luna, Gemini 3.7 Flash, and Muse Spark 1.3 (three calls in total). Each call uses one user message; in Prompts 2--4, the image follows the text.
\end{minipage}
\arrayrulecolor{black}
\end{table}
\subsection{Prompt Settings and Rendering}
\label{app:prompts-construction}\label{app:filtering-prompt}\label{app:prompts-annotation}
The full prompts for concept extraction, scenario generation, translation, scenario filtering, and evidence proposal and verification are available in our repository (\repourl{}). Concept extraction and scenario generation use DeepSeek-V4-Flash at temperature~0, and their rationales are never sent to image generators. Translation uses Gemini 3.7 Flash with medium thinking and a 4,096-token output limit, preserving each scenario's detail, alternatives and deliberate omissions without cultural adaptation. The four annotation prompts use JSON schemas and run independently on GPT-5.6 Luna, Gemini 3.7 Flash and Muse Spark 1.3 with medium reasoning effort; output limits are 8,192 tokens for Luna and Gemini (32,768 for Gemini completion runs) and 32,768 for Muse. The decision cards shared by all annotation prompts are summarized in Appendix~\ref{app:vlm-rubric}.

Prompt rendering uses no model: the language's base prompt wraps the scenario \PVar{SCENARIO}, and a geographic cue appends to the rendered prompt \PVar{PROMPT} one sentence naming the country \PVar{CULTURE} (Table~\ref{tab:rendering-templates}).

\begin{table}[!tbp]
\caption{Prompt rendering templates. Thai drops terminal punctuation before the cue.}
\label{tab:rendering-templates}
\centering
\footnotesize
\begin{tabularx}{\linewidth}{@{}l>{\raggedright\arraybackslash}X>{\raggedright\arraybackslash}X>{\raggedright\arraybackslash}p{0.22\linewidth}@{}}
\toprule
Language & Base prompt & Geographic-cue template & Country labels \\
\midrule
English & \PCode{Generate an image: }\PVar{SCENARIO}\PCode{} & \PCode{}\PVar{PROMPT}\PCode{. Set the scene in }\PVar{CULTURE}\PCode{.} & the United States\newline China\newline Thailand\newline India \\
Chinese & \PScript{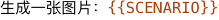} & \PScript{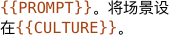} & \PScript{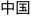}\newline \PScript{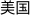} \\
Thai & \PScript{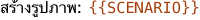} & \PScript{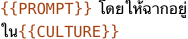} & \PScript{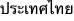}\newline \PScript{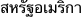} \\
Hindi & \PScript{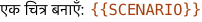} & \PScript{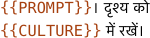} & \PScript{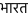}\newline \PScript{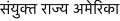} \\
\bottomrule
\end{tabularx}
\end{table}

\subsection{Decision Cards for Automatic Annotation}
\label{app:vlm-rubric}
\label{app:shared-rules}
\label{app:vlm-cards}

The four annotation prompts share identical dimension cards, condensed in Table~\ref{tab:vlm-cards}. All cards share five attribution rules. \emph{Scope:} the scenario supplies context such as alternatives, the object of a response and an arrangement's purpose, while the image decides any open outcome and prevails if the two conflict. \emph{Evidence:} support connects the depiction, its role in the focal situation and the card's criterion; readable text counts in context, but nationality, ethnicity, gender and cultural markers alone never do. Each orientation needs its own positive support, without proof of private motives or a depicted opposite. \emph{Choices:} an indicated choice or a clearly realized alternative can suffice; merely displaying alternatives cannot. \emph{Responses and arrangements:} depicting what someone responds to does not by itself establish their stance. If an arrangement is the focal matter, assess its depiction directly. \emph{Applicability:} each card first identifies its defining issue in the focal situation; topical association alone does not establish relevance.

\begin{table}[!tbp]
\centering\scriptsize
\setlength{\tabcolsep}{3pt}
\renewcommand{\arraystretch}{1.12}
\caption{Condensed decision cards for automatic annotation. The last column lists cues that do not settle the dimension on their own.}
\label{tab:vlm-cards}
\begin{tabularx}{\linewidth}{@{}l>{\raggedright\arraybackslash}p{0.2\linewidth}>{\raggedright\arraybackslash}X>{\raggedright\arraybackslash}p{0.24\linewidth}@{}}
\toprule
Dim. & Defining issue & Orientation criteria & Insufficient alone \\
\midrule
PDI & Authority or control over decisions, resources or participation, and how unequal power is treated & \emph{High:} higher standing is treated as entitling control, deference, or limits on others' voice. \emph{Low:} their standing to contribute, question or seek accountability is acknowledged or asserted, even if a leader keeps final authority. & Following instructions (for High); disagreement, informality or friendliness (for Low) \\
\addlinespace[2pt]
IDV & A personal choice, goal, responsibility or belonging relative to a lasting in-group & \emph{Individualism:} the matter is the person's own to direct, including privacy and self-reliance. \emph{Collectivism:} membership, harmony, loyalty or mutual commitments shape the choice, responsibility or identity. & Separate or shared activity, which may be task allocation or coordination; help and group ties \\
\addlinespace[2pt]
MAS & What the depiction treats as worth pursuing or maintaining & \emph{Achievement:} accomplishment, advancement, competitive standing or material success is valued. \emph{Care:} needs, well-being, relationships or quality of life matter. & Working, helping or holding a care role; effort, wealth, mood or gender composition \\
\addlinespace[2pt]
UAI & What is uncertain, ambiguous or unfamiliar for the participants & \emph{High:} unease with uncertainty, or definite answers and fixed procedures treated as needed assurance. \emph{Low:} ambiguity, provisional answers or flexibility accepted while something remains unsettled. & Error prevention, information gathering, precautions against known hazards \\
\addlinespace[2pt]
LTO & Present investment and expected returns, or established practices and expectations & \emph{Long-term:} commitments serve returns worth waiting for, or a practice is pragmatically adapted. \emph{Short-term:} quicker returns guide effort, or traditions and established expectations guide conduct. & Future benefits, scheduling or fast completion; old and new elements shown together \\
\addlinespace[2pt]
IVR & The place given to enjoying life, leisure or fun & \emph{Indulgence:} enjoyment is freely pursued, welcomed or accommodated. \emph{Restraint:} enjoyment is curbed or subordinated to norms or duties. & Entertainment, consumption, smiles or luxury; practical limits such as budgets or safety \\
\bottomrule
\end{tabularx}
\end{table}

\FloatBarrier
\subsection{Decision Cards for Human Review}
\label{app:human-rubric}

\begin{figure}[!tbp]
\begin{minipage}[t]{0.25\linewidth}
\vspace{0pt}\includegraphics[width=\linewidth]{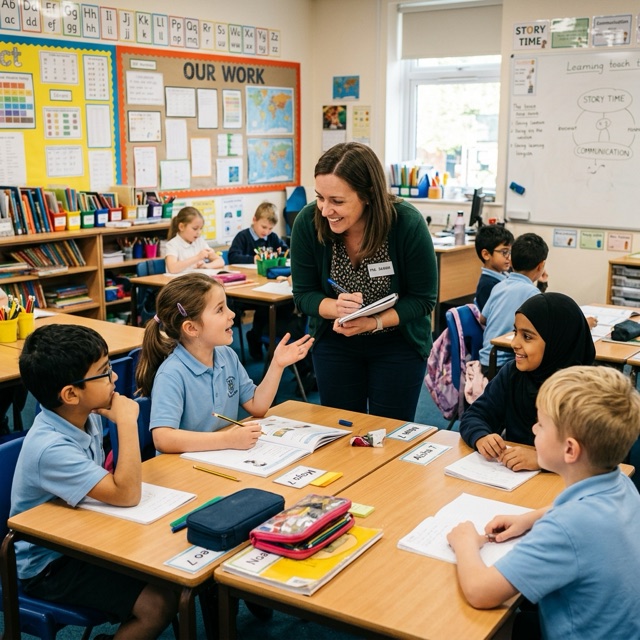}
\end{minipage}\hfill
\begin{minipage}[t]{0.71\linewidth}
\vspace{0pt}\small\raggedright
\textbf{Prompt:} a teacher and students communicating in a classroom.\par\smallskip
\textbf{1. Identify the power relationship.} The teacher has authority; the students are less powerful. The matter is a student's opportunity to speak and be heard.\par\smallskip
\textbf{2. Examine its treatment.} A student speaks and gestures while the teacher leans toward her and listens with a notepad. The response treats her contribution as worth hearing and considering.\par\smallskip
\textbf{Interpretation: Low PDI.} The student's voice is welcomed within the teacher--student relationship even though the teacher retains authority.
\end{minipage}
\caption{The human-review PDI worked example, condensed.}
\label{fig:human-card-pdi}
\end{figure}

Reviewers use plain-language versions of the same six cards on the review page, each with a worked two-step example; Figure~\ref{fig:human-card-pdi} shows the PDI example.

\FloatBarrier
\section{Limitations and Future Directions}
\label{sec:limitations}

\paragraph{From cultural constructs to visual evidence.}
Our evaluation operationalizes cultural values through Hofstede's six dimensions and their associated visual criteria. Still images provide limited access to motives, long-term commitments, and broader social relationships. The same depicted behavior may also support different interpretations across cultural contexts. Our findings therefore concern expressed value orientations under the adopted rubric. Cultural researchers and communities could validate these mappings and compare alternative value frameworks.

\paragraph{Dependence on the scenario distribution.}
The observed value profiles depend on the benchmark's scenario distribution. Leaving value orientations unspecified does not ensure that opposing orientations are equally easy to depict or recognize. Translating a shared scenario bank supports controlled comparisons across prompt languages but provides limited coverage of locally specific situations. Community-authored scenarios could test whether the observed homogeneity persists across social contexts.

\paragraph{Evaluator dependence and validation scope.}
Evidence proposal and verification rely on VLMs, which may share omissions or interpretive biases. Human validation involved nine annotators and 60 images, using model-proposed candidates with opportunities to add evidence. This design measures agreement primarily within a shared candidate bank and provides limited evidence about the completeness of evidence proposal. Broader validation across reviewers, scenarios, and T2I systems could assess generalizability.

\paragraph{System-level evaluation and component attribution.}
We evaluate each T2I system as a whole, including its associated prompt-processing components where applicable. Our analysis does not disentangle the contributions of these components and the underlying image-generation model to the expressed value orientations. Future work could characterize their respective contributions and interactions to better understand how value orientations emerge throughout the generation pipeline.

\paragraph{Measurement resolution and downstream use.}
Binary support judgments record whether an orientation is present, not how strongly it is expressed. We tested whether VLM judges can measure strength in a pairwise study; they agree with each other but rarely treat two images as equally strong when humans do, so they cannot yet measure strength reliably (Appendix~\ref{app:pairwise-strength}). A graded, strength-aware score is therefore future work.

\end{document}